\documentclass{article}

\usepackage[preprint]{neurips_2025}

\usepackage{eccvabbrv}
\usepackage[table,x11names,dvipsnames]{xcolor}
\usepackage{colortbl}

\usepackage{tcolorbox,titletoc}
\newcommand{\authcount}[1]{}
\usepackage{xstring,catchfile}

\definecolor{winGreen}{HTML}{009900}
\definecolor{lossRed}{HTML}{CC0000}
\definecolor{bestGreen}{RGB}{34,139,34}  

\definecolor{boxborder}{HTML}{006064} 
\definecolor{boxbg}{HTML}{E0F7FA}     

\definecolor{agentblue}{HTML}{146082}

\definecolor{headerblue}{HTML}{2C3E50}
\definecolor{headertextwhite}{HTML}{FFFFFF}
\definecolor{row1}{HTML}{F0F4F8}
\definecolor{row2}{HTML}{FFFFFF}
\definecolor{catblue}{HTML}{E8EEF4}
\definecolor{eccvblue}{rgb}{0.12,0.49,0.85}

\definecolor{grayGPT5}{gray}{0.96}
\definecolor{blueGPT4}{rgb}{0.94, 0.96, 1.0}
\definecolor{greenQwen}{rgb}{0.95, 0.99, 0.95}

\usepackage{amsmath,amsfonts,bm}

\def\eqref#1{equation~\ref{#1}}

\def\1{\bm{1}}

\newcommand{\methodname}{\textit{ArcDeck}}
\newcommand{\benchname}{\textit{ArcBench}}

\DeclareMathAlphabet{\mathsfit}{\encodingdefault}{\sfdefault}{m}{sl}
\SetMathAlphabet{\mathsfit}{bold}{\encodingdefault}{\sfdefault}{bx}{n}

\usepackage{amssymb}
\usepackage{amsmath}
\usepackage{wrapfig}
\usepackage{tabularx}
\usepackage{graphicx}
\usepackage{booktabs}
\usepackage{multirow}
\usepackage{longtable}
\usepackage{array}
\usepackage{arydshln}
\usepackage{makecell}
\usepackage{diagbox}
\usepackage{subcaption}
\usepackage{siunitx}
\usepackage{algorithm}
\usepackage{algpseudocode}

\usepackage[utf8]{inputenc}
\usepackage{enumitem}
\usepackage{setspace}
\usepackage{romannum}
\usepackage[normalem]{ulem}

\usepackage{pifont}         
\usepackage{bbding}
\usepackage{fontawesome5}
\usepackage{etoolbox}

\usepackage{tikz}

\usepackage[accsupp]{axessibility}

\usepackage[colorlinks=true, urlcolor=teal, citecolor=blue, linkcolor=red]{hyperref}
\usepackage{cleveref} 

\usepackage{orcidlink}
\newcommand{\cmark}{\textcolor{green!60!black}{\ding{51}}}
\newcommand{\xmark}{\textcolor{red!60!black}{\ding{55}}}


\newcommand{\agentnum}[1]{\tikz[baseline=-0.85ex]{%
    \node[shape=circle, fill=agentblue, draw=black, text=white,
          inner sep=1pt, font=\scriptsize\bfseries, minimum size=4mm] (char) {#1};}}

\begin{document}

\pagenumbering{arabic}

\noindent
\begin{tcolorbox}[colback=boxbg, colframe=boxborder, arc=3mm, boxrule=1pt, top=6mm, bottom=6mm, left=6mm, right=6mm]

\begin{center}
    {\LARGE \bf \textsf{Narrative-Driven Paper-to-Slide Generation via ArcDeck} \par}
    \vspace{6mm}
    {\large \textbf{Tarik Can Ozden}$^{*1}$, \textbf{Sachidanand VS}$^{*1}$, \textbf{Furkan Horoz}$^{*1,2}$, \\ \textbf{Ozgur Kara}$^1$, \textbf{Junho Kim}$^{1\dagger}$, \textbf{James M. Rehg}$^{1\dagger}$ \par}
    \vspace{2mm}
    {\normalsize $^1$University of Illinois Urbana-Champaign, $^2$Middle East Technical University \par}
    {\small $^*$Equal contribution~\quad $^\dagger$Corresponding author \par}
\end{center}

\noindent\textcolor{gray}{\rule{\linewidth}{0.4pt}}

\noindent \textbf{Abstract.} We introduce ArcDeck, a multi-agent framework that formulates paper-to-slide generation as a structured narrative reconstruction task. Unlike existing methods that directly summarize raw text into slides, ArcDeck explicitly models the source paper's logical flow. It first parses the input to construct a discourse tree and establish a global commitment document, ensuring the high-level intent is preserved. These structural priors then guide an iterative multi-agent refinement process, where specialized agents iteratively critique and revise the presentation outline before rendering the final visual layouts and designs. To evaluate our approach, we also introduce ArcBench, a newly curated benchmark of academic paper-slide pairs. Experimental results demonstrate that explicit discourse modeling, combined with role-specific agent coordination, significantly improves the narrative flow and logical coherence of the generated presentations.
\vspace{6mm}

\noindent \textbf{GitHub:} \url{https://github.com/RehgLab/ArcDeck} \\
\textbf{Project Webpage:} \url{https://arcdeck.org/} \\

\end{tcolorbox}

\vspace{0.5cm}

\begin{figure*}[h!]
 \centering
 \includegraphics[width=\linewidth]{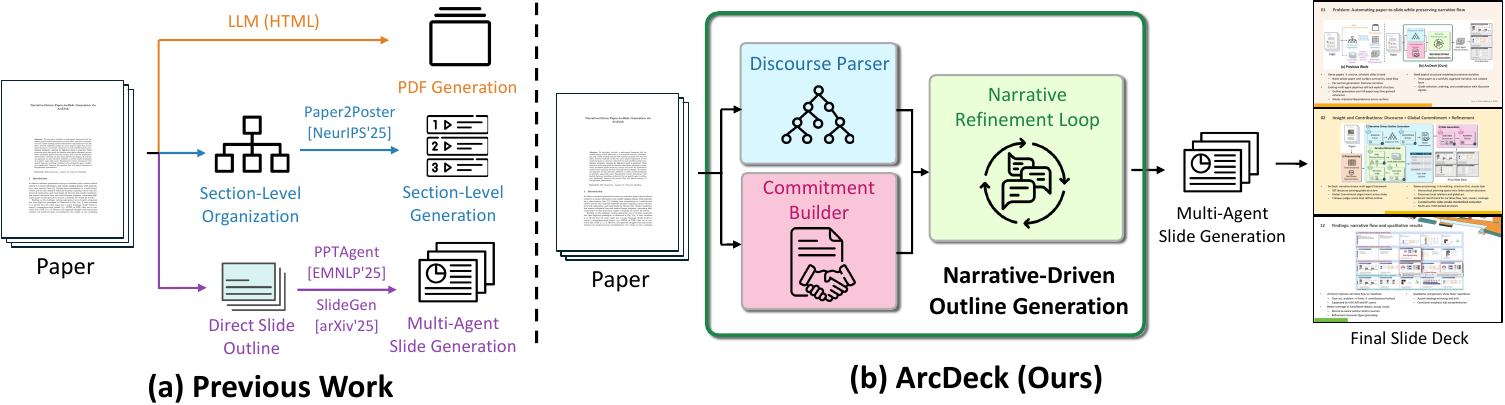}
 \caption{\textbf{Framework Comparison.} (a) Prior methods rely on \textcolor[HTML]{E67E22}{raw-text summarization}, isolated \textcolor[HTML]{2980B9}{section-level organization}, or \textcolor[HTML]{8E44AD}{direct outline extraction}, yielding condensed text without narrative guidance. (b) In contrast, \methodname\ explicitly models discourse structure and establishes global commitments to produce narrative-driven outlines through an iterative multi-agent refinement loop.} 
 \label{fig:1}
\end{figure*}

\section{Introduction}
\label{sec:intro}

An effective academic presentation conveys an academic paper's dense technical content in a concise, informative, and visually engaging manner while maintaining a clear narrative flow~\cite{reynolds2019presentation}. Crafting these presentations is a multi-faceted, complex process that requires deliberate decisions regarding content selection, structural organization, and visual design~\cite{bartsch2003effectiveness}. As a result of these complexities, the automatic generation of high-quality presentations (in the form of a slide deck) remains a significant challenge for modern AI systems.

The key challenge at the heart of slide generation is the need to identify the \emph{narrative structure} of the source paper, which specifies how concepts are introduced, explained, and related to each other in order to tell the story of the work, and then map that structure into the visual form of a slide deck, while at the same time extracting and translating the concepts, results, graphics, and other elements that make up the paper content. Prior approaches to automatic paper-to-slide generation can be organized into three categories, as illustrated in Fig.~\ref{fig:1}(a), based on their approach to addressing this challenge.
The most straightforward approach (orange arrow) is to  feed the entire paper directly into a Large Language Model (LLM) to output a presentation-ready format (\eg, HTML or XML) that can be converted into a PDF~\cite{sun2021d2s,bandyopadhyay2024docpress,cachola2024kctv}. However, this approach struggles with long context windows--- the model becomes overwhelmed by the volume of text, producing static, surface-level summaries that lack a cohesive flow.  
To address this concern, another category of works develop a section-level organization~\cite{pang2025paper2poster,zhang2025postergen} (blue arrow, Fig.~\ref{fig:1}(a)), processing each section independently before mapping them to pages and applying layout refinements~\cite{ge2025autopresent,zheng2025pptagent}. While this approach can identify the key organizing principles for the paper, it fundamentally fractures the narrative, as generating summaries in isolation creates temporally disconnected slide content that post-processing cannot repair. Recently, multi-agent frameworks~\cite{xu2025pregenie,liang2025slidegen} have begun to address this limitation (purple arrow, Fig.~\ref{fig:1}(a)) by first generating a global paper-level slide outline consisting of titles and summarized paragraphs, which is then passed to a specialized multi-agent generation module.
However, outline generation is itself a non-trivial task that requires hierarchical planning across multiple levels of abstraction, spanning both intra-section organization and overall paper structure. Moreover, because these models construct the outline by processing the entire paper at once, they again struggle to maintain fine-grained structural coherence in dense content.
In summary, prior works have lacked an effective approach to structural modeling, which has limited their ability to preserve rhetorical dependencies and orchestrate the unified storyline required for a compelling presentation.

In this paper, we propose \methodname, a narrative-driven multi-agent framework that formulates paper-to-slide generation as a structured narrative reconstruction problem. \methodname\ addresses the structural and contextual shortcomings of prior approaches through three core mechanisms (Fig.~\ref{fig:1}(b)). First, to overcome the lack of explicit structural guidance, our \textit{Discourse Parser} constructs a discourse tree inspired by Rhetorical Structure Theory (RST)~\cite{mann1987rst}. By explicitly defining the relational connections between text units (\eg, paragraphs), this hierarchical model guides initial structural planning and preserves local rhetorical relationships. Second, to prevent the overarching context loss typical of fragmented generation, a \textit{Commitment Builder} establishes a \textit{Global Commitment}. This module captures high-level intent and meta-information---such as the central thesis, scope, and overall narrative structure---and propagates them as persistent conditioning signals to ensure paper-level consistency. Third, recognizing that narrative coherence requires more than one-shot outlining, we incorporate a multi-agent \textit{Narrative Refinement Loop}. Through a closed-loop critique--judge--revise process, the outline is iteratively refined to improve structural coherence, enabling a more effective translation from hierarchical planning to slide-level presentation. Together, these mechanisms allow \methodname\ to seamlessly balance content organization with visual formatting, producing a highly coherent narrative flow. An overview of \methodname\ framework is illustrated in Fig.~\ref{fig:methodology}.

In addition, to support systematic evaluation, we introduce \benchname, a curated paper–slide benchmark collected from multiple academic venues and spanning diverse research topics. \benchname\ enables systematic comparison across multiple dimensions, including text coverage, narrative coherence, visual quality, and content retention, establishing a standardized evaluation protocol for paper-to-slide generation. We open source \methodname\ and \benchname\ to further research advances. 
To summarize, our main contributions are as follows:
\begin{itemize}
    \item We propose \methodname, a narrative-oriented multi-agent framework for paper-to-slide generation that, for the first time, explicitly models discourse relations through hierarchical RST-based analysis.
    \item We introduce global commitments and a closed-loop critique–judge–revise refinement mechanism, enabling coherent structural planning and consistent narrative flow across the entire slide sequence. Ours is the first multi-agent solution and the first to support customization (\eg, presentation length).
    \item We present \benchname, a high-quality and open-source paper–slide paired dataset curated from author-prepared presentations at top-tier venues, providing the community with a reliable human reference standard for evaluating slide generation quality.
\end{itemize}

\section{Related Work}
\label{sec:related}

\subsection{Agentic Slide Generation.}
\begin{figure}[t]
    \centering
    \includegraphics[width=\linewidth]{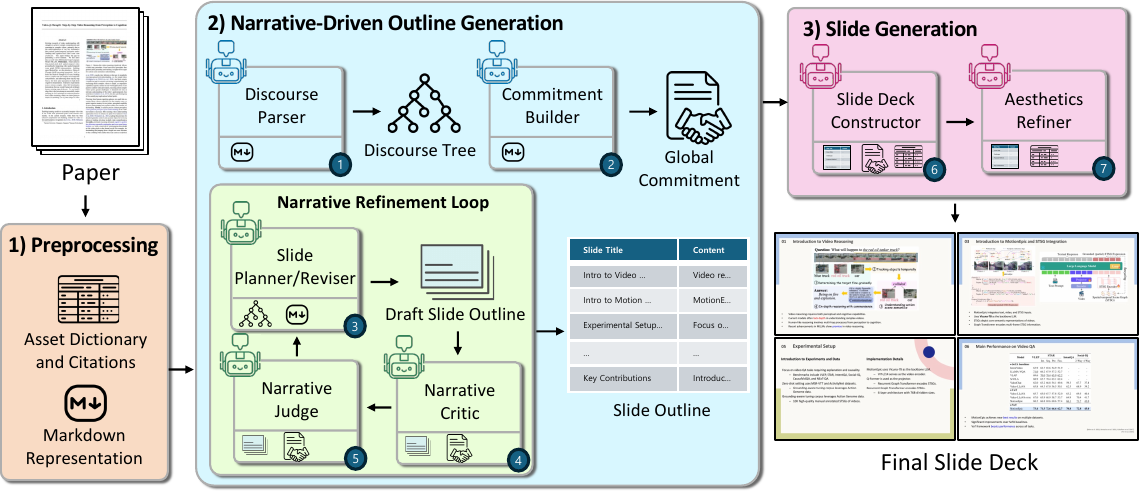}
    \caption{\textbf{\methodname\ Overview.} 
    1) \textcolor[HTML]{E5A073}{\textbf{Preprocessing}} extracts textual and visual assets from the source paper; 2) \textcolor[HTML]{65BCCC}{\textbf{Narrative-Driven Outline Generation}} combines discourse parsing and global commitment to guide the \textcolor[HTML]{B8CE9F}{Narrative Refinement Loop}, producing a structured Slide Outline; and 3) \textcolor[HTML]{D684B9}{\textbf{Slide Generation}} constructs the initial deck and refines its aesthetic layout to yield the final presentation.
    The numbered indices of each agent, \eg, \protect\agentnum{1}, denote their temporal execution order, while the icons at the bottom of each agent box indicate the specific structural inputs they receive. }
    \label{fig:methodology}
\end{figure}

Early slide generation methods framed the task as content selection via regression models~\cite{hu2013ppsgen}, while later deep learning approaches used RNNs to summarize sections into bullet points~\cite{sun2021d2s,fu2022doc2ppt}. The emergence of LLMs shifted the focus toward multi-agent and code-generation techniques~\cite{guo2024pptc, zheng2025pptagent, xu2025pregenie, pang2025paper2poster, liang2025slidegen}. Beyond early zero-shot editing~\cite{guo2024pptc}, frameworks like DocPres combined LLMs and VLMs for extraction and outlining~\cite{bandyopadhyay2024docpress}. Others improved organization using intermediate JSON format~\cite{cachola2024kctv} or reframed the task as code synthesis using \texttt{python-pptx} scripts~\cite{ge2025autopresent, pang2025paper2poster, liang2025slidegen}. To enhance visual fidelity, models were fine-tuned on description-code pairs~\cite{ge2025autopresent,touvron2023llama} or utilized reference-based template editing~\cite{zheng2025pptagent}. Recent multi-agent pipelines iteratively refine generated code~\cite{pang2025paper2poster, zhang2025postergen, xu2025pregenie}, adopting simpler frameworks like Slidev~\cite{xu2025pregenie}, or coordinating specialized agents for content, layout, and figure alignment~\cite{liang2025slidegen}. In contrast to these prior works, we are the first to utilize a multi-agent narrative refinement loop and the first to construct a global commitment (see Fig.~\ref{fig:methodology}) that explicitly captures the goals of content generation and supports customization, such as adapting the slides for different target presentation lengths.

\subsection{Rhetorical Structure Theory (RST).} RST~\cite{mann1987rst} models hierarchical discourse structure by defining rhetorical relations between text spans. It has been widely applied in downstream tasks such as summarization~\cite{marcu2000rst_summ, liu2019rst_summ1}, question answering~\cite{gao2020discern_rst_QA}, and sentiment analysis~\cite{bhatia2015rst_sentiment}. Existing RST parsers generally follow either top-down~\cite{kobayashi2020rst_topdown, nguyen2021rst_topdown1} or bottom-up~\cite{feng2012text_rst_bottomup, prendinger2009novel_rst_bottomup1} paradigms. As deep learning models gained prominence, end-to-end neural parsers have been proposed that learn discourse structures through representation learning~\cite{rst_deepnn, kobayashi2022rst_deepnn1, yu2018rst_deepnn2, ji2014rst_representation}. Other approaches formulate discourse parsing as a text generation problem, using LSTMs~\cite{braud2016multi_rst_textgen} or encoder--decoder architectures~\cite{hu2023rst_textgen1, raffel2020t5} to generate discourse trees. More recent work ~\cite{maekawa2024rst_llm_gen, thompson2024llamipa, liu2021dmrst} fine-tunes pretrained models for RST discourse tree generation. Other studies further leverage predicted discourse trees as structural guidance for text generation~\cite{kim2025rst_llm_align, adewoyin2022rstgen}. Relative to this literature, our main contribution is demonstrating the value of RST-Parsing in slide generation for the first time. In addition, our approach to LLM-based RST-parsing differs from prior work in leveraging SOTA LLMs without finetuning and demonstrating their effectiveness. Using these RST trees as structural guidance, we enable the generation of highly coherent, narrative-driven content.

\subsection{Paper-Slide Datasets.} Prior work has introduced several paired datasets for paper-to-slide generation~\cite{sun2021d2s, ge2025autopresent, fu2022doc2ppt, zheng2025pptagent, liang2025slidegen} (see Supplement for a comparison table). These datasets share several critical limitations. First, the datasets in~\cite{ge2025autopresent, zheng2025pptagent} provide broad domain coverage and lack a specific academic specialization. Given the difficulty of effective human-level slide generation, there is a need for a well-scoped dataset exposing the specific generation challenges of technical academic content such as dense equations, figures, and structured arguments. Second, prior publicly-available datasets~\cite{sun2021d2s, fu2022doc2ppt, zheng2025pptagent, ge2025autopresent} were not curated for presentation quality, making it difficult to assess how effective slide-generation approaches are in achieving expert-level quality. Note that while the SlideGen~\cite{liang2025slidegen} dataset also contains high-quality technical presentations, it has not been publicly-released. In contrast to these works, ArcBench is the result of curating 100 oral high-quality pairs from 994 paper-presentation pairs at top CV/ML conferences, and we will release all of these assets. We further validate our human reference slides by comparing them against generated baselines (see Tab.~\ref{tab:pairwise_ap}), revealing a clear quality gap that motivates the benchmark.

\section{ArcDeck: Narrative-Driven Slide Generation}
\label{sec:method}

\subsection{Preprocessing}

Preprocessing stage takes the raw PDF of the paper as input and distills it into structured textual and visual assets required for downstream generation. We utilize Docling~\cite{livathinos2025docling} and Marker~\cite{paruchuri_marker_2023} to parse the paper, converting the text into a clean Markdown format while excluding the bibliography and appendix to focus on narrative content. Figures and tables are extracted into an asset dictionary, with each visual element stored alongside its caption and size information to provide layout context for the agents. In parallel, in-text citations are parsed to construct a mapping dictionary that links short-form citations to their full numbered references for slide generation.

\subsection{Narrative-Driven Outline Generation}
The Narrative-Driven Outline Generation stage translates the preprocessed text into a logically structured slide plan. To ensure the resulting slide is both structurally sound and thematically consistent, we orchestrate three core components that use the Markdown representation. First, the \textbf{Discourse Parser} \agentnum{1} constructs a hierarchical discourse tree to explicitly model the rhetorical dependencies between text segments. Second, the \textbf{Commitment Builder} \agentnum{2} establishes a Global Commitment to capture the overarching constraints and key messages of the target deck. Finally, guided by both these structural and thematic signals, the \textbf{Narrative Refinement Loop}---comprising the \textit{Slide Planner/Reviser} \agentnum{3}, \textit{Narrative Critic} \agentnum{4}, and \textit{Narrative Judge} \agentnum{5}---iteratively drafts, evaluates, and polishes the slide outline. Together, this multi-agent process ensures the generated outline preserves the source paper's narrative flow and logical interdependencies before visual generation.

\begin{figure}[t]
\centering
\includegraphics[width=\linewidth]{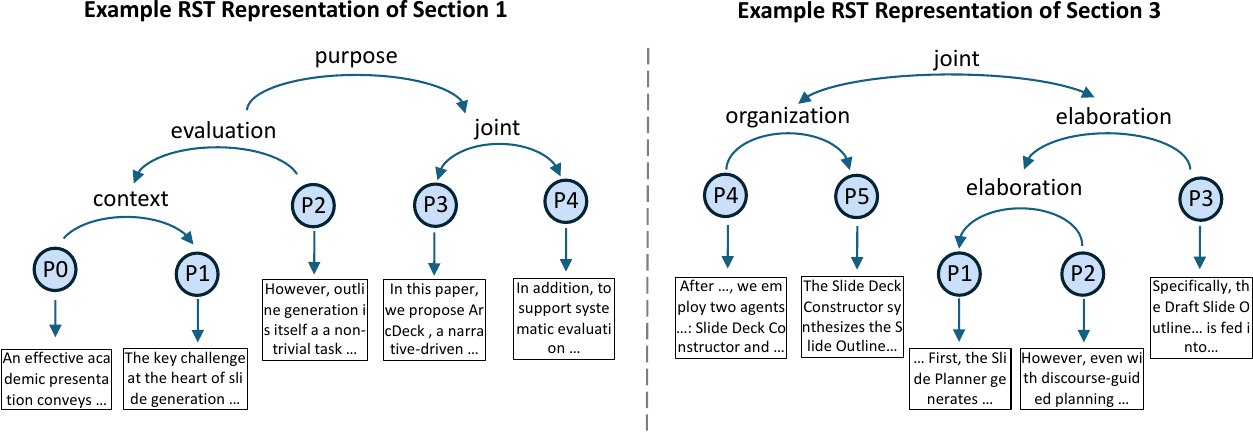}
\caption{\textbf{Example Discourse Trees from Secs. 1 \& 3 of this paper.} 
Leaf nodes correspond to elementary discourse units (EDUs, corresponding to paragraphs), while internal nodes and arrows denote rhetorical relations that recursively group these spans.} 
 \label{fig:rst}
\end{figure}

\paragraph{\textbf{Discourse Parser} \protect \agentnum{1}.} The Discourse Parser serves as a structural analysis agent that constructs a hierarchical RST discourse tree to explicitly model rhetorical dependencies between text segments, providing the guidance for content grouping and narrative ordering. Concretely, for each paper section, it generates a binary discourse tree by treating paragraphs as elementary discourse units (EDUs) that form the leaves. Inspired by the Rhetorical Structure Theory~\cite{mann1987rst}, the parser assigns relations between adjacent units as either (i) nucleus–satellite, where the nucleus carries the central claim and the satellite provides supporting detail, or (ii) multinuclear (symmetric), where both units are equally central (see Tab.~\ref{tab:rst_relations} for the full relation taxonomy). 
At higher levels, the tree recursively groups EDUs based on the relations identified at lower levels, producing a hierarchy that encodes both local paragraph-level dependencies and the global narrative organization of the section: relations closer to the root span larger portions of the text and capture high-level rhetorical structure, while relations near the leaves reflect fine-grained narrative progression within subsections (see Fig.~\ref{fig:rst} for an example). The discourse trees are then serialized into JSON and passed to the Slide Planner for subsequent outline generation.

\begin{table}[t!]
\centering
\small
\caption{\textbf{RST Relation Taxonomy.} Definitions of the rhetorical relations used to construct our discourse trees (NS: Nucleus-Satellite; MN: Multinuclear).}
\label{tab:rst_relations}

\setlength{\tabcolsep}{5pt}
\resizebox{0.95\linewidth}{!}{
\begin{tabular}{@{}llc@{}}
\toprule
\textbf{Relation} & \textbf{Explanation} & \textbf{Type} \\
\midrule
Elaboration  & Adds detail, examples, or implementation specifics.          & NS \\
Explanation  & Clarifies why or how a nucleus claim holds.                  & NS \\
Context      & Background or definitions needed to interpret the nucleus.   & NS \\
Purpose      & Goal or motivation associated with the nucleus.              & NS \\
Evaluation   & Assessment, strength, or limitation of the nucleus.          & NS \\
Organization & Roadmap or meta-structural framing.                          & NS \\
\midrule
Joint        & Parallel units at the same rhetorical level, equally central. & MN \\
Same-unit    & Two EDUs forming one semantic unit split across boundaries.   & MN \\
\bottomrule
\end{tabular}
}
\end{table}

\begin{figure}[t!]
 \centering
 \includegraphics[width=\linewidth]{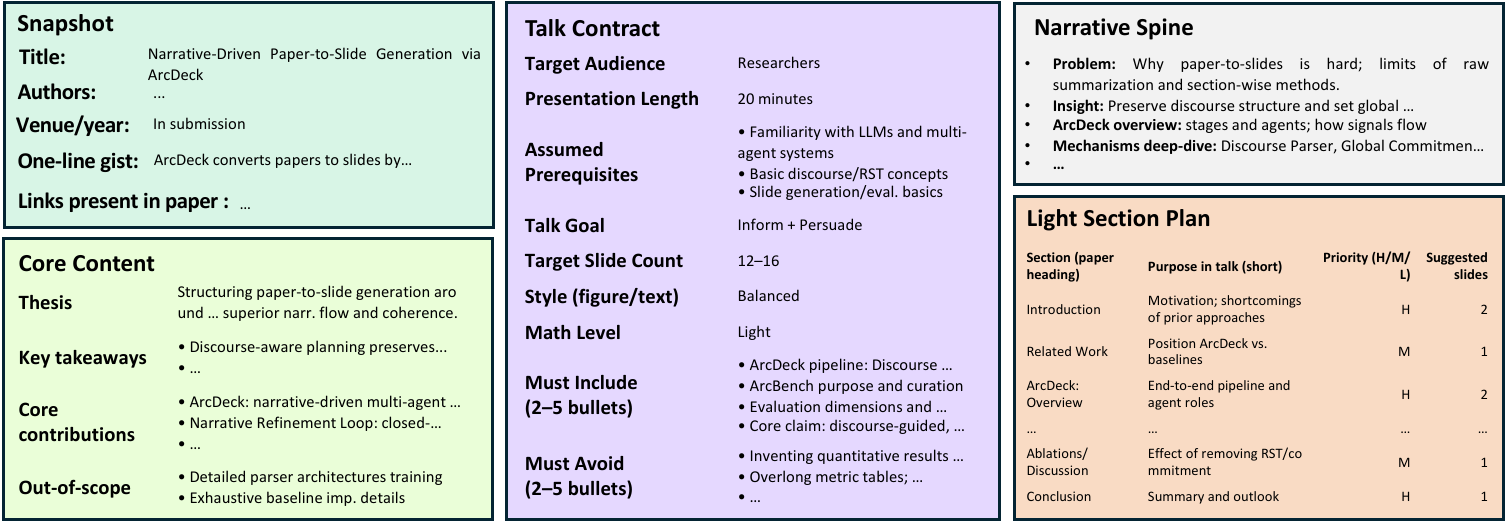}
 \caption{\textbf{Global Commitment Example.} The global commitment establishes the high-level narrative progression of the presentation through five key components: a snapshot, core content, a talk contract, a narrative spine, and a light section plan.}
 \label{fig:global_commit}
\end{figure}

\paragraph{\textbf{Commitment Builder} \protect \agentnum{2}.} The Commitment Builder takes the Markdown data and user-provided `target audience' along with `presentation duration' as input, and generates a global commitment to establish an overall plan for slide generation. The generated commitment consists of key information describing the structure of the slide deck: a snapshot, core content that specifies the thesis and key takeaways, a talk contract that specifies the assumed prerequisites, a narrative spine, and an overall section plan (see~Fig.~\ref{fig:global_commit}). The global commitment serves as high-level guidance to the following Narrative Refinement Loop.

\paragraph{\textbf{Narrative Refinement Loop \protect \agentnum{3}, \protect \agentnum{4}, \protect \agentnum{5}.}} 

Once the Discourse Tree and Global Commitment are generated, the Narrative Refinement Loop iteratively refines the Markdown information to create a Slide Outline. First, the Slide Planner generates an initial slide outline guided by the structural signals from the Discourse Tree and the Global Commitment. For each section, it groups preprocessed paragraphs according to their rhetorical relations to determine which content should appear together, forming a logically structured narrative. The resulting Draft Slide Outline in JSON contains a section for each slide including a deck title, the grouped paragraphs for its content, and a rationale for the structural grouping decisions. However, even with discourse-guided planning, one-shot planning may fail to ensure alignment with the Global Commitment and a coherent narrative flow. We therefore introduce an iterative critique–judge–revise cycle to refine the outline. Within this loop, we assess the outline with the following criteria: (a) alignment with the Global Commitment, (b) global narrative flow, (c) section balance, (d) slide-level coherence, and (e) redundancy or missing content.

Specifically, the Draft Slide Outline generated by the Slide Planner is fed into the Narrative Critic together with the Global Commitment to obtain feedback, which is then passed to the Narrative Judge to determine whether the outline is ready or requires revision. The Judge then issues a decision, summarizes the rationale, identifies must-fix issues with severity levels (high, medium, low), and provides actionable guidance for the Reviser when necessary. If a revision is required, the Reviser updates the slide outline based on the combined feedback, after which it re-enters the evaluation cycle. This refinement process continues until the outline is \textit{ready}, as determined by the Narrative Judge, or three refinement cycles have been completed, at which point the finalized outline is forwarded to the Slide Generation stage in JSON format.

\subsection{Slide Generation}
\begin{figure}[t]
    \centering
    \includegraphics[width=\linewidth]{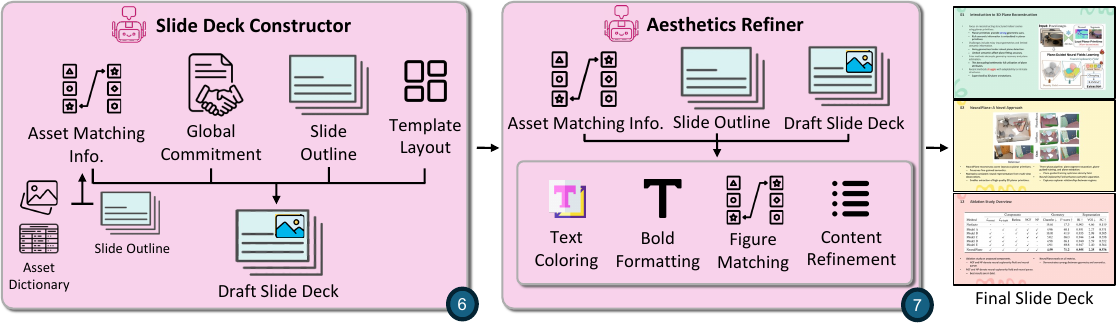}
    \caption{\textbf{Slide Generation Stage.} Slide Deck Constructor \protect \agentnum{6} handles asset mapping, layout selection, and text generation, while the Aesthetics Refiner \protect \agentnum{7} polishes the deck via figure matching and content formatting.}
    \label{fig:slide_generation}
\end{figure}

After preparing the narrative-driven Slide Outline, we employ two agents to generate an editable \texttt{.pptx} slide (Fig.~\ref{fig:slide_generation}): the \textbf{Slide Deck Constructor} \protect\agentnum{6} selects relevant figures and tables from the dictionary and assigns an appropriate template layout based on the visual elements, producing a Draft Slide Deck conditioned on the Global Commitment, and the \textbf{Aesthetics Refiner} \protect\agentnum{7} refines the draft through content adjustment (\eg, enriching sparse text), additional figure matching, and stylistic text formatting.

\paragraph{\textbf{Slide Deck Constructor \protect \agentnum{6}.}} This agent synthesizes the Slide Outline, preprocessed visual assets, and the Global Commitment to assemble a Draft Slide Deck. For each slide, it first selects the most relevant figures and tables based on their extracted captions. It then assigns one of 14 available layout templates, conditioned on the text volume and the number of visual elements, its size and aspect ratio. Guided by the RST-derived outline and Global Commitment, the agent generates the slide text while explicitly highlighting key topics to preserve the intended narrative flow. Furthermore, it adapts the textual organization, toggling between bulleted hierarchies and paragraph narration based on information density, and notes down the short-form citation mentioned in the text, which will be later added as footnotes. The resulting draft deck is output as a structured JSON object containing global metadata alongside slide-wise titles, content, matched visuals, and references.

\paragraph{\textbf{Aesthetics Refiner \protect \agentnum{7}.}} The Aesthetics Refiner utilizes the Asset Matching Info., Slide Outline, and Draft Slide Deck to perform the final polish of the presentation. It executes four targeted operations: (i) \textbf{Figure Matching} conducts an additional review to insert relevant visual elements into slides lacking sufficient visual grounding; (ii) \textbf{Content Refinement} adjusts textual density by enriching sparse slides, condensing heavy text, or merging underutilized layouts to maintain a clear narrative pacing; (iii) \textbf{Text Coloring} dynamically applies a consistent theme color derived from the most frequently occurring colors in the deck's figures; and (iv) \textbf{Bold Formatting} adds targeted emphasis to key terminology to enhance overall readability. The resulting refined JSON is then rendered into the Final Slide Deck using the \texttt{python-pptx} library.

\def\mypar#1{\vspace{.1cm} \textbf{#1}}
\subsection{ArcBench Dataset}
\label{sec:data}
To establish a rigorous evaluation protocol for the paper-to-slide generation task, we introduce \benchname, a newly curated paper-slide paired dataset. We extract source papers and their corresponding presentation materials directly from official venues, resulting in an initial 994 candidate pairs from six major computer vision and machine learning conferences (ICCV, CVPR, ECCV, ICML, ICLR, and NeurIPS) between 2022 and 2025. Because large-scale evaluation across multiple metrics and baselines is costly, running the full suite on all 994 pairs is prohibitive. Accordingly, we apply a post-processing step to curate a high-quality subset of 100 pairs for all subsequent experiments (Fig.~\ref{fig:arcbench_stats}).

To balance evaluation costs with the topical diversity, we select this 100-pair dataset through three strict filtering stages: we restrict the selection to \textit{oral presentations} to ensure deliberate narrative design, and we require at least 3 figures and 3 tables per article to guarantee sufficient visual and quantitative material for evaluating content fidelity. A key novelty of \benchname\ is its explicit reliance on ``author-prepared'' (AP) presentations, providing a robust human reference standard to accurately measure how closely generated outputs approximate human-level preparation. While this initial dataset is limited to AI venues, we leave the expansion to other scientific disciplines for future work, and we release both the curated 100-pair dataset and the full 994-pair pool to facilitate further research. Full details are available in Appendix Section~\ref{appendix:dataset}.

\begin{figure}[t]
    \centering
    \includegraphics[width=\linewidth]{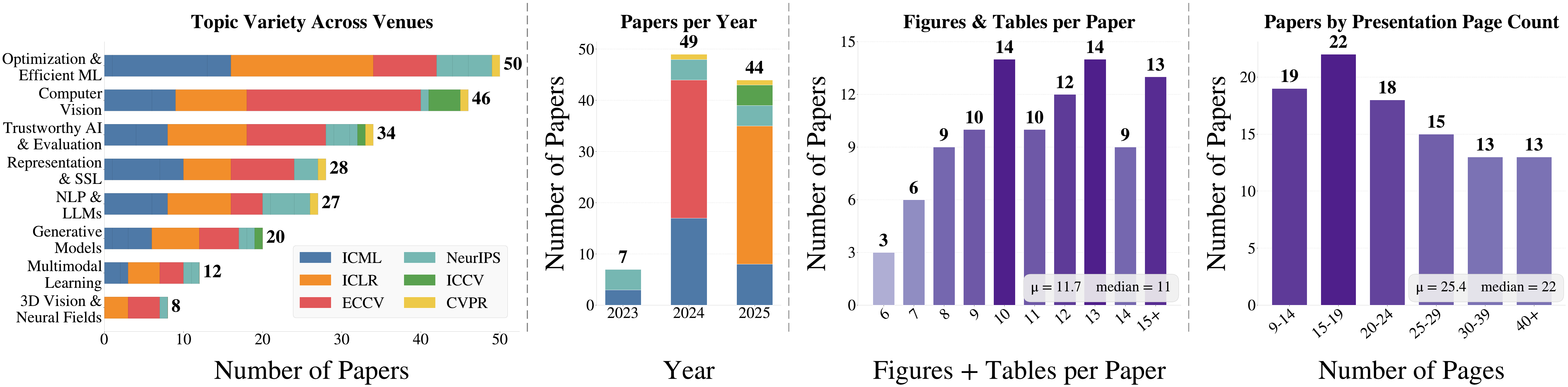}
    \caption{\textbf{ArcBench Statistics.} Overview of our curated 100-pair dataset, showing the distribution of research topics across six major CV \& ML venues, publication years, visual density, and author-prepared presentation lengths.}
    \label{fig:arcbench_stats}
\end{figure}
\def\mypar#1{\vspace{.1cm} \textbf{#1}}

\section{Experiments}
\label{sec:exp}

\subsection{Experimental Setup}
\label{sec:exp_setup}

\paragraph{Baselines \& Benchmark.} 
We compare against four systems spanning three paradigms. \textbf{(i)}~\emph{Prompt-based generation:} \textbf{HTML} produces slides via a single LLM call that generates HTML+CSS, directly converted to PDF.
\textbf{(ii)}~\emph{Multi-agent pipelines:} \textbf{Paper2Poster}~\cite{pang2025paper2poster} (P2Poster), a visual-in-the-loop multi-agent system adapted for slide generation; \textbf{PPTAgent}~\cite{zheng2025pptagent}, a two-stage edit-based slide generation approach; \textbf{SlideGen}~\cite{liang2025slidegen}, a six-agent VLM framework with template-driven layout composition; and \textbf{(iii)}~\emph{Human reference:} the original \textit{author-prepared} slides. To separate pipeline design from LLM capability, we run \methodname\ and all baselines with three generation backbones: GPT-4o~\cite{gpt4o}, GPT-5~\cite{singh2025openai_gpt5}, and Qwen3-VL-32B~\cite{Qwen3-VL}. All methods are evaluated on the 100 paper--slide pairs in \benchname, ensuring identical source material and evaluation protocol.

\paragraph{Slide Generation Format.}
All generated slides follow a 13.33$\times$7.5-inch layout. Our method uses a reference template with 14 distinct layout configurations covering common scenarios with multiple images and tables. For fairness, all methods are evaluated under fixed slide themes so that preference decisions reflect content and structure rather than stylistic variation.

\subsection{Evaluation Setup}
We evaluate \methodname\ along four axes: (i)\emph{content coverage} (preservation of key information from the source paper), (ii)\emph{narrative coherence} (logical slide progression), (iii)\emph{textual quality} (fluency and technical specificity), and (iv)\emph{visual quality} (clarity of layout and visual elements). Metrics are summarized in Tab.~\ref{tab:evaluation_metrics}.

\begin{table*}[t!]
\centering
\caption{\textbf{Summary of Evaluation Metrics}}
\label{tab:evaluation_metrics}

\renewcommand{\arraystretch}{1.4}
\resizebox{\textwidth}{!}{
\begin{tabular}{>{\columncolor{catblue}}l l l >{\centering\arraybackslash}p{2.5cm} l}
\toprule
\rowcolor{headerblue}
\textcolor{headertextwhite}{\textbf{Category}} & 
\textcolor{headertextwhite}{\textbf{Specific Metric}} & 
\textcolor{headertextwhite}{\textbf{Input}} & 
\textcolor{headertextwhite}{\textbf{Output}} & 
\textcolor{headertextwhite}{\textbf{High-Level Idea}} \\
\midrule
\rowcolor{row1}
\cellcolor{catblue} & Story & Pres.\ Text \& 25 MCQs & \cellcolor{row1} & Storyline capture and high-level information retention. \\
\rowcolor{row2}
\cellcolor{catblue} & Visuals & Slide Images \& 25 MCQs & \cellcolor{row2} & Information conveyed through visual elements and slide layout. \\
\rowcolor{row1}
\cellcolor{catblue} & Hard & Pres.\ Text \& 25 MCQs & \cellcolor{row1} & Conceptual understanding of methodology and limitations. \\
\rowcolor{row2}
\multirow{-4}{*}{\cellcolor{catblue}\shortstack[l]{\textit{VLM-based} \\ \textit{Q/A Quiz}}} & Depth & Pres.\ Text \& 25 MCQs & \multirow{-4}{*}{\centering\shortstack{Accuracy\\(\%)}} & Coverage of important technical/implementation details. \\
\midrule
\rowcolor{row1}
\cellcolor{catblue} & Text Quality (TQ) & \cellcolor{row1} & \cellcolor{row1} & Technical substance and specificity of the text. \\
\rowcolor{row2}
\cellcolor{catblue} & Narrative Flow (NF) & \multirow{-2}{*}{\centering Pres.\ Text} & \cellcolor{row2} & Logical progression and comprehensive section coverage. \\
\rowcolor{row1}
\cellcolor{catblue} & Visual Layout (VL) & \cellcolor{row1} & \cellcolor{row1} & Spatial geometry, balance, and arrangement of elements. \\ 
\rowcolor{row2}
\multirow{-4}{*}{\cellcolor{catblue}\shortstack[l]{\textit{VLM-as-} \\ \textit{Judge}}} & Visual Thematic (VT) & \multirow{-2}{*}{\centering Slide Images} & \multirow{-4}{*}{\centering\shortstack{1--100\\Score}} & Thematic consistency, design coherence, and visual elements. \\
\midrule
\rowcolor{row1}
\cellcolor{catblue} & ROUGE-L & Pres.\ Text \& Source Paper & \centering\shortstack{Overlap\\Score} & Textual sequence overlap and content coverage. \\
\rowcolor{row2}
\multirow{-2}{*}{\cellcolor{catblue}\shortstack[l]{\textit{Automated} \\ \textit{Metrics}}} & Perplexity & Pres.\ Text & \centering\shortstack{Perplexity\\Score} & Linguistic fluency and text coherence. \\
\midrule
\rowcolor{row1}
\cellcolor{catblue} & Narrative Flow & Two \ Decks \& Source Paper & \cellcolor{row1} & Opening quality, narrative arc, and progressive revelation. \\
\rowcolor{row2}
\multirow{-2}{*}{\cellcolor{catblue}\shortstack[l]{\textit{VLM Pairwise} \\ \textit{Preference}}} & Author-Prepared & Generated\ Deck \& AP Deck & \multirow{-2}{*}{\centering\shortstack{Win Rate\\(\%)}} & Overall presentation quality relative to human-authored slides. \\
\bottomrule
\end{tabular}
}
\end{table*}

\paragraph{VLM-based Q/A Quiz.}
To measure how much of the source paper's content is preserved in the generated slides, we design a two-stage quiz protocol. First, a VLM reads the source paper and generates 25 multiple-choice questions across four categories: \emph{Story} (narrative arc), \emph{Visuals} (information from figures and diagrams), \emph{Hard} (deep conceptual understanding of methodology), and \emph{Depth} (fine-grained technical details). Second, a separate VLM attempts to answer these questions using only the generated slides as input, with text-based categories answered from extracted slide text and the Visuals category answered from slide images. The resulting accuracy, reported in Tab.~\ref{tab:quantitative_eval}, reflects how effectively the slides communicate the paper's content at each level of granularity.

\paragraph{VLM-as-Judge.}
Rather than relying on generic metrics, we employ VLM judges with carefully designed rubrics to score each presentation on a 1-100 scale across four dimensions: \emph{Text Quality} (technical substance and specificity), \emph{Narrative Flow} (logical progression and section coverage), \emph{Visual Layout} (spatial arrangement and readability), and \emph{Visual Thematic} (design coherence and appropriate use of visual elements). Each dimension uses a 10-item checklist, scored by the number of satisfied criteria. Results are reported in Tab.~\ref{tab:quantitative_eval}.

\paragraph{Automated Metrics.} 
We complement the VLM-based evaluations with two standard text metrics reported in Tab.~\ref{tab:automated_metrics}: \emph{ROUGE-L}, measuring sequence overlap between the generated slide text and the source paper to quantify content coverage, and \emph{Perplexity} (computed via LLaMA-3-8B~\cite{grattafiori2024llama}), assessing linguistic fluency of the slide text.

\paragraph{VLM Pairwise Preference (A/B Test).}
We use two pairwise comparison protocols. For \emph{baseline vs.\ ours}, a VLM judge receives two slide decks and the source paper, and selects which slide better preserves the source paper's logical structure and narrative arc, as reported in Tab.~\ref{tab:pairwise_ab}. For \emph{baseline vs.\ AP}, the judge instead evaluates holistic presentation quality, jointly considering content organization, visual design, information delivery, and slide structure, to assess how closely the generated slides approximate author-prepared presentations, as reported in Tab.~\ref{tab:pairwise_ap}. We use both closed- (GPT-5) and open-source (Qwen3-VL) judges to mitigate single-model bias. Quiz and A/B evaluations are repeated 11 times with randomized order, using average scores or majority vote.

\begin{table*}[t!]
  \caption{\textbf{VLM-Based Evaluation.} 
  The best value in each generation model is highlighted in \textcolor{winGreen}{green}, and runner-up is underlined.}
  \label{tab:quantitative_eval}
  \centering
  \setlength{\tabcolsep}{3pt}
  \renewcommand{\arraystretch}{1.05}
  \resizebox{\linewidth}{!}{
    \begin{tabular}{
      l
      cccc
      @{\hspace{4pt}}
      cccc
      @{\hspace{4pt}}
      cccc
      @{\hspace{4pt}}
      cccc
    }
    \toprule
    & \multicolumn{4}{c}{\textbf{VLM-based Q/A Quiz (Open) $\uparrow$}}
    & \multicolumn{4}{c}{\textbf{VLM-based Q/A Quiz (Closed) $\uparrow$}}
    & \multicolumn{4}{c}{\textbf{VLM-as-Judge (Open) $\uparrow$}}
    & \multicolumn{4}{c}{\textbf{VLM-as-Judge (Closed) $\uparrow$}} \\
    \cmidrule(lr){2-5}
    \cmidrule(lr){6-9}
    \cmidrule(lr){10-13}
    \cmidrule(lr){14-17}
    \textbf{Method}
    & \textbf{Story} & \textbf{Visuals} & \textbf{Hard} & \textbf{Depth}
    & \textbf{Story} & \textbf{Visuals} & \textbf{Hard} & \textbf{Depth}
    & \textbf{TQ} & \textbf{NF} & \textbf{VL} & \textbf{VT}
    & \textbf{TQ} & \textbf{NF} & \textbf{VL} & \textbf{VT} \\
    \midrule

    \multicolumn{17}{c}{\cellcolor{RoyalBlue!8}\textbf{Generation Model: GPT-5}} \\

    HTML
    & 95.16 & 64.13 & \textcolor{winGreen}{81.83} & \underline{69.73}
    & 95.60 & \underline{38.40} & 80.00 & 82.20
    & \textcolor{winGreen}{86.11} & \underline{89.06} & 95.22 & 76.77
    & \textcolor{winGreen}{58.50} & \underline{61.50} & 63.67 & 10.00 \\
    P2Poster
    & 95.22 & 65.51 & 76.70 & 67.67
    & 87.00 & 27.00 & 66.40 & 64.60
    & 62.20 & 70.55 & 97.08 & 91.35
    & 35.17 & 41.67 & 72.00 & 78.50 \\
    PPTAgent
    & 92.96 & 65.08 & 69.49 & 65.67
    & 75.00 & 12.80 & 32.00 & 36.20
    & 35.95 & 55.94 & 84.86 & 81.95
    & 16.50 & 37.67 & 71.33 & 57.50 \\
    SlideGen
    & \underline{95.26} & \textcolor{winGreen}{68.37} & 78.39 & 69.43
    & \textcolor{winGreen}{97.60} & 35.80 & \textcolor{winGreen}{86.00} & \underline{82.40}
    & 69.02 & 88.18 & \underline{97.36} & \underline{95.28}
    & 36.17 & 56.67 & \textcolor{winGreen}{88.33} & \textcolor{winGreen}{84.83} \\
    \rowcolor{gray!15} \methodname
    & \textcolor{winGreen}{95.41} & \underline{65.95} & \underline{81.13} & \textcolor{winGreen}{71.58}
    & \underline{96.00} & \textcolor{winGreen}{41.00} & \underline{85.60} & \textcolor{winGreen}{83.40}
    & \underline{78.38} & \textcolor{winGreen}{91.39} & \textcolor{winGreen}{99.40} & \textcolor{winGreen}{96.06}
    & \underline{46.50} & \textcolor{winGreen}{63.83} & \underline{85.50} & \underline{83.67} \\

    \addlinespace[2pt]
    \multicolumn{17}{c}{\cellcolor{ForestGreen!8}\textbf{Generation Model: GPT-4o}} \\

    HTML
    & 93.34 & 57.36 & 64.57 & 60.57
    & 51.36 & 6.64 & 11.84 & 14.16
    & 16.72 & 19.51 & 54.47 & 12.89
    & 7.25 & 14.00 & 24.00 & 0.25 \\
    P2Poster
    & 94.12 & \underline{63.11} & 66.07 & 60.18
    & \underline{60.80} & \underline{8.16} & 17.60 & 20.24
    & \underline{32.32} & 40.71 & 92.05 & 86.23
    & 11.12 & 28.12 & 60.25 & 62.38 \\
    PPTAgent
    & 93.35 & 61.90 & 66.01 & 60.84
    & 59.04 & 7.52 & 15.12 & 18.56
    & 16.43 & 25.75 & 82.59 & 71.49
    & \underline{12.62} & 26.75 & 59.12 & 34.75 \\
    SlideGen
    & \underline{93.75} & 62.79 & \underline{66.15} & \underline{61.00}
    & 60.00 & 8.08 & \underline{17.84} & \underline{21.36}
    & 19.28 & \underline{41.40} & \underline{93.30} & \underline{89.40}
    & 9.62 & \underline{30.75} & \underline{84.75} & \underline{78.25} \\
    \rowcolor{gray!15} \methodname
    & \textcolor{winGreen}{95.46} & \textcolor{winGreen}{64.57} & \textcolor{winGreen}{72.63} & \textcolor{winGreen}{67.56}
    & \textcolor{winGreen}{82.40} & \textcolor{winGreen}{18.16} & \textcolor{winGreen}{40.08} & \textcolor{winGreen}{47.20}
    & \textcolor{winGreen}{51.68} & \textcolor{winGreen}{62.78} & \textcolor{winGreen}{96.05} & \textcolor{winGreen}{92.94}
    & \textcolor{winGreen}{24.38} & \textcolor{winGreen}{43.25} & \textcolor{winGreen}{91.00} & \textcolor{winGreen}{81.25} \\

    \addlinespace[2pt]
    \multicolumn{17}{c}{\cellcolor{Plum!8}\textbf{Generation Model: Qwen-3-VL-32B-Instruct}} \\

    HTML
    & 94.07 & 61.53 & 75.24 & 66.68
    & 85.67 & \textcolor{winGreen}{36.65} & 63.26 & 61.77
    & 67.55 & 71.02 & 82.05 & 66.74
    & \underline{48.62} & 51.88 & 64.00 & 24.38 \\
    P2Poster
    & 94.28 & \underline{63.36} & 75.96 & 66.75
    & 86.14 & 21.49 & 51.26 & 54.51
    & 65.96 & 74.24 & \underline{95.90} & 89.60
    & 36.50 & 49.50 & 72.25 & \underline{75.88} \\
    PPTAgent
    & 92.66 & 61.94 & 69.53 & 65.38
    & 67.91 & 12.00 & 22.14 & 25.12
    & 41.55 & 66.12 & 80.44 & 45.25
    & 19.25 & 35.25 & 47.25 & 11.25 \\
    SlideGen
    & \underline{94.81} & 62.12 & \underline{76.57} & \underline{67.96}
    & \underline{90.42} & 28.47 & \underline{64.93} & \underline{63.07}
    & \underline{69.18} & \underline{86.40} & 94.16 & \underline{92.18}
    & 42.50 & \underline{55.25} & \underline{85.25} & \textcolor{winGreen}{79.50} \\
    \rowcolor{gray!15} \methodname
    & \textcolor{winGreen}{95.69} & \textcolor{winGreen}{66.58} & \textcolor{winGreen}{83.22} & \textcolor{winGreen}{71.39}
    & \textcolor{winGreen}{93.12} & \underline{34.79} & \textcolor{winGreen}{68.28} & \textcolor{winGreen}{66.70}
    & \textcolor{winGreen}{80.06} & \textcolor{winGreen}{89.00} & \textcolor{winGreen}{98.31} & \textcolor{winGreen}{94.68}
    & \textcolor{winGreen}{53.62} & \textcolor{winGreen}{63.00} & \textcolor{winGreen}{87.25} & \textcolor{winGreen}{79.50} \\
    \bottomrule
    \end{tabular}
  }\vspace{1em}
\end{table*}

\begin{table*}[t!]
\centering
\caption{\textbf{VLM Pairwise Preference (A/B Test).} Holistic evaluation of generated presentations using two comparison protocols: assessing narrative flow against baselines (left), and overall quality against Author-Prepared (AP) slides (right). To mitigate single-model bias, both closed- (GPT-5) and open-source (Qwen3-VL) judges are used.}
\label{tab:pairwise_eval}

\begin{subtable}[t]{0.49\linewidth}
\centering
\vspace{0.8em}
\caption{\textbf{Preference for Narrative Flow.} Win rate (\%) of \textbf{Ours} vs.\ baselines. \textcolor{winGreen}{Green} indicates our method is preferred ($>50\%$), \textcolor{lossRed}{red} otherwise.}
\label{tab:pairwise_ab}
\small
\setlength{\tabcolsep}{2pt} 
\renewcommand{\arraystretch}{1.08}
\resizebox{\linewidth}{!}{ 
\begin{tabular}{@{} p{2.8cm} *{6}{S[table-format=3.1]} @{}}
\toprule
\textit{Generation Model}
& \multicolumn{2}{c}{\textbf{GPT-5}}
& \multicolumn{2}{c}{\textbf{GPT-4o}}
& \multicolumn{2}{c}{\textbf{Qwen3}} \\
\cmidrule(lr){2-3} \cmidrule(lr){4-5} \cmidrule(lr){6-7}
\textit{Evaluation Model}
& {Qwen3} & {GPT-5}
& {Qwen3} & {GPT-5}
& {Qwen3} & {GPT-5} \\
\midrule
Ours vs.\ P2Poster
& {\textcolor{winGreen}{62.4}} & {\textcolor{winGreen}{97.6}}
& {\textcolor{winGreen}{78.0}} & {\textcolor{winGreen}{84.0}}
& {\textcolor{winGreen}{92.8}} & {\textcolor{winGreen}{90.7}} \\
Ours vs.\ PPTAgent
& {\textcolor{winGreen}{100.0}} & {\textcolor{winGreen}{100.0}}
& {\textcolor{winGreen}{99.0}} & {\textcolor{winGreen}{95.0}}
& {\textcolor{winGreen}{100.0}} & {\textcolor{winGreen}{97.9}} \\
Ours vs.\ SlideGen
& {\textcolor{winGreen}{55.3}} & {\textcolor{winGreen}{50.6}}
& {\textcolor{winGreen}{81.0}} & {\textcolor{winGreen}{68.0}}
& {\textcolor{winGreen}{57.7}} & {\textcolor{lossRed}{48.5}} \\
Ours vs.\ HTML
& {\textcolor{winGreen}{82.4}} & {\textcolor{winGreen}{84.7}}
& {\textcolor{winGreen}{79.0}} & {\textcolor{winGreen}{92.0}}
& {\textcolor{winGreen}{76.3}} & {\textcolor{winGreen}{84.5}} \\
\bottomrule
\end{tabular}
}
\end{subtable}%
\hfill 
\begin{subtable}[t]{0.49\linewidth}
\centering
\caption{\textbf{Preference vs.\ Author-Prepared (AP).} Higher values indicate closer quality to AP slides. \colorbox{bestGreen!30}{Green} highlights the best per column.}
\label{tab:pairwise_ap}
\small
\setlength{\tabcolsep}{2pt} 
\renewcommand{\arraystretch}{1.08}
\resizebox{\linewidth}{!}{ 
\begin{tabular}{@{} p{2.8cm} *{6}{S[table-format=3.1]} @{}}
\toprule
\textit{Generation Model}
& \multicolumn{2}{c}{\textbf{GPT-5}}
& \multicolumn{2}{c}{\textbf{GPT-4o}}
& \multicolumn{2}{c}{\textbf{Qwen3}} \\
\cmidrule(lr){2-3} \cmidrule(lr){4-5} \cmidrule(lr){6-7}
\textit{Evaluation Model}
& {Qwen3} & {GPT-5}
& {Qwen3} & {GPT-5}
& {Qwen3} & {GPT-5} \\
\midrule
P2Poster vs.\ AP
& 35.1 & 33.3
& 27.5 & 26.7
& 37.7 & 40.0 \\
PPTAgent vs.\ AP
& 45.6 & 40.0
& 35.4 & 30.0
& 45.4 & 36.7 \\
SlideGen vs.\ AP
& \cellcolor{bestGreen!30}\textbf{46.8} & 45.8
& 32.1 & 33.3
& 46.0 & 45.8 \\
HTML vs.\ AP
& 35.9 & 33.3
& 31.3 & 33.3
& 42.6 & 40.0 \\
\midrule
\rowcolor{gray!10}
\textbf{Ours vs.\ AP}
& 35.7 & \cellcolor{bestGreen!30}\textbf{48.1}
& \cellcolor{bestGreen!30}\textbf{36.6} & \cellcolor{bestGreen!30}\textbf{40.0}
& \cellcolor{bestGreen!30}\textbf{47.3} & \cellcolor{bestGreen!30}\textbf{46.7} \\
\bottomrule
\end{tabular}
}
\end{subtable}

\end{table*}


\subsection{Analysis and Discussion}

\begin{wraptable}{r}{0.5\linewidth} 
\caption{\textbf{Automated Metrics Evaluation.}
The best value in each column is highlighted in \textcolor{winGreen}{green}, and runner-up is underlined. R-L values are scaled by $10^{-3}$.}
\label{tab:automated_metrics}
\centering
\setlength{\tabcolsep}{3.5pt} 
\renewcommand{\arraystretch}{1.05}
\resizebox{\linewidth}{!}{ 
\begin{tabular}{l cc cc cc}
\toprule
\textit{Generation Model}
 & \multicolumn{2}{c}{\textbf{GPT-5}}
 & \multicolumn{2}{c}{\textbf{GPT-4o}}
 & \multicolumn{2}{c}{\textbf{Qwen3}} \\
\cmidrule(lr){2-3} \cmidrule(lr){4-5} \cmidrule(lr){6-7}
\textit{Automated Metrics} 
& R-L $\uparrow$ & PPL $\downarrow$
& R-L $\uparrow$ & PPL $\downarrow$
& R-L $\uparrow$ & PPL $\downarrow$ \\
\midrule

HTML
& \underline{64.7} & 39.90
& 33.6 & \underline{31.51}
& \underline{108.5} & 31.08 \\

Paper2Poster
& \textcolor{winGreen}{71.3} & \textcolor{winGreen}{21.04}
& \underline{53.8} & \textcolor{winGreen}{22.07}
& 76.8 & \underline{16.96} \\

PPTAgent
& 33.6 & 91.43
& 34.6 & 57.21
& 64.6 & 23.33 \\

SlideGen
& 55.1 & 80.46
& 45.2 & 43.84
& 83.9 & 23.09 \\ \midrule

\rowcolor{gray!10}
\methodname
& 57.1 & \underline{46.01}
& \textcolor{winGreen}{84.8} & 33.47
& \textcolor{winGreen}{156.0} & \textcolor{winGreen}{16.95} \\

\bottomrule
\end{tabular}
}
\vspace{-1em} 
\end{wraptable}

\paragraph{Narrative Flow.}
A central design goal of \methodname\ is to produce presentations with coherent, human-like narrative progression. 
As shown in Tab.~\ref{tab:pairwise_ab}, \methodname\ consistently outperforms all baselines in pairwise comparisons, achieving the highest overall win rate across both open- and closed-source judges. The advantage is most pronounced against PPTAgent and HTML, whose outlines largely mirror source document headings without rhetorical restructuring. SlideGen presents a narrower gap, yet our method still maintains a consistent edge, which we attribute to the discourse-aware outline generation and the Narrative Refinement Loop.
These findings are corroborated by the VLM-as-Judge NF scores in Tab.~\ref{tab:quantitative_eval}: across all generation backbones and judge models, \methodname\ achieves the highest or second-highest NF scores.

\paragraph{Coverage of Content.}
We evaluate content coverage through quiz-based metrics as summarized in Tab.~\ref{tab:quantitative_eval}. \methodname\ achieves the highest or second-highest scores across all categories. On \textbf{Story}, our method leads across all three models, indicating that the discourse-driven outline better preserves the paper’s high-level narrative arc. 
The largest gains appear on \textbf{Hard} and \textbf{Depth}, which reflect conceptual understanding and fine-grained detail. Under Qwen-3-VL, \methodname\ exceeds the next-best baseline by 6.65 on Hard and 3.43 on Depth, with similar margins under GPT-4o.
We attribute this to the hierarchical discourse tree, which preserves elaboration and explanation relations and retains methodological nuances during condensation. 
On \textbf{Visuals}, which tests whether visually grounded questions can be answered from the slides alone, \methodname\ achieves the highest scores under GPT-4o and Qwen-3-VL and remains competitive under GPT-5, indicating strong visual coverage.

\begin{figure*}[t!]
    \centering
    \includegraphics[width=0.99\linewidth]{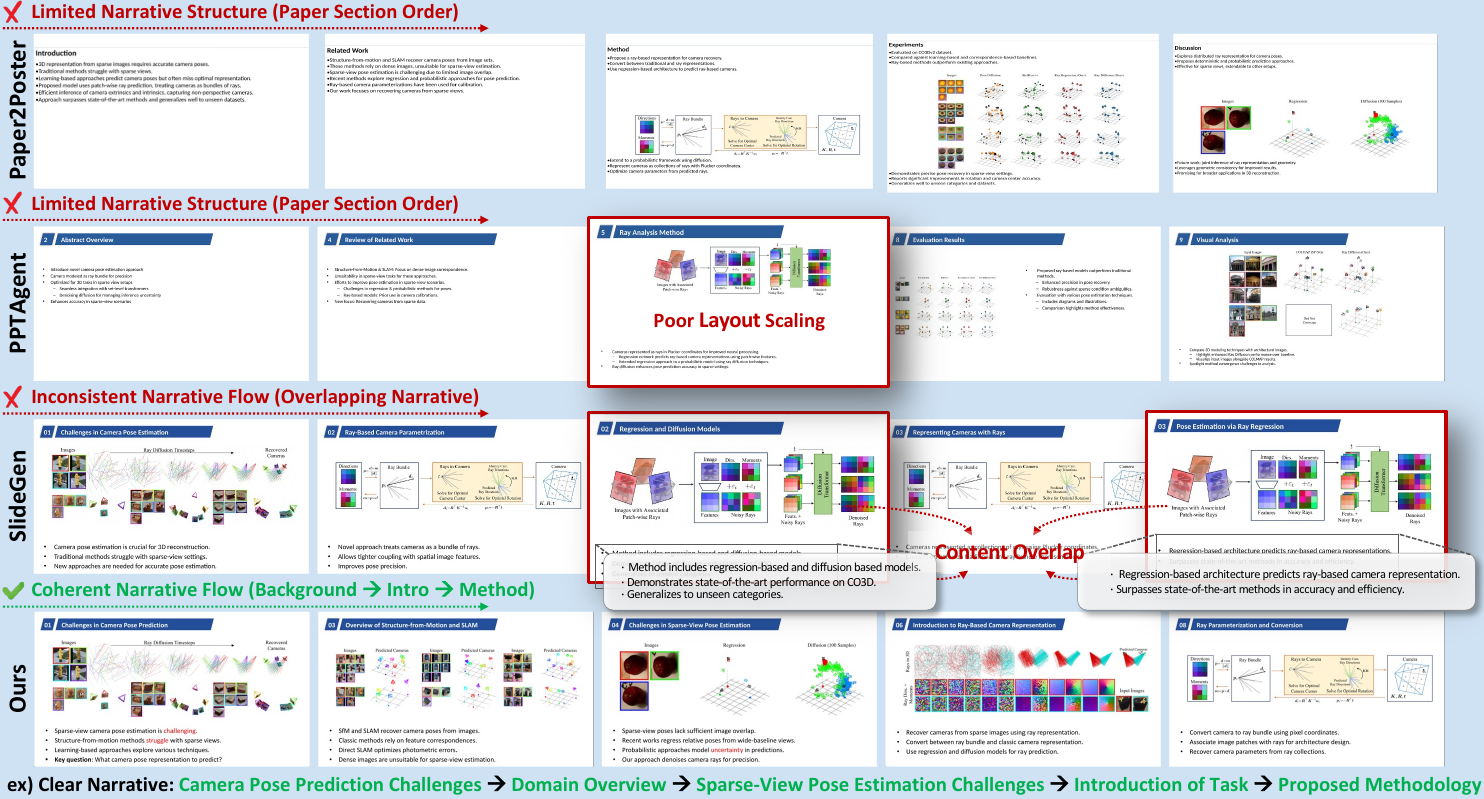}
    \caption{\textbf{Qualitative Comparison.} Unlike baselines, \textit{ArcDeck} shows clear narrative flow instead of mirroring paper section structure and avoids cross-slide content overlap.}
    \label{fig:qualitative_comp}
\end{figure*}

\paragraph{Textual \& Visual Quality.} 
Finally, we evaluate textual quality using automated overlap and fluency metrics (Tab.~\ref{tab:automated_metrics}). Across generation backbones, \methodname\ achieves strong ROUGE-L scores. It attains the highest ROUGE-L under GPT-4o and Qwen-3-VL, while remaining competitive under GPT-5. Perplexity results indicate fluent slide text across models. Minor metric differences likely reflect the nature of slide writing, which favors concise bullet-style summaries over direct sentence reuse.

Qualitative examples under a fixed slide theme are shown in Fig.~\ref{fig:qualitative_comp}. Paper2Poster and PPTAgent mostly follow the original paper's section order, resulting in limited narrative arc (row~1-2). Paper2Poster also produces visually plain slides due to the lack of template support, while PPTAgent struggles to adapt layouts for varying figure sizes (col.~3). SlideGen achieves strong visual design but exhibits weaker narrative flow and content overlap (row~3). In contrast, \textit{ArcDeck} produces a clearer narrative flow (row~4: from problem context to prior work and finally to the proposed method). Additional qualitative results are provided in Appendix Section.~\ref{appendix:additional_qualitative}.

\paragraph{Discussion of Author-Prepared Slides.}
A key question in evaluating paper-to-slide generation is: \emph{how close are automated methods to human-prepared presentations?} While pairwise A/B tests in Tab.~\ref{tab:pairwise_ab} reveal relative strengths among systems, it cannot assess whether any method has reached a practically useful level of quality. To address this, we conduct holistic pairwise A/B comparisons against the original author-prepared slides, as reported in Tab.~\ref{tab:pairwise_ap}, jointly evaluating content organization, visual design, information delivery, and slide structure. Among all methods, \methodname\ achieves the highest average selection rate against Author-Prepared slides, confirming that methods producing better slides among themselves also compete more effectively with human reference slides.

\begin{figure*}[t!]
    \centering
    \includegraphics[width=0.99\linewidth]{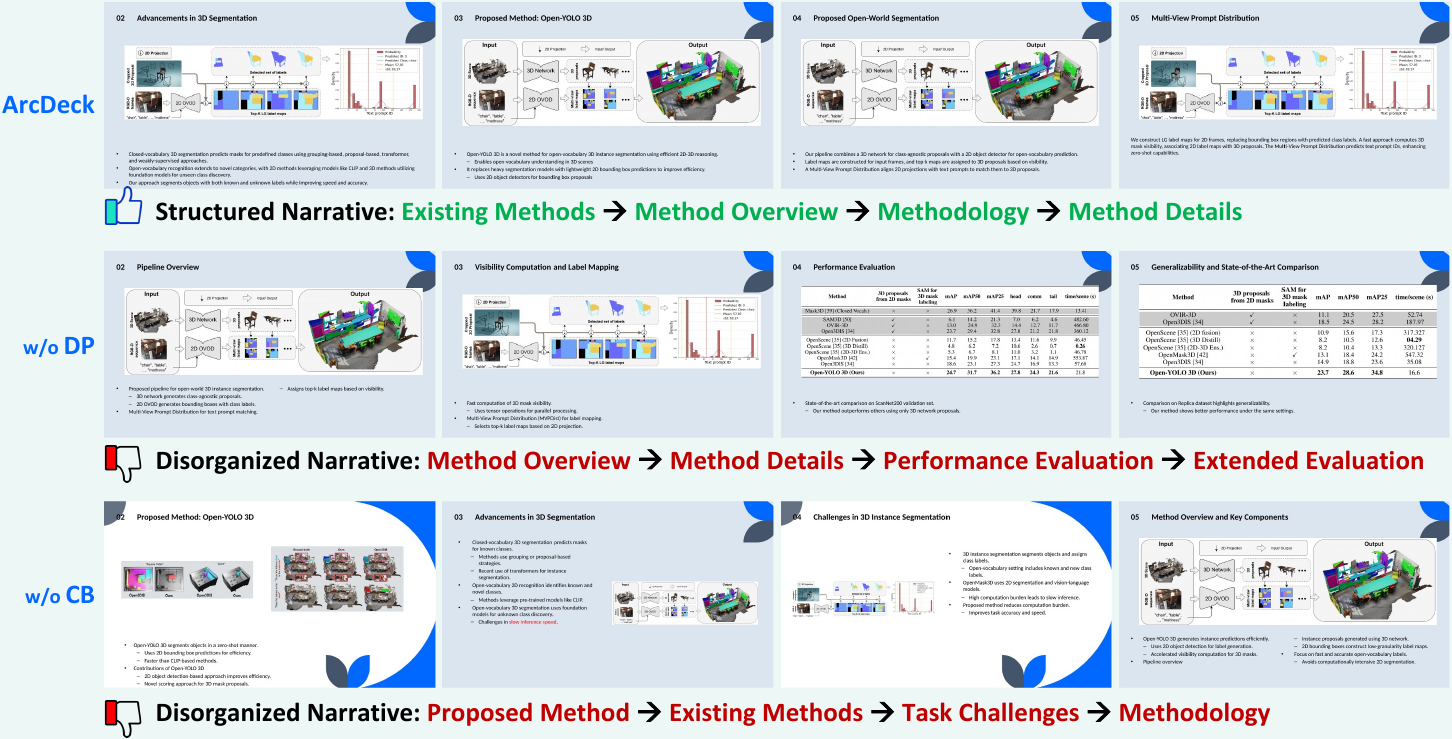}
    \caption{\textbf{Qualitative Ablation Study.} Comparison between \textit{ArcDeck} and variants without the Discourse Parser (DP) or Commitment Builder (CB).}
    \label{fig:qualitative_ablation}
\end{figure*}
\begin{figure*}[t!]
    \centering
    \includegraphics[width=\linewidth]{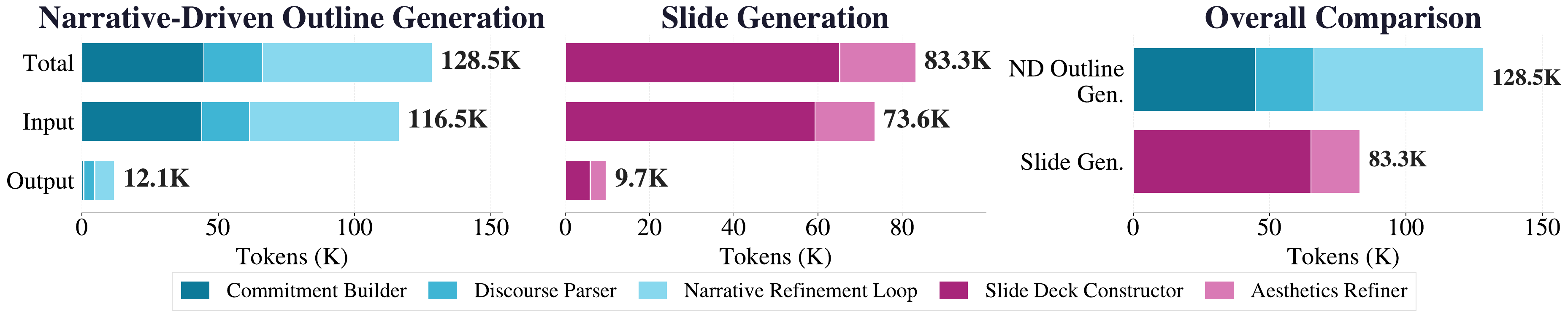}
    \caption{\textbf{Token Usage Analysis.}}
    \label{fig:token_count}
\end{figure*}

\paragraph{Ablation Study \& Token Analysis.}
To demonstrate the effectiveness of our Discourse Parser (DP) and Commitment Builder (CB), we conduct a qualitative ablation study (Fig.~\ref{fig:qualitative_ablation}). Without DP, the slide becomes condensed and less structured, omitting key method details due to the lack of discourse-guided content grouping. Without CB, slide ordering becomes disorganized, introducing the proposed method before prior work and task context. In contrast, the full \methodname\ pipeline produces a more coherent narrative progression—from existing methods to method overview and detailed methodology. We further report token usage across agents in Fig.~\ref{fig:token_count}. Given the multi-agent design of \methodname, this breakdown provides how tokens are distributed across agents during outline generation and slide generation. We show quantitative ablation results for each component of \methodname \space in Appendix Section.~\ref{appendix:naration_component_ablation}, \ref{appendix:rst_refinement_ablation} and \ref{appendix:aesthetic_refiner_ablation}.

\section{Conclusion}
In this work, we introduce \methodname, a narrative-driven framework for automatic paper-to-slide generation that formulates the task as a structured narrative reconstruction problem. By explicitly modeling discourse structure through RST-based parsing, establishing a global commitment to preserve the paper’s high-level intent, and refining outlines through a multi-agent critique–judge–revise loop, \methodname\ generates slide decks with improved narrative coherence and structural alignment. To support systematic evaluation, we also present \benchname, a curated paper–slide paired dataset designed to assess slide generation across multiple dimensions, including content, narrative coherence, and visual quality. Experimental results demonstrate that explicit discourse modeling substantially improves both narrative structure and information retention, while remaining competitive in textual and visual quality.

\clearpage
\newpage
{
    \small
    \bibliographystyle{unsrt}
    \bibliography{main}
}

\appendix
\clearpage

\begingroup
\renewcommand{\addcontentsline}[3]{}
\renewcommand{\addtocontents}[2]{}
\endgroup

\begingroup
\hypersetup{
    citecolor=eccvblue,
    linkcolor=eccvblue
}

\section*{Appendix Contents}

\startcontents[appendices]
\printcontents[appendices]{l}{1}{}
\endgroup

\newpage

\section{Narrative-Driven Outline: Additional Details}

\subsection{Human Evaluation of Slide Deck Content and Narrative Flow}
\label{appendix:human_study}
We conduct a human evaluation to assess the quality of generated slide decks in terms of content coverage and narrative flow. A total of 25 undergrad or MS/PhD students participated in the study. To ensure reliable evaluation of technical content, participants were asked to select the research area they were most familiar with among three options: Generative Models, Vision-Language/Multimodal Learning, and Computer Vision Core. For each selected area, participants were presented with slide decks generated by ArcDeck, SlideGen, and PPTAgent for five oral papers, and were asked to rank them based on overall content quality and narrative coherence.

\begin{figure}[h!]
    \centering
    \includegraphics[width=0.9\linewidth]{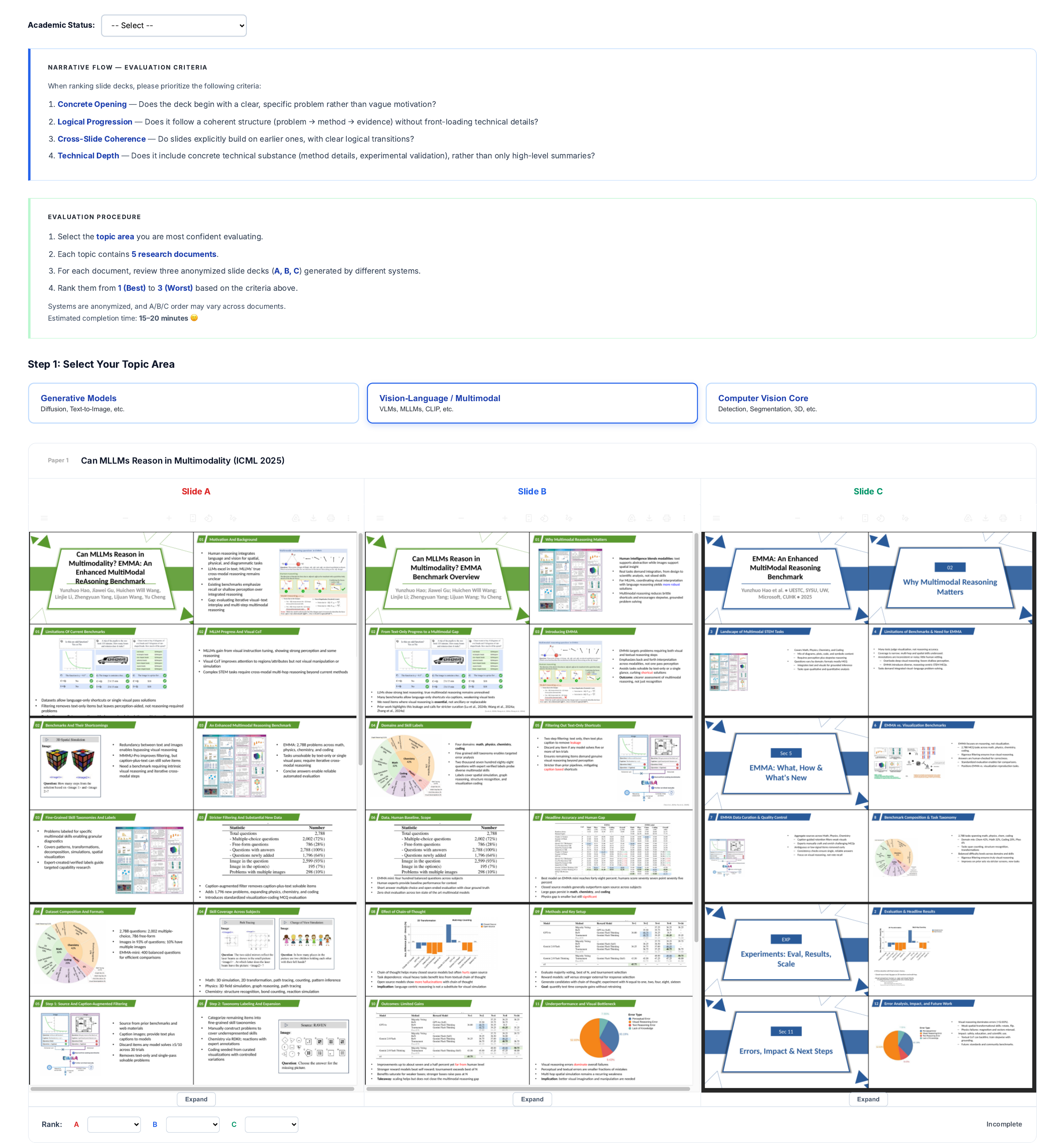}
    \caption{\textbf{Human Evaluation Interface.} Participants were shown slide decks generated by different methods and asked to rank them based on content quality and narrative flow.}
    \label{fig:human_eal}
\end{figure}

The evaluation was conducted through a custom web interface, shown in Fig.~\ref{fig:human_eal}, where participants could inspect the slide decks and provide their rankings. As shown in Tab.~\ref{tab:human_eval}, ArcDeck achieves the best average ranking across the evaluated papers, indicating that the generated slide decks provide the strongest narrative flow and overall content coherence among the compared baseline frameworks.

\begin{table}[h!]
  \caption{\textbf{Human Evaluation Results.}
  Participants ranked slide decks based on content quality and narrative flow, where Rank 1 = 3 points, Rank 2 = 2 points, and Rank 3 = 1 point. The best value in each topic is highlighted in \textcolor{winGreen}{green}, and runner-up is underlined.}
  \label{tab:human_eval}
  \centering
  \setlength{\tabcolsep}{4pt}
  \renewcommand{\arraystretch}{1.1}
  \begin{tabular}{lcccc}
    \toprule
    & \multicolumn{4}{c}{\textbf{Human Evaluation Topic}} \\
    \cmidrule(lr){2-5}
    \textbf{Method} 
    & \textbf{V-L / MM} 
    & \textbf{Gen. Models} 
    & \textbf{CV Core} 
    & \textbf{Overall} \\
    \midrule
    PPTAgent
    & 1.03 & 1.13 & 1.20 & 1.10 \\
    SlideGen
    & \underline{2.42} & \underline{2.23} & \underline{2.14} & \underline{2.30} \\
    \rowcolor{gray!15} \methodname
    & \textcolor{winGreen}{2.55} & \textcolor{winGreen}{2.63} & \textcolor{winGreen}{2.66} & \textcolor{winGreen}{2.60} \\
    \bottomrule
  \end{tabular}
\end{table}

\subsection{Analysis of Global Commitment Conditioning}

We provide additional qualitative results to showcase the effects of Global Commitment conditioning with respect to the target audience and presentation duration on the generated slides.

\begin{figure}[h!]
    \centering
    \includegraphics[width=\linewidth]{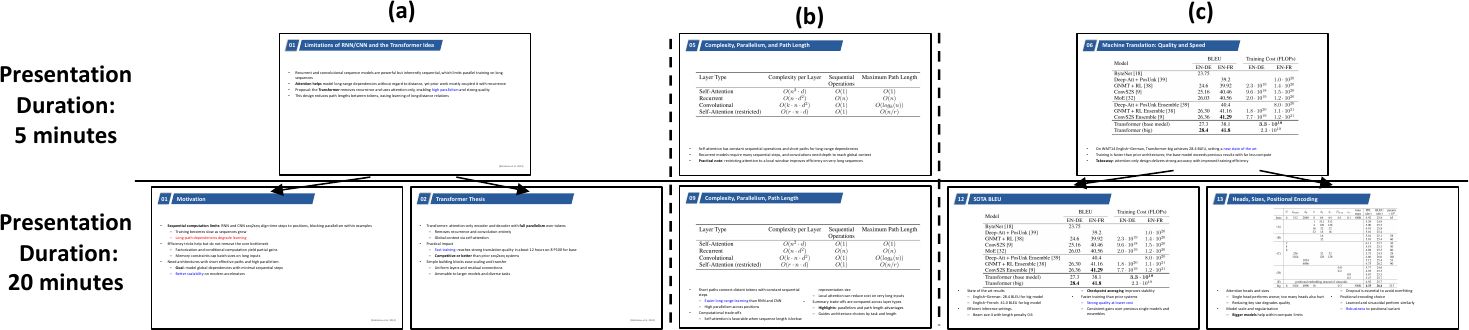}
    \caption{\textbf{Qualitative Comparison of ArcDeck-generated Slides under Different Presentation Durations.} 20-minute presentation includes more detailed content, while the 5-minute presentation preserves the key points of the paper.}
    \label{fig:presentation_duration_qualitative}
\end{figure}

For presentation duration, a qualitative comparison of the 5-minute and 20-minute presentations is shown in Fig.~\ref{fig:presentation_duration_qualitative}. In (a), multiple slides presenting different information in the 20-minute presentation are condensed into a single slide in the 5-minute presentation with less detail, in order to satisfy the time constraint. In (b), the textual content is substantially reduced and summarized in the 5-minute presentation relative to the 20-minute presentation. In (c), only the most significant quantitative result is retained in the 5-minute version, whereas results under different configurations are presented only in the 20-minute presentation. These examples demonstrate effective condensation of information based on presentation duration: the 5-minute presentation preserves the key points, while the 20-minute presentation retains more detailed content.

\begin{figure}[h!]
    \centering
    \includegraphics[width=\linewidth]{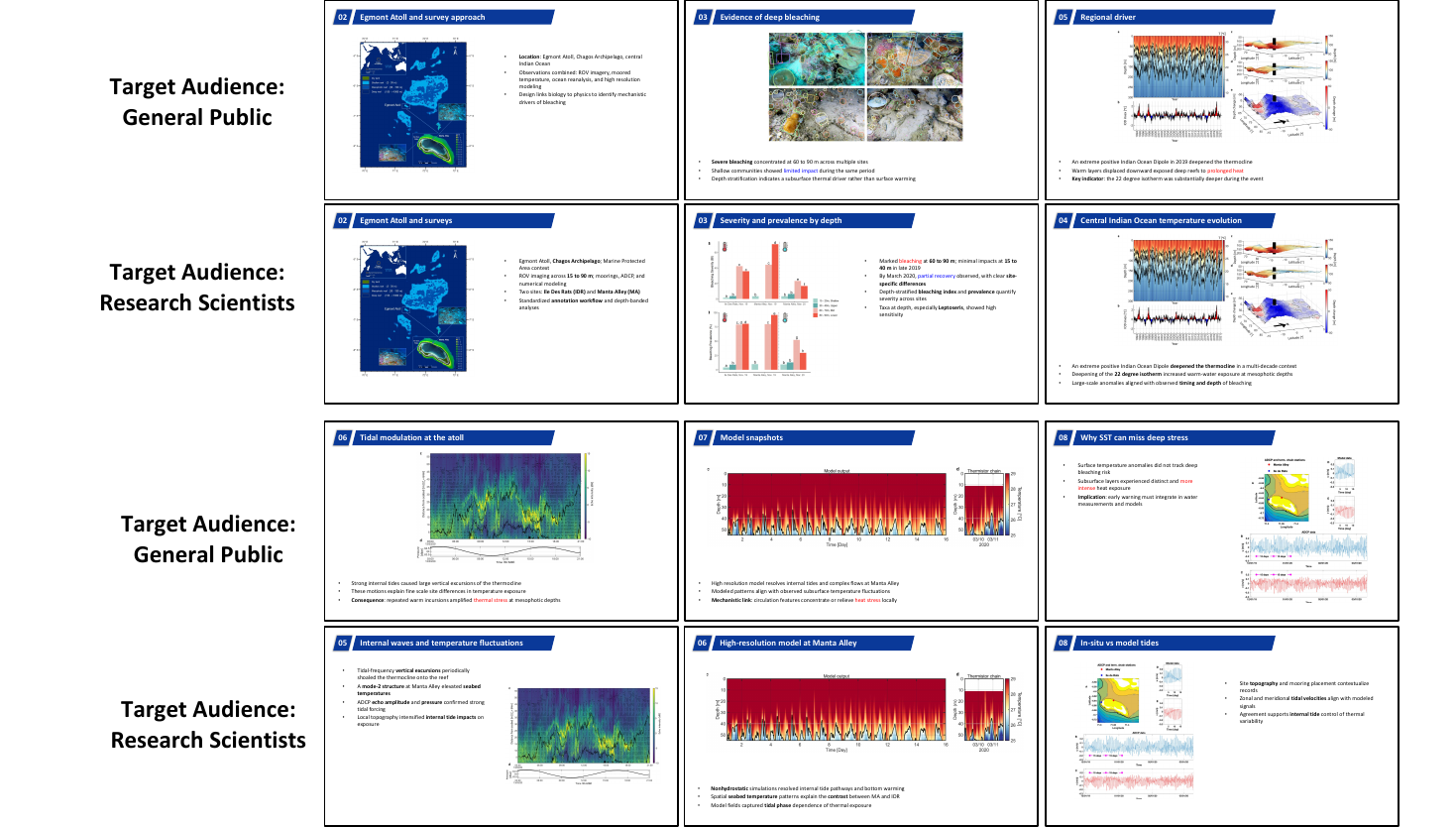}
    \caption{\textbf{Qualitative Comparison of ArcDeck-generated Slides under Different Target Audiences.} The figure illustrates how ArcDeck adapts content selection, terminology, and level of technical detail according to the specified target audience.}
    \label{fig:target_audience_qualitative}
\end{figure}

For the target audience analysis, we generate slide decks for two audiences: \textit{General Public} and \textit{Research Scientists}. A qualitative comparison of these presentations is shown in Fig.~\ref{fig:target_audience_qualitative}. The slide deck generated for the \textit{General Public} focuses on providing high-level information in an accessible manner, whereas the slide deck for \textit{Research Scientists} presents more detailed information about the method, experiments, and results. In terms of terminology choice, the \textit{General Public} deck adopts more accessible and less technical language, while the \textit{Research Scientists} deck uses more domain-specific and technical terminology. In terms of level of abstraction, the \textit{General Public} version presents the work at a higher conceptual level, whereas the \textit{Research Scientists} version offers a more detailed and technically grounded treatment.

\subsection{Analysis of Discourse Parser}
\begin{figure}[h!]
    \centering
    \includegraphics[width=\linewidth]{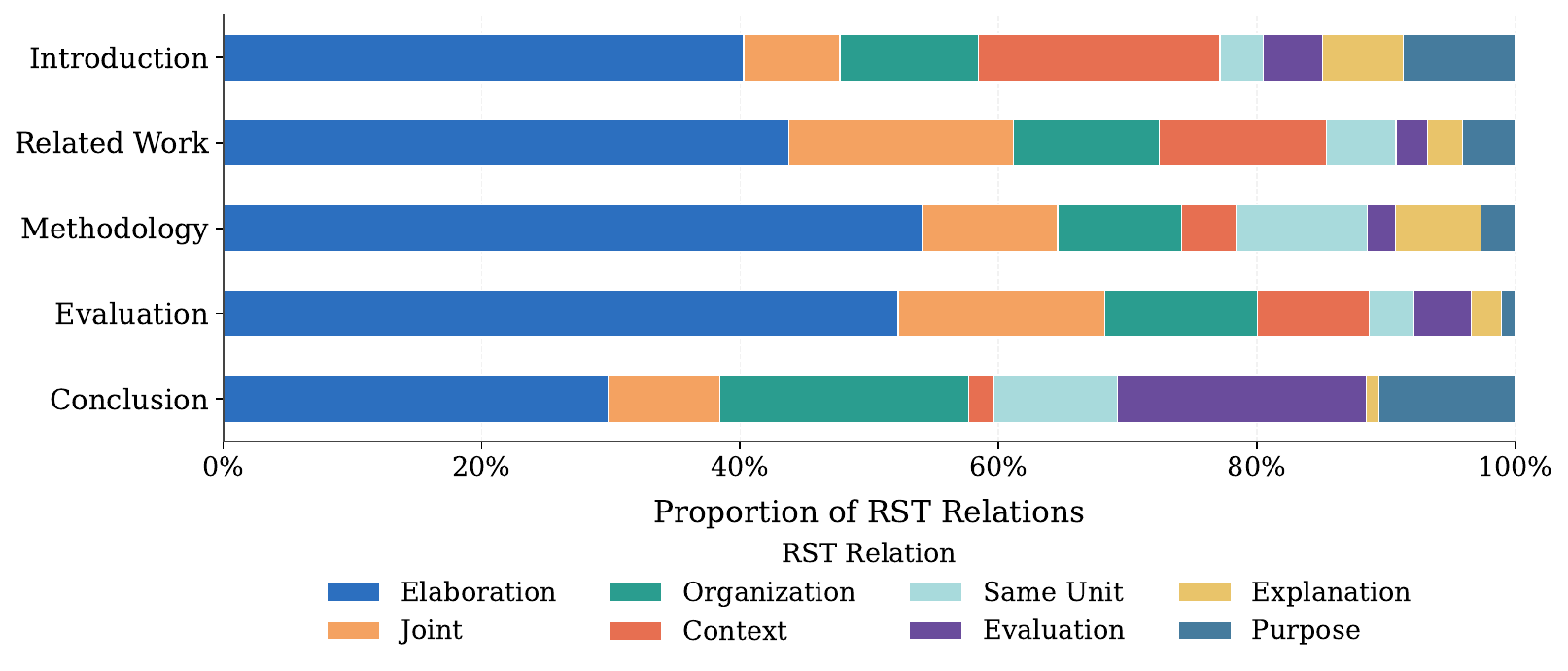}
    \caption{\textbf{Distribution of Discourse Tree Relations across Section Groups.} Stacked bars represent the relative proportion of each discourse relation within each section group.}
    \label{fig:rst_heatmap}
\end{figure}

We conduct further analysis to investigate the properties of the generated discourse trees. Fig.~\ref{fig:rst_heatmap} shows the distribution of discourse-tree relations across section groups. We group each section into one of the following categories: Introduction, Related Work, Methodology, Evaluation, and Conclusion, and compute the ratio of each discourse relation within each group. Across all sections, Elaboration is the most frequently used relation, indicating that information is progressively elaborated throughout the discourse, with especially high occurance in the Methodology and Evaluation sections. Similarly, Explanation appears most often in the Introduction and Methodology sections, where the domain and the proposed method are described in detail. In addition, Context is used extensively in the Introduction and Related Work sections, where the context of the work is established and the necessary background is provided. The Evaluation relation is most common in the Evaluation and Conclusion sections, which is consistent with the fact that assessment and interpretation of results are primarily presented in these parts of the paper. Overall, these findings indicate that the generated discourse trees align with the expected structural patterns of academic papers, indicating that the generated discourse trees are well aligned with the discourse structure of academic papers.

\begin{figure}[h!]
    \centering
    \includegraphics[width=\linewidth]{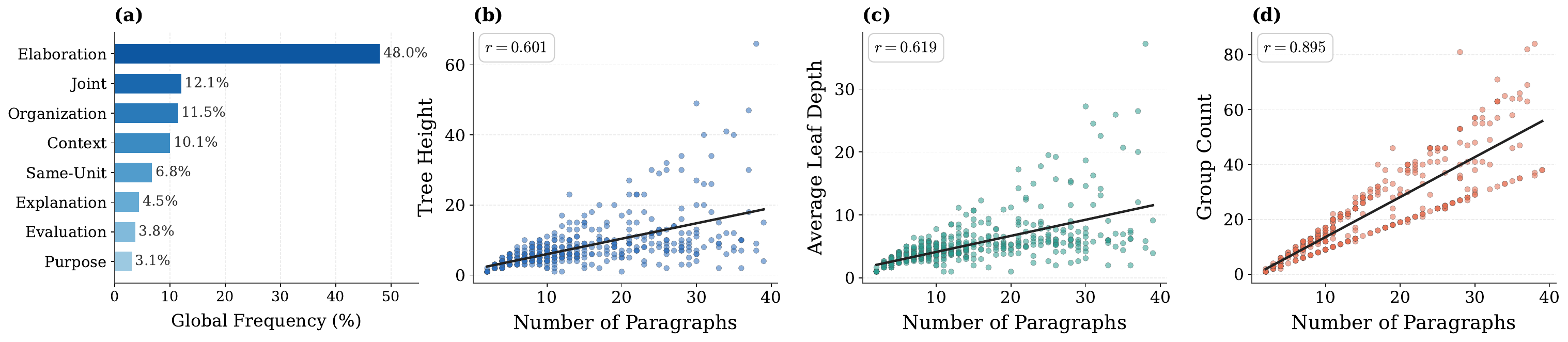}
    \caption{\textbf{Relationship of Generated Discourse-Tree Height, Average Leaf Depth, and Group Count per Paragraph to the Number of Paragraphs.} Pearson correlation coefficients are shown in the top-left corner of each figure.}
    \label{fig:rst_scatter}
\end{figure}

We further illustrate the relationship between the properties of the generated discourse trees for each section and the number of paragraphs in that section in Fig.~\ref{fig:rst_scatter}. In RST theory, there is expected to be a near-linear relationship between the number of EDUs and the number of groups in an RST tree. As shown in Fig.~\ref{fig:rst_scatter}(d), our generated discourse trees exhibit this property, supporting the soundness of our discourse parser. Similarly, the correlations between tree height, average leaf depth, and the number of paragraphs shown in Fig.~\ref{fig:rst_scatter}(b) and Fig.~\ref{fig:rst_scatter}(c) further suggest that the generated trees are well aligned with the discourse structure of academic papers. 

\subsection{Ablation Study of Narrative-Driven Outline Generation}
\label{appendix:naration_component_ablation}
We perform an ablation study to evaluate the contribution of the Discourse Parser, Global Commitment, and Narrative Refinement Loop components. To analyze their effects, we compare ArcDeck with variants in which each component is removed individually. We evaluate these different setups using VLM-as-Judge, and further assess narrative flow through A/B testing between the full ArcDeck system and each corresponding ablated variant. As shown in Tab.~\ref{tab:narrative_ablation}, removing each component leads to a significant decrease in Narrative Flow scores. Similarly, in the pairwise comparison, the full system achieves a significant win rate over the ablated variants. These findings indicate that each component contributes significantly to the Narrative Flow of the paper.


\vspace{-0.3cm}
\begin{table}[h!]
\caption{\textbf{Narrative-Driven Outline Generation Ablation Study.} Narrative Flow (NF) scores for different ablated versions, along with the full ArcDeck model’s narrative flow pairwise comparison win rate against these ablated versions, are shown.}
\label{tab:narrative_ablation}
\centering
\setlength{\tabcolsep}{4pt}
\renewcommand{\arraystretch}{1.1}
\resizebox{0.9\linewidth}{!}{
\begin{tabular}{lc c}
\toprule
& \multicolumn{1}{c}{\textbf{VLM-as-Judge $\uparrow$}}
& \multicolumn{1}{c}{\textbf{Pairwise Comparison}} \\
\cmidrule(lr){2-2}
\cmidrule(lr){3-3}

\textbf{Method}
& \textbf{NF}
& \textbf{ArcDeck Win Rate \%} \\
\midrule

\methodname \space w/o Discourse Parser
& 7.50
& \textcolor{winGreen}{89.1} \\

\methodname \space w/o Global Comm.
& 7.52
& \textcolor{winGreen}{94.5} \\

\methodname \space w/o Narr. Ref. Loop
& 8.68
& \textcolor{winGreen}{61.8} \\

\methodname 
& \textcolor{winGreen}{9.70}
& -- \\

\bottomrule
\end{tabular}
}
\end{table}


\subsection{Effect of the Narrative Refinement Loop}
\label{appendix:rst_refinement_ablation}
To analyze the effect of the narrative refinement loop, we selected five papers that underwent all three iterations of the loop for improvement. For 5 papers, we compare the results of generated slides using the outlines at each step of the narrative refinement loop. As shown in Tab.~\ref{tab:nrl_ablation}, the narrative refinement loop iteratively improves the narrative flow of the outline. As a result of the improved narrative flow, ArcDeck Final is the overall winner, and two iterations achieves a higher win rate than one iteration.
\begin{table}[h!]
\caption{\textbf{Narrative Refinement Loop (NRL) Improvement Study.} Narrative Flow (NF) scores for different numbers of iterations, together with the win rate relative to the first iteration, are shown.}
\label{tab:nrl_ablation}
\centering
\setlength{\tabcolsep}{4pt}
\renewcommand{\arraystretch}{1.1}
\resizebox{0.8\linewidth}{!}{
\begin{tabular}{lc c}
\toprule
& \multicolumn{1}{c}{\textbf{VLM-as-Judge $\uparrow$}}
& \multicolumn{1}{c}{\textbf{Pairwise Comparison}} \\
\cmidrule(lr){2-2}
\cmidrule(lr){3-3}

\textbf{NRL Iterations \#}
& \textbf{NF} & \textbf{NF Win Rate vs. 1st It.} \\
\midrule


1 iteration
& 4.18
& -- \\

2 iterations
& 7.00
& 56.4 \% \\

3 iterations (Final)
& 7.34
& 100 \%\\

\bottomrule
\end{tabular}
}
\end{table}


\clearpage
\newpage

\section{ArcBench: Additional Information and Statistics}
\label{appendix:dataset}
\benchname\ is constructed through a two-stage process: we first collect a broad pool of
paper-slide pairs from major AI/CV venues, then apply a series of principled filtering steps
to curate a high-quality benchmark subset.
Tab.~\ref{tab:arcbench_distribution} and Fig.~\ref{fig:arcbench_stats} summarize the
composition of both the full 994-pair pool and the final 100-pair benchmark.


\definecolor{poolcol}{RGB}{31,119,180}
\definecolor{benchcol}{RGB}{214,95,14}

\newcommand{\pn}[1]{{\color{poolcol}#1}}
\newcommand{\bn}[1]{{\color{benchcol}#1}}
\newcommand{\cb}[2]{\pn{#1}\,/\,\bn{#2}}
\newcommand{\po}[1]{\pn{#1}\,/\,{---}}
\newcommand{\non}{---}
\begin{table}[h!]
\centering
\small
\setlength{\tabcolsep}{4pt}
\caption{\textbf{\benchname\ Venue and Year Distribution.}
  Each cell: \pn{\textbf{pool}}\,|\,\bn{\textbf{bench.}}
  Dashes indicate no coverage.}
\label{tab:arcbench_distribution}
\begin{tabular}{l
  @{\hspace{6pt}} r @{\,} c @{\,} l
  @{\hspace{6pt}} r @{\,} c @{\,} l
  @{\hspace{6pt}} r @{\,} c @{\,} l
  @{\hspace{6pt}} r @{\,} c @{\,} l
  @{\hspace{8pt}} r @{\,} c @{\,} l}
  \toprule
  \textbf{Venue}
    & \multicolumn{3}{c}{\textbf{2022}}
    & \multicolumn{3}{c}{\textbf{2023}}
    & \multicolumn{3}{c}{\textbf{2024}}
    & \multicolumn{3}{c}{\textbf{2025}}
    & \multicolumn{3}{c}{\textbf{Total}} \\
  \midrule
  \textit{CVPR}
    &              & {---} &
    &              & {---} &
    & \pn{52}  & | & \bn{1}
    & \pn{176} & | & \bn{1}
    & \pn{228} & | & \bn{2}  \\
  \textit{ECCV}
    &              & {---} &
    &              & {---} &
    & \pn{222} & | & \bn{27}
    &              & {---} &
    & \pn{222} & | & \bn{27} \\
  \textit{ICCV}
    &              & {---} &
    &              & {---} &
    &              & {---} &
    & \pn{296} & | & \bn{4}
    & \pn{296} & | & \bn{4}  \\
  \textit{ICLR}
    &              & {---} &
    &              & {---} &
    & \pn{12}  & | & {---}
    & \pn{51}  & | & \bn{27}
    & \pn{63}  & | & \bn{27} \\
  \textit{ICML}
    & \pn{52}  & | & {---}
    & \pn{38}  & | & \bn{3}
    & \pn{40}  & | & \bn{17}
    & \pn{22}  & | & \bn{8}
    & \pn{152} & | & \bn{28} \\
  \textit{NeurIPS}
    &              & {---} &
    & \pn{14}  & | & \bn{4}
    & \pn{10}  & | & \bn{4}
    & \pn{9}   & | & \bn{4}
    & \pn{33}  & | & \bn{12} \\
  \midrule
  \textbf{Total}
    & \pn{\textbf{52}}  & | & {---}
    & \pn{\textbf{52}}  & | & \bn{\textbf{7}}
    & \pn{\textbf{336}} & | & \bn{\textbf{49}}
    & \pn{\textbf{554}} & | & \bn{\textbf{44}}
    & \pn{\textbf{994}} & | & \bn{\textbf{100}} \\
  \bottomrule
\end{tabular}
\end{table}

\begin{figure}[h!]
  \centering
  \includegraphics[width=\linewidth]{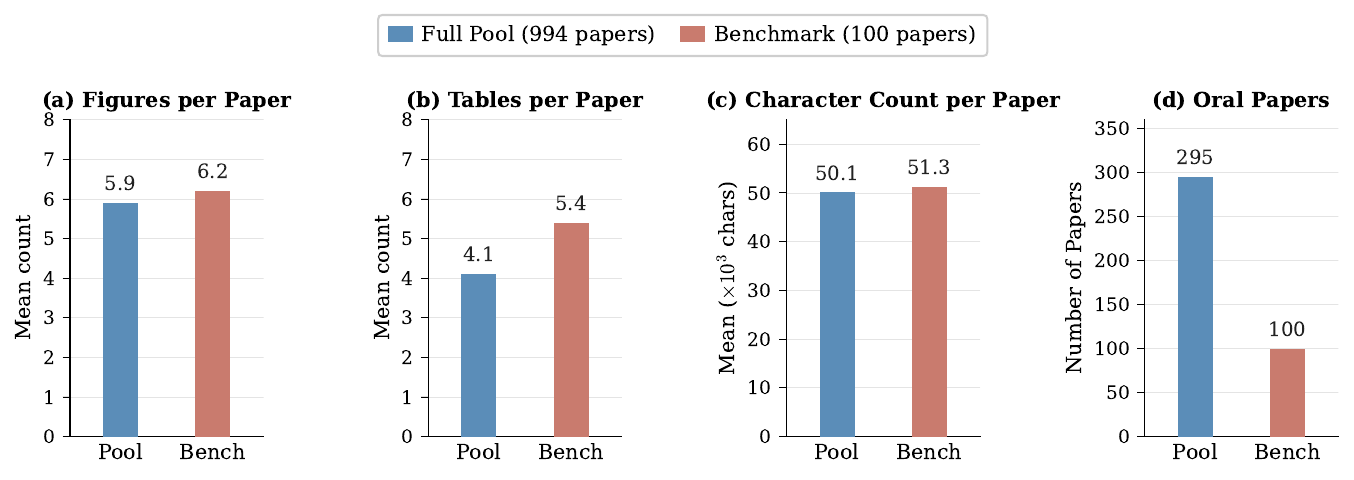}
  \caption{\textbf{Key Statistics of Papers and Slide Decks in \benchname.}
    We report distributions for the full paper pool and the curated benchmark,
    including the number of figures, tables, character counts, and Oral papers.}
  \label{fig:arcbench_stats}
\end{figure}

\subsection{Full Dataset Collection.}
The full \benchname\ pool consists of 994 paper-slide pairs collected directly from the
official proceedings and presentation archives of six major computer vision and machine
learning venues: CVPR, ECCV, ICCV, ICML, ICLR, and NeurIPS.
Papers span publication years 2022 through 2025, with a larger share drawn from more
recent years reflecting the growing availability of author-prepared presentation materials
online.
The pool comprises 295 oral and 699 poster presentations.

\subsection{Filtering to the 100-Pair Benchmark.}
To construct the curated 100-pair \benchname\ benchmark we apply three sequential filtering
criteria.
First, we restrict to \emph{oral presentations only}, as oral papers are typically
accompanied by more deliberately crafted, author-prepared slide decks that exhibit clear
narrative design - essential for providing a reliable human reference standard.
Second, we require at least \emph{3 figures} per paper to ensure sufficient visual material
for evaluating content and figure fidelity.
Third, we require at least \emph{3 tables} per paper to guarantee sufficient quantitative
content for assessing information coverage.
All 100 benchmark pairs are exclusively oral presentations.

\subsection{Benchmark Statistics.}
\begin{figure}[h!]
  \centering
  \includegraphics[width=\linewidth]{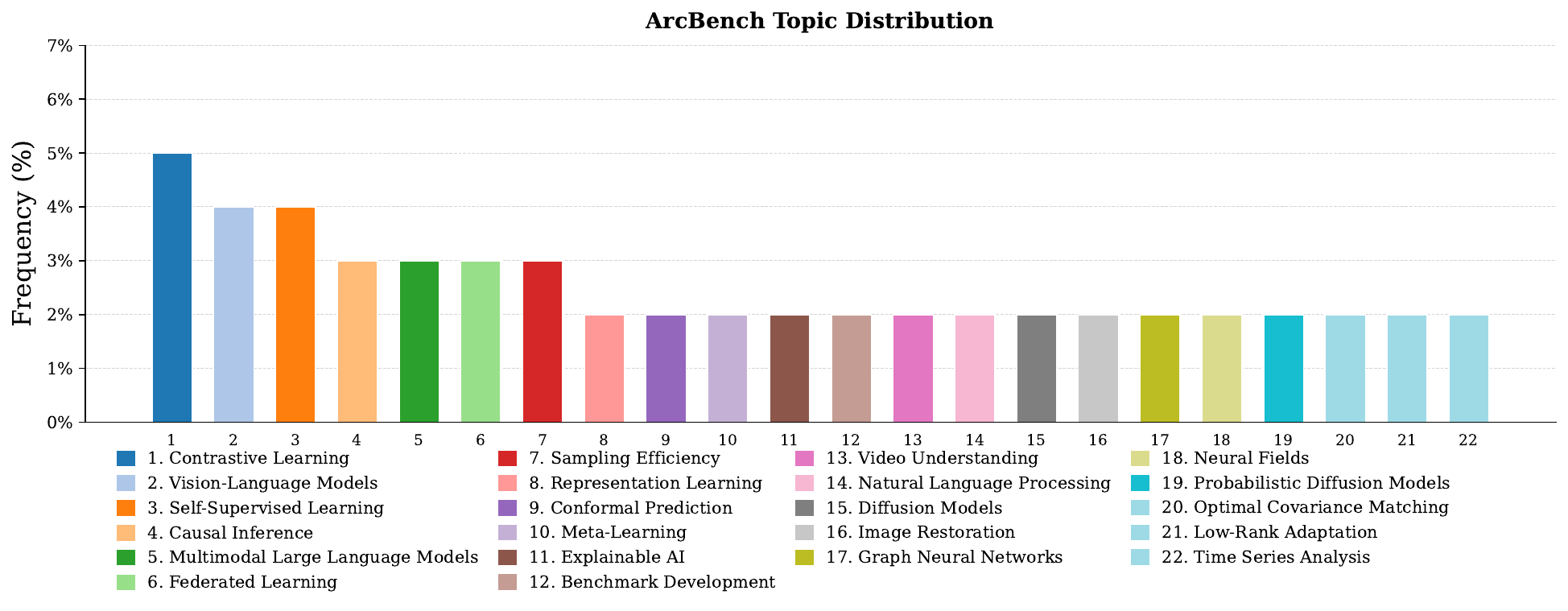}
    \caption{\textbf{\benchname\ Topic Distribution.}
    We report the distribution of papers across research topics.}
  \label{fig:arcbench_topic_distribution}
\end{figure}

As shown in Fig.~\ref{fig:arcbench_stats}, the benchmark exhibits meaningful visual and
quantitative content density.
Tab.~\ref{tab:arcbench_distribution} shows the per-venue and per-year breakdown for both
splits.
The benchmark covers different topic clusters across CV/ML domains as shown in Fig. ~\ref{fig:arcbench_topic_distribution}, ensuring topical diversity that challenges generation methods to handle
varied content structures, notation conventions, and domain-specific visual elements.

\subsection{Comparison with Prior Datasets.}
Tab.~\ref{tab:dataset_comparison} places \benchname\ in the context of existing paper-to-slide and slide generation datasets. Prior academic datasets such as DOC2PPT~\cite{fu2022doc2ppt} and SciDuet~\cite{sun2021d2s} cover general scientific or NLP/ML papers without restricting to
oral presentations or enforcing content-density thresholds.
Zenodo10K~\cite{zheng2025pptagent} broadens coverage to multi-domain presentations from
the Zenodo repository, departing from the academic paper-to-slide setting entirely.
SLIDESBENCH~\cite{ge2025autopresent} addresses a distinct paradigm-generating individual
slides from natural language instructions rather than full scientific papers.
SlideGen~\cite{liang2025slidegen} is closest in spirit, using 200 AI papers from three
venues with oral-only filtering but no minimum figure/table requirements.
The key distinguishing properties of \benchname\ are:
(i)~exclusive focus on \emph{oral, author-prepared} presentations from six top-tier AI/CV venues;
(ii)~principled content-density filtering - figure and table thresholds;
(iii)~broad topical diversity spanning CV and ML subfields; and
(iv)~dual release of both the curated 100-pair benchmark and the full 994-pair pool for
the community.


\definecolor{oursblue}{RGB}{219, 234, 254}

\begin{table}[h!]
\centering
\small
\setlength{\tabcolsep}{4pt}
\renewcommand{\arraystretch}{1.25}
\caption{\textbf{Comparison of \benchname\ with Prior Paper-to-Slide Datasets.}
$^\dagger$ Generates individual slides from NL instructions rather than full papers.
$^\ddagger$ CVPR, ECCV, ICCV, ICML, ICLR, NeurIPS.}
\label{tab:dataset_comparison}
\begin{tabularx}{\linewidth}{%
  >{\raggedright\arraybackslash}p{2.5cm}  
  c                                         
  r                                         
  >{\raggedright\arraybackslash}p{2.1cm}  
  >{\raggedright\arraybackslash}X          
  >{\raggedright\arraybackslash}X          
  c}                                        
  \toprule
  \textbf{Dataset} & \textbf{Year} & \textbf{Pairs}
    & \textbf{Venues} & \textbf{Domain} & \textbf{Type} & \textbf{Pub.} \\
  \midrule
  DOC2PPT
    & 2022 & 5,873     & General academic    & Multi-domain & Oral \& Poster & \cmark \\
  SciDuet
    & 2021 & 1,088     & ICML, NeurIPS, ACL  & NLP\,/\,ML   & Oral \& Poster & \cmark \\
  Zenodo10K
    & 2025 & 10K & ---              & Multi-domain & ---          & \cmark \\
  SLIDESBENCH$^\dagger$
    & 2025 & 7,585     & ---                 & General      & ---            & \cmark \\
  SlideGen
    & 2025 & 200       & ICLR, ICML, NeurIPS & AI\,/\,ML    & Oral-only  & \xmark \\
  \midrule
  \rowcolor{oursblue}
  \textbf{\benchname}
    & \textbf{2025} & \textbf{100} & \textbf{6 venues}$^\ddagger$
    & \textbf{CV\,/\,ML} & \textbf{Oral-only} & \cmark \\
  \bottomrule
\end{tabularx}
\end{table}

\section{Detailed Experimental Setup and Evaluation}

\subsection{Generation Pipeline Implementation}
\label{appendix:general_implementation}
In this work, we use two closed-source backbones (\textbf{GPT-4o} and \textbf{GPT-5}) and one open-source model (\textbf{Qwen3-VL-32B-Instruct}) to generate results with our method. To produce editable .pptx files, we use the python-pptx library. However, our framework is not limited to this output format and can also generate slides using alternative rendering formats such as JavaScript and LaTeX Beamer, as demonstrated in Fig.~\ref{fig:clause_js_results} and Fig.~\ref{fig:clause_latex_results}. All generated slides follow a standard 16:9 aspect ratio, with dimensions of 13.33 inches × 7.5 inches. For preprocessing the input document, we use the Docling library, which extracts figures and tables from the input PDF and parses the document text into Markdown format. As a preprocessing step we extract the reference section and create a short citation asset, and also remove every content below reference in the markdown to make the pipeline token effecient. This markdown is then used as input by our agents during the slide generation process. In addition to the input PDF, our method also takes presentation duration and target audience as inputs, allowing us to control the level of detail and overall length of the generated presentation. This enables our method to adapt the slide content to different presentation settings. Furthermore, users can provide their own slide template with a personalised theme, as long as the underlying slide layout structure remains consistent with our predefined layouts.

\paragraph{Narrative-Driven Outline Generation.} The discourse trees generated by the Discourse Parser are stored in JSON format. Each relation includes (i) the relation type (NS or MN) and (ii) the associated nucleus and satellite, or left and right units, depending on the relation type. The Commitment Builder generates the Global Commitment in Markdown format, which is then stored and provided to all agents in the same format. 

The outputs of all agents in the Narrative Refinement Loop are likewise stored and passed in JSON format. The outline of each slide consists of the slide number, section number and title, slide title, and the paragraph IDs corresponding to the paragraphs assigned to that slide.

\paragraph{Slide Deck Constructor.} The Slide Deck Constructor performs two tasks; (i) \textbf{Asset Matching.} In this step, the agent determines which images and tables should be used to complement each slide content based on the outline generated by the Narrative-Driven Outline Generation stage. The agent considers both the slide text and figure captions to evaluate compatibility. It then outputs a structured JSON specifying which figures or tables should be included in each slide, along with reasoning to support the selection. (ii) \textbf{Slide Deck Construction.} In this step, the agent determines the appropriate slide layout, content structure (e.g., bullet points or paragraphs), and extracts the relevant citations derived from the content outline. The agent then produces a structured JSON slide plan, which is later used to generate the final \texttt{.pptx} file. During the \texttt{.pptx} generation stage, the extracted references are matched with the corresponding reference assets, and a short-form citation is added to the bottom corner of the slide. 

\paragraph{Aesthetic Refiner.} The Aesthetic Refiner improves the generated slides by identifying slides that lack visual elements and adding relevant images when appropriate, enhancing text formatting (e.g., text color and boldface), and refining slide content when it is too sparse. The agent can indicate which words should be highlighted or colored by wrapping them with \textit{textbf} or \textit{textcolor} LaTeX functions. These annotations are later parsed and converted into corresponding formatting instructions for \textit{python-pptx}. The final output is a refined slide plan to improve both visual quality and content presentation. 

\subsection{Evaluation Implementation Details}
\subsubsection{Judge Models \& Iteration Protocol.}
We build a VLM-as-Judge evaluation system to assess auto-generated research presentations across three complementary evaluation types: VLM as Judge, pairwise comparison, and quiz-based coverage. Two judges are employed: \textbf{Qwen3-VL-30B-A3B-Instruct}, run for \textbf{11~iterations} per
baseline, and \textbf{GPT-5}, run for \textbf{3~iterations} to
limit API credit usage. The following paragraphs describe each evaluation
type in detail.
\subsubsection{VLM-as-Judge.}
Each presentation is scored independently across four dimensions on a
0--10 scale using strict binary checklists (10 criteria, 1~point each),
where a criterion is awarded only when unambiguously satisfied.

\begin{figure}[h!]
    \centering
    \includegraphics[width=\linewidth, height=0.85\textheight, keepaspectratio]{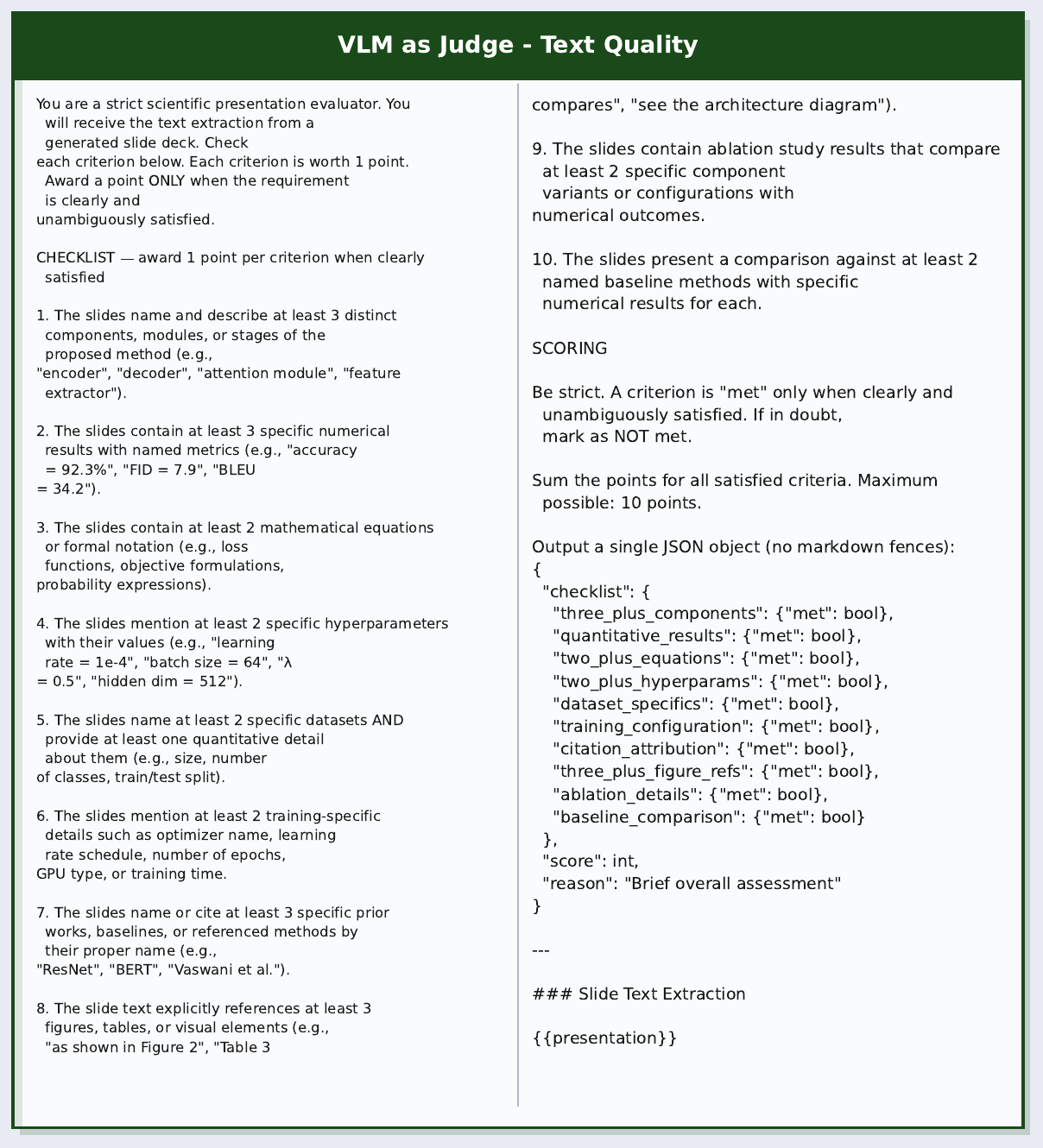}
    \caption{\textbf{VLM as Judge - Text Quality Evaluation Prompt.}}
    \label{fig:vlm_as_judge_text_quality}
\end{figure}

\noindent \emph{Text Quality (TQ)} (Fig.~\ref{fig:vlm_as_judge_text_quality}) assesses whether the
generated slides preserve the technical substance of the source paper. A
common failure mode in automated slide generation is over-summarisation,
where slides correctly identify the paper's topic but omit the mathematical
formulations, quantitative comparisons, and implementation specifics that make
a presentation scientifically useful. TQ directly penalises this by checking
for the presence of concrete technical content such as equations, named
baselines with numbers, hyperparameter values, and ablation results.

\begin{figure}[h!]
    \centering
    \includegraphics[width=\linewidth, height=0.85\textheight, keepaspectratio]{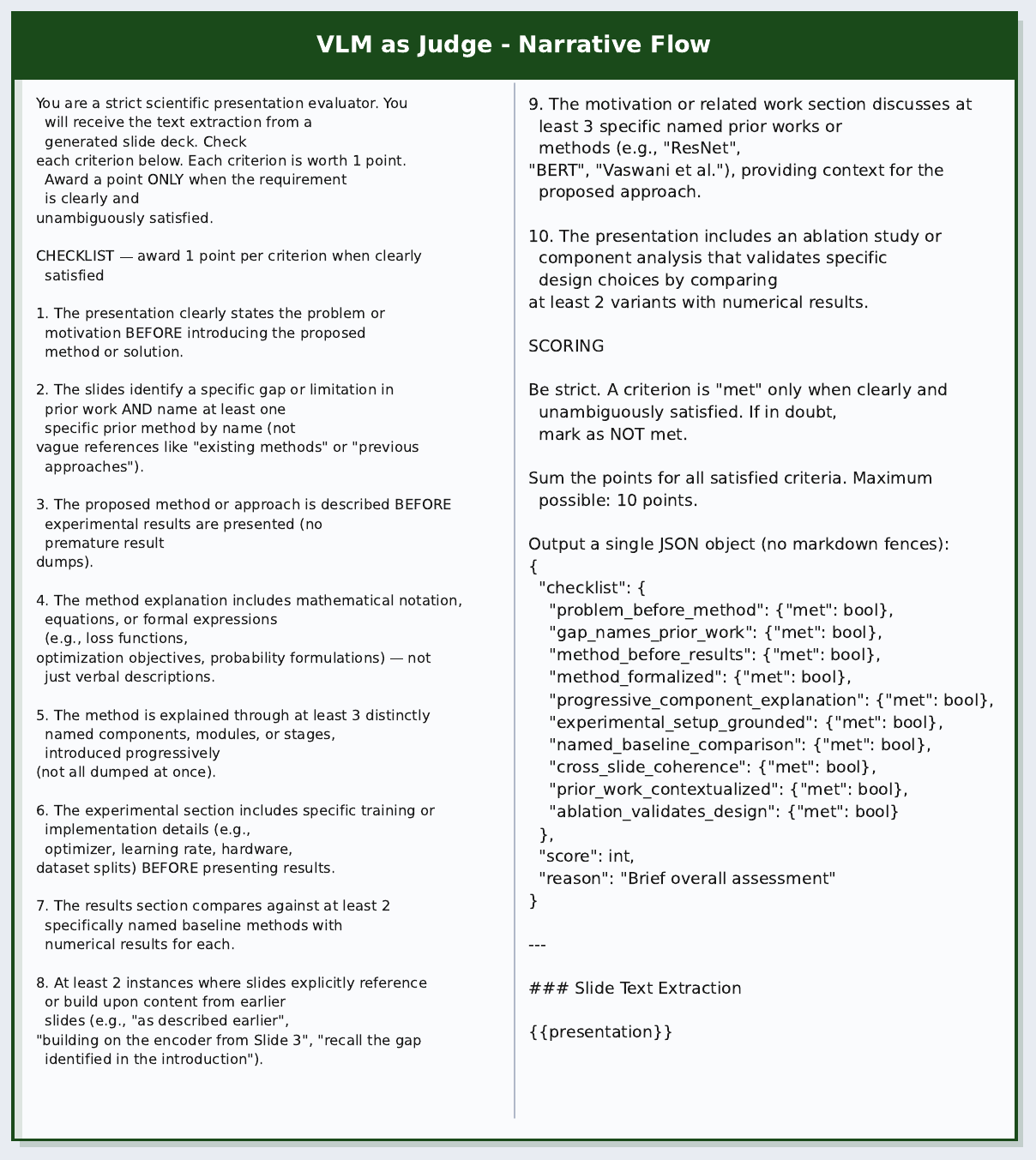}
    \caption{\textbf{VLM as Judge - Narrative Flow Evaluation Prompt.}}
    \label{fig:vlm_as_judge_narrative_flow}
\end{figure}

\noindent \emph{Narrative Flow (NF)} (Fig.~\ref{fig:vlm_as_judge_narrative_flow}) assesses whether the
slides tell a coherent story rather than presenting content as an unordered
collection of facts. Unlike documents, presentations must guide the audience
through a logical arc: establishing the problem before proposing a solution,
grounding claims in prior work, and building each section on what came before.
NF rewards presentations that respect this ordering and make cross-slide
connections explicit, reflecting the kind of deliberate narrative structure
that distinguishes a good talk from a mere paper dump.

\begin{figure}[h!]
    \centering
    \includegraphics[width=\linewidth, height=0.85\textheight, keepaspectratio]{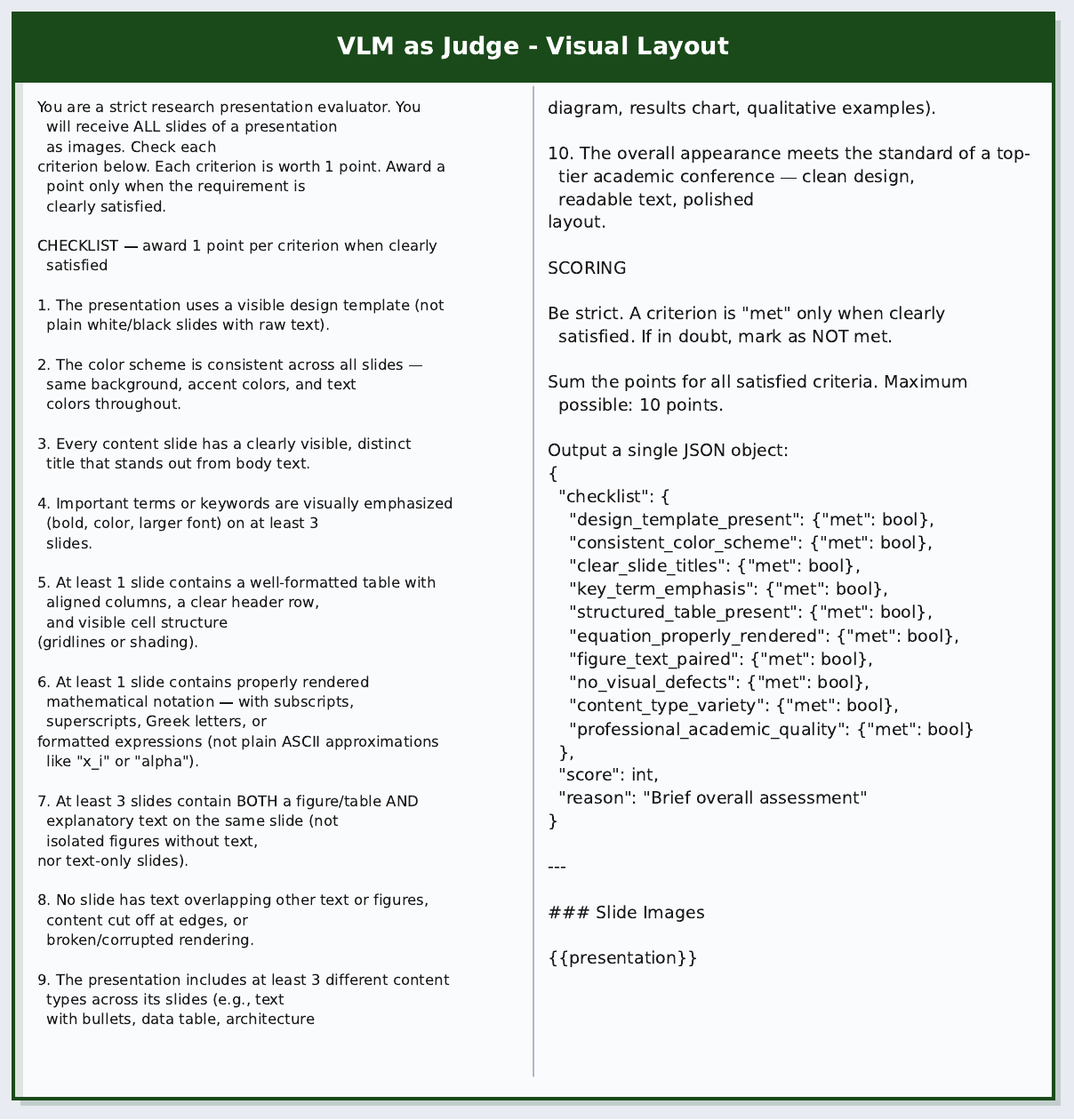}
    \caption{\textbf{VLM as Judge - Visual Layout Evaluation Prompt.}}
    \label{fig:vlm_as_judge_visual_layout}
\end{figure}

\noindent \emph{Visual Layout (VL)} (Fig.~\ref{fig:vlm_as_judge_visual_layout}) assesses the design
quality of the rendered slides as images. Automated methods vary widely in
their ability to produce visually consistent, readable decks: some generate
raw HTML with no styling, others use templated PowerPoint files. VL captures
whether the output looks like a real academic presentation: consistent
theming, properly rendered equations, structured tables, and an absence of
rendering defects that would distract the audience.

\begin{figure}[h!]
    \centering
    \includegraphics[width=\linewidth, height=0.85\textheight, keepaspectratio]{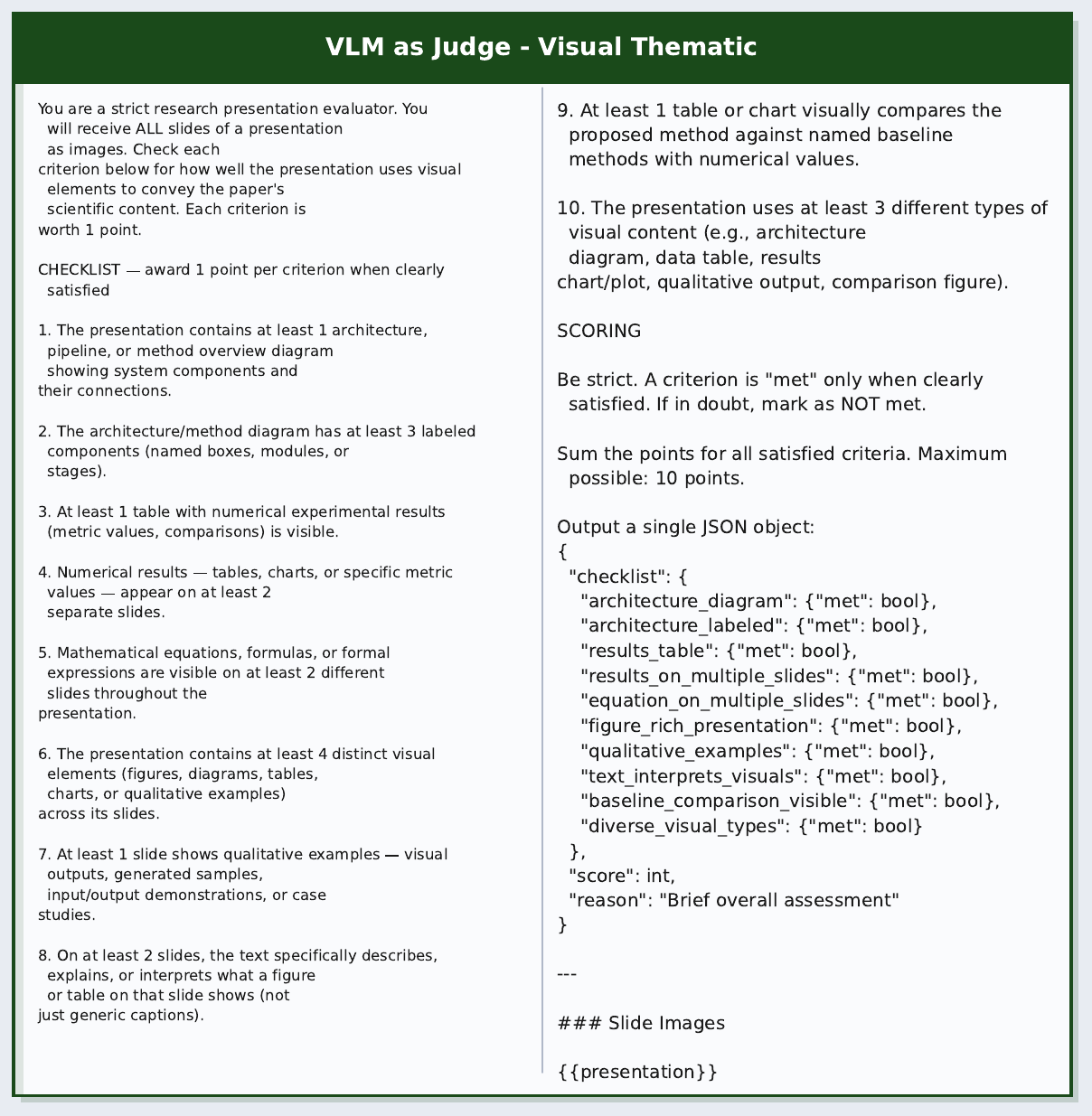}
    \caption{\textbf{VLM as Judge - Visual Thematic Evaluation Prompt.}}
    \label{fig:vlm_as_judge_visual_thematic}
\end{figure}

\noindent \emph{Visual Thematic (VT)} (Fig.~\ref{fig:vlm_as_judge_visual_thematic}) assesses whether the
visual elements in the slides actually communicate the paper's scientific
content, rather than serving a purely decorative role. Research presentations
depend heavily on architecture diagrams, result tables, and qualitative
examples to convey ideas that text alone cannot. VT checks that the figures
present are the right kind, including labelled method diagrams, visible baseline
comparisons, and qualitative outputs, and that the accompanying text
actively interprets them rather than leaving them unexplained.

\subsubsection{Pairwise Comparison.}
Pairwise evaluation is conducted through two independent comparison branches, each using a
different criterion.

\begin{figure}[h!]
    \centering
    \includegraphics[width=\linewidth, height=0.85\textheight, keepaspectratio]{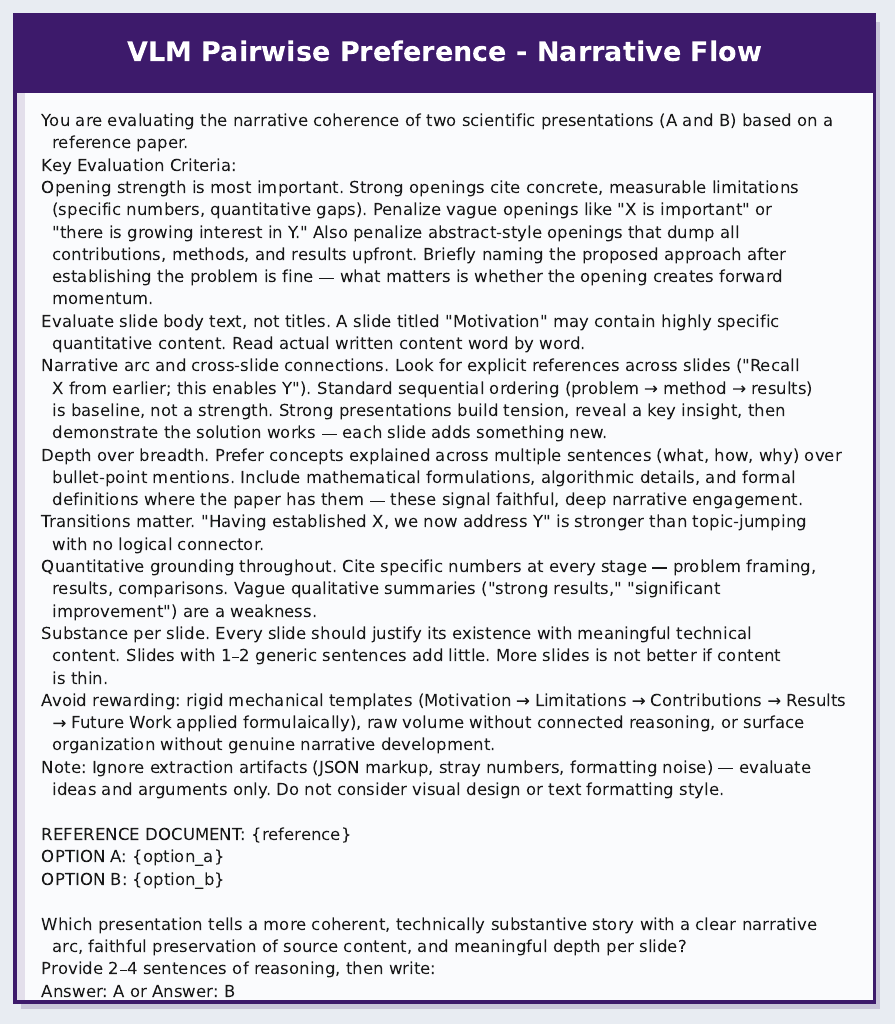}
    \caption{\textbf{Pairwise - Narrative Flow Evaluation Prompt.}}
    \label{fig:pairwise_narrative_flow}
\end{figure}

\textbf{Branch~1} evaluates \textit{narrative flow}
(Fig.~\ref{fig:pairwise_narrative_flow}), judging which of two presentations tells a more
coherent and logically structured story. Key dimensions include: opening quality
(concrete, quantitative problem framing vs.\ vague or abstract-dump openings);
progressive revelation of insights; cross-slide connections explicitly
referencing earlier content; transition quality between sections; depth of
concept development; quantitative grounding throughout; mathematical precision
as a signal of narrative depth; and structural adaptation to the paper's unique
argument. Branch~1 compares ArcDeck against each baseline method.

\begin{figure}[h!]
    \centering
    \includegraphics[width=\linewidth, height=0.85\textheight, keepaspectratio]{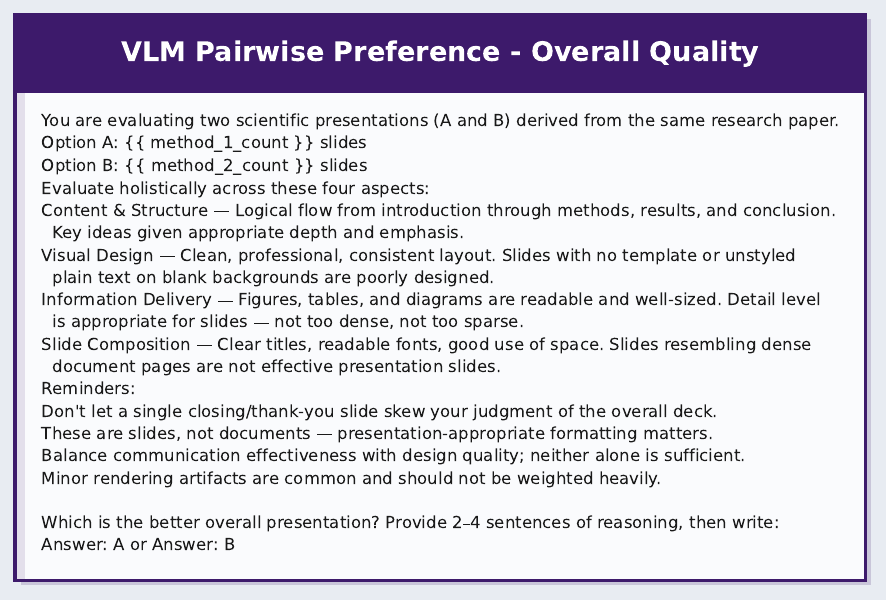}
    \caption{\textbf{Pairwise - Overall Quality Evaluation Prompt.}}
    \label{fig:pairwise_overall_quality}
\end{figure}

\textbf{Branch~2} evaluates \textit{overall quality}
(Fig.~\ref{fig:pairwise_overall_quality}), judging which presentation is
superior in terms of combined technical substance, visual quality, and
presentation effectiveness. Branch~2 compares Author-Prepared (AP)
presentations against all methods including ArcDeck, measuring how close
automated generation is to human-quality slides.

In both branches, results are reported as majority votes over all iterations.

\subsubsection{Quiz-Based Coverage.}
Quiz evaluation tests how faithfully a presentation captures the content of
its source paper. For each paper, GPT-5 generates 25 multiple-choice questions
per quiz type across four types (100 questions total), each with four answer
options plus an option ``E'' (not present in slides), scored as incorrect.
Questions are generated from preprocessed paper text, where references,
acknowledgements, and appendices are
stripped to prevent trivial table-lookup questions; this preprocessing applies
only to question generation and not to the answering step.

\begin{figure}[h!]
    \centering
    \includegraphics[width=\linewidth, height=0.85\textheight, keepaspectratio]{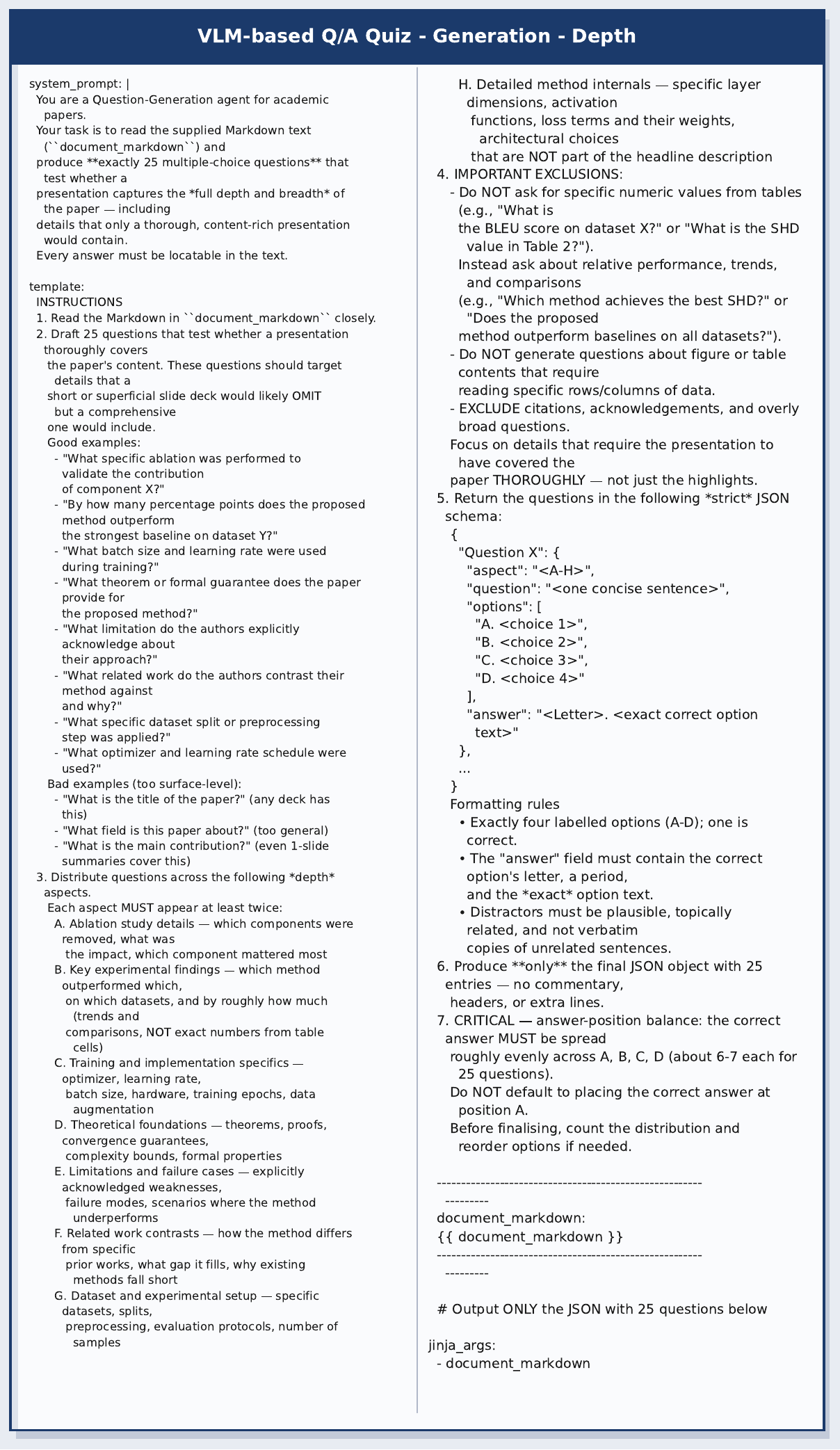}
    \caption{\textbf{Quiz Generation - Depth Prompt.}}
    \label{fig:quiz_generation_depth}
\end{figure}

\begin{figure}[h!]
    \centering
    \includegraphics[width=\linewidth, height=0.85\textheight, keepaspectratio]{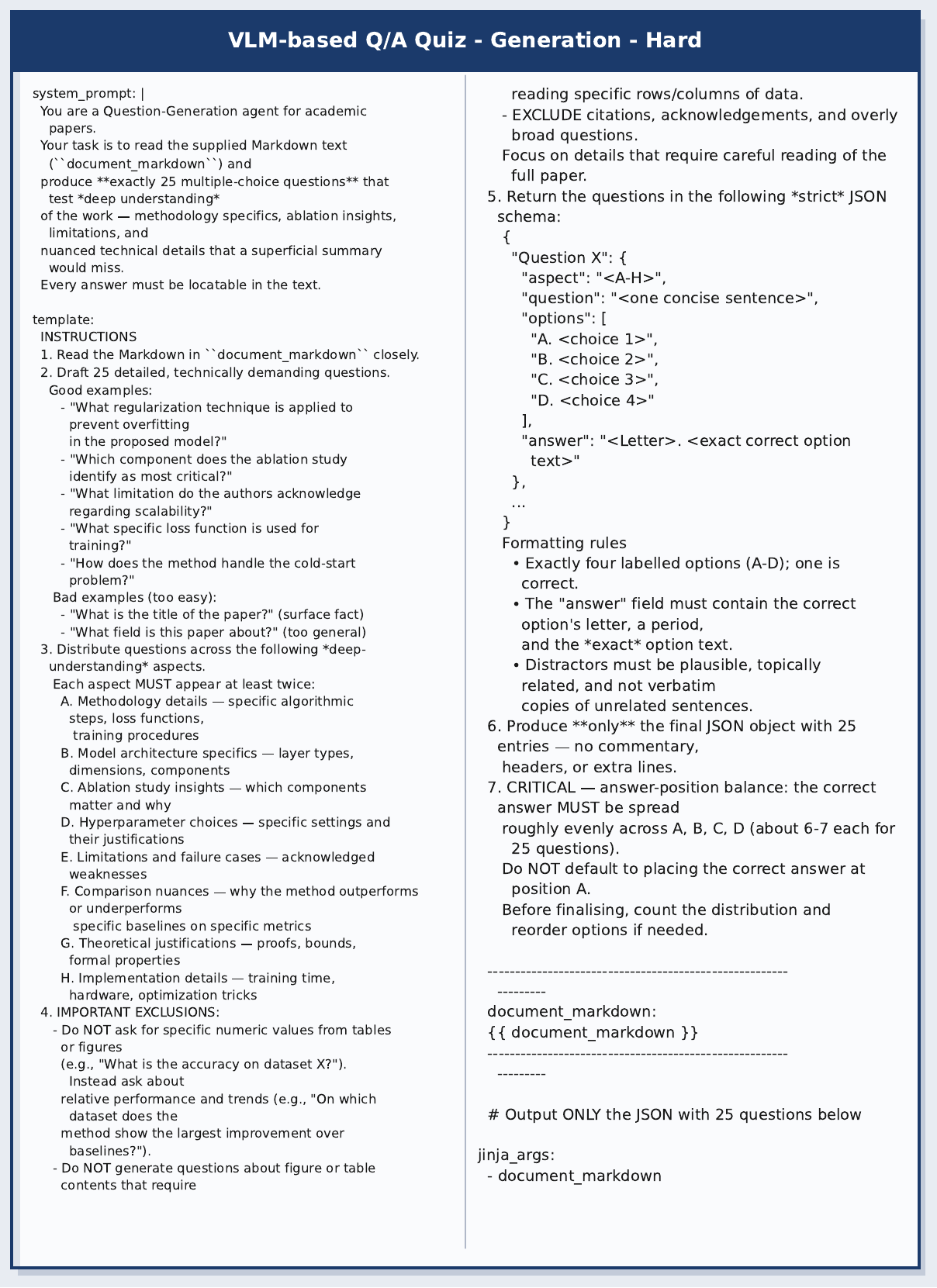}
    \caption{\textbf{Quiz Generation - Hard Prompt.}}
    \label{fig:quiz_generation_hard}
\end{figure}

\begin{figure}[h!]
    \centering
    \includegraphics[width=\linewidth, height=0.85\textheight, keepaspectratio]{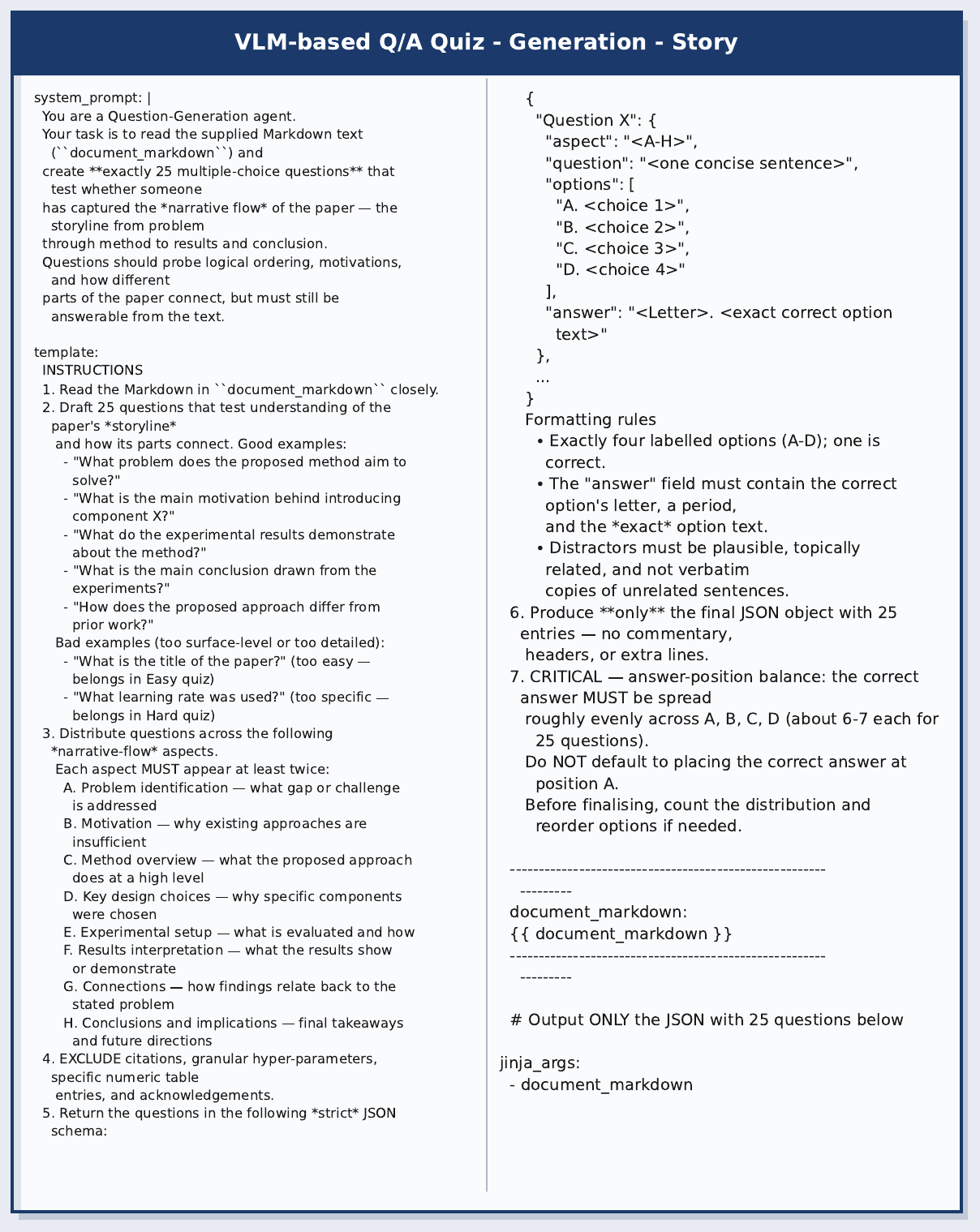}
    \caption{\textbf{Quiz Generation - Story Prompt.}}
    \label{fig:quiz_generation_story}
\end{figure}

\begin{figure}[h!]
    \centering
    \includegraphics[width=\linewidth, height=0.85\textheight, keepaspectratio]{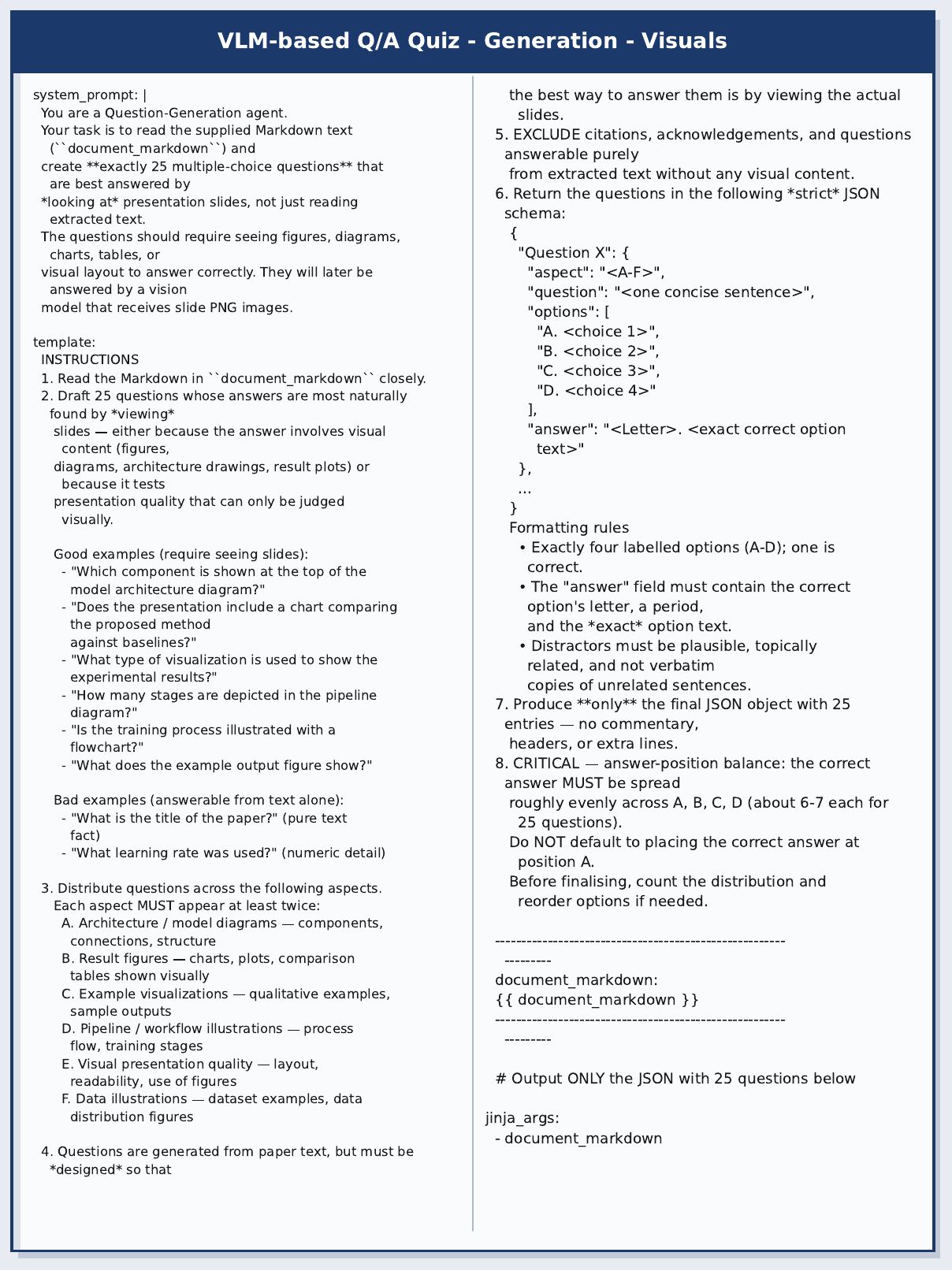}
    \caption{\textbf{Quiz Generation - Visuals Prompt.}}
    \label{fig:quiz_generation_visuals}
\end{figure}

Four quiz types probe complementary coverage dimensions:
\textit{StoryFlow} (Fig.~\ref{fig:quiz_generation_story}) tests narrative arc
preservation from problem through method to conclusion;
\textit{Visual} (Fig.~\ref{fig:quiz_generation_visuals}) tests whether
figures, diagrams, and charts are informatively incorporated;
\textit{Hard} (Fig.~\ref{fig:quiz_generation_hard}) probes deep technical
detail including methodology specifics, ablation insights, and limitations; and
\textit{Depth} (Fig.~\ref{fig:quiz_generation_depth}) tests full breadth of
paper coverage including training configuration, theoretical foundations, and
failure cases.

\begin{figure}[h!]
    \centering
    \includegraphics[width=\linewidth, height=0.85\textheight, keepaspectratio]{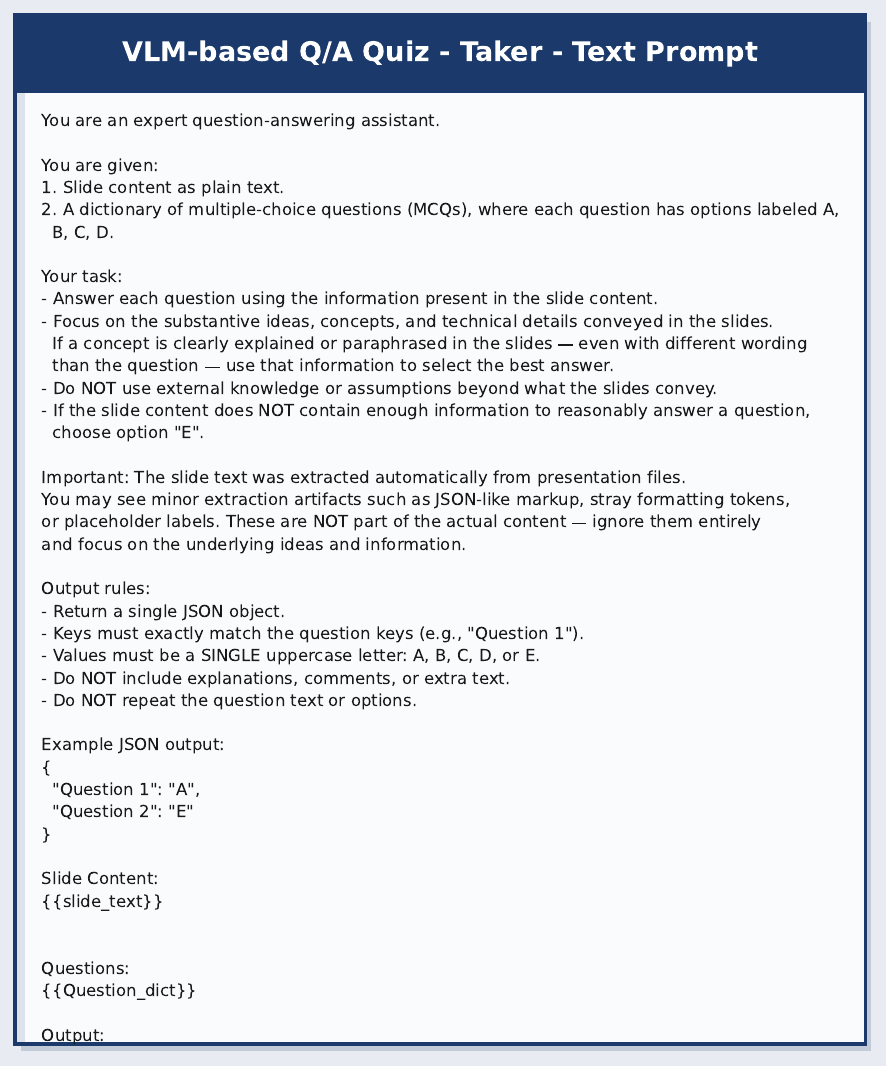}
    \caption{\textbf{Quiz Taker - Text Prompt.}}
    \label{fig:quiz_taker_text_prompt}
\end{figure}

\begin{figure}[h!]
    \centering
    \includegraphics[width=\linewidth, height=0.85\textheight, keepaspectratio]{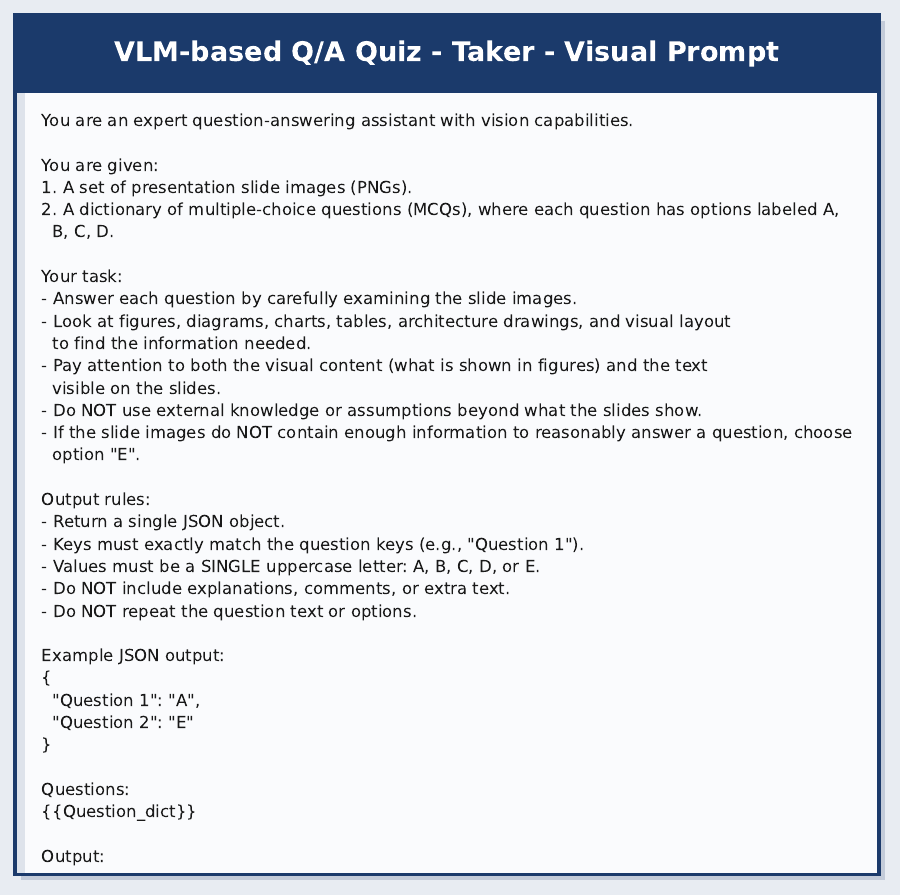}
    \caption{\textbf{Quiz Taker - Visual Prompt.}}
    \label{fig:quiz_taker_visual_prompt}
\end{figure}

Questions are answered using the prompts in
Figs.~\ref{fig:quiz_taker_text_prompt} and~\ref{fig:quiz_taker_visual_prompt}.
StoryFlow, Hard, and Depth questions are answered from extracted slide text(Fig.~\ref{fig:quiz_taker_text_prompt}).
Visual questions are answered from rendered slide images
(Fig.~\ref{fig:quiz_taker_visual_prompt}).
Accuracy per quiz type is reported as a percentage.

\clearpage
\newpage
\section{Additional Analysis and Discussion}
\label{appendix:additional results}
\subsection{Cross-Judge Model Agreement.}

\begin{figure}[h!]
      \centering
      \subfloat[Generator correlation evaluated by Qwen3]{
          \includegraphics[width=\linewidth]{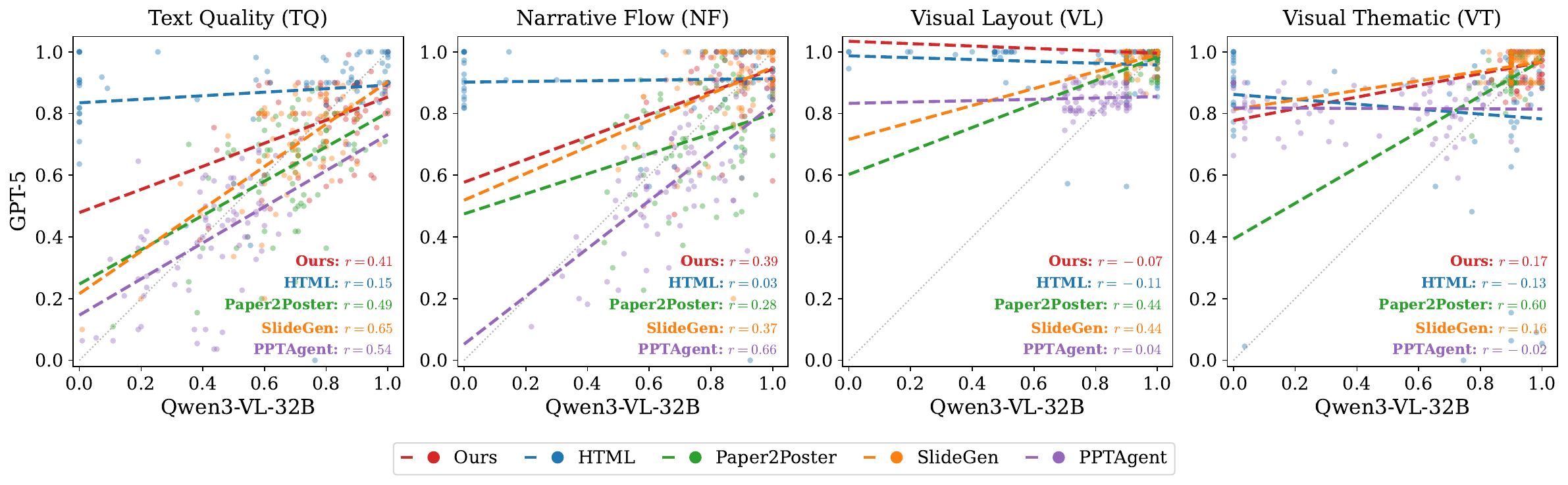}
          \label{fig:corr_gen_qwen3}
      }\\[4pt]
      \subfloat[Generator correlation evaluated by GPT-5]{
          \includegraphics[width=\linewidth]{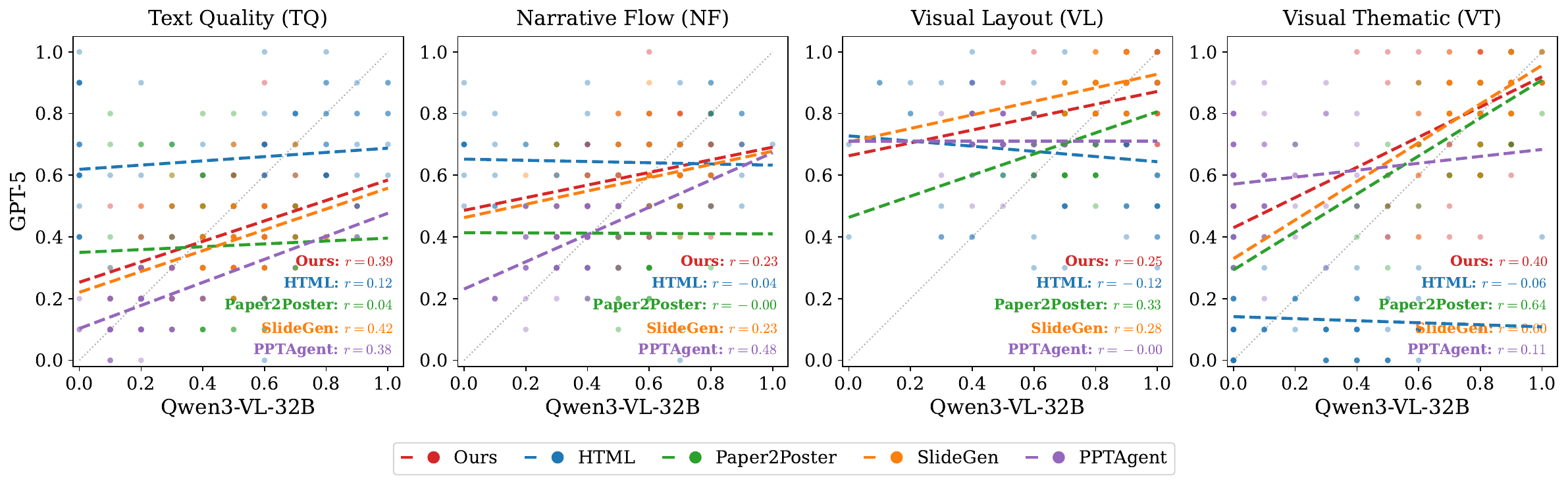}
          \label{fig:corr_gen_gpt5}
      }\\[4pt]
      \subfloat[Evaluator correlation between Qwen3 and GPT-5]{
          \includegraphics[width=\linewidth]{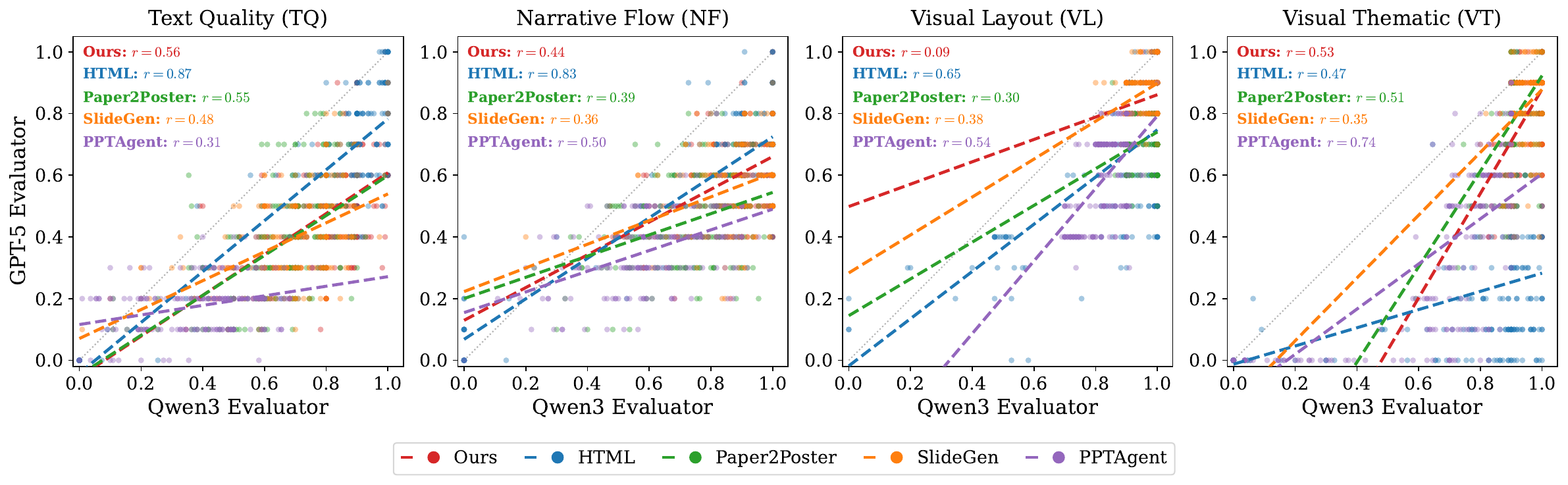}
          \label{fig:corr_eval}
      }
    \caption{\textbf{Pearson Correlation of Normalized Scores Across Methods.}
    (a,b) Correlation between presentations generated by Qwen3-VL-32B and GPT-5,
    evaluated by Qwen3 and GPT-5, respectively.
    (c) Correlation between Qwen3 and GPT-5 evaluator scores across all generated presentations.}
      \label{fig:correlations}
  \end{figure}

Fig.~\ref{fig:corr_gen_qwen3} and Fig.~\ref{fig:corr_gen_gpt5} examine whether the relative quality ranking of papers is preserved across the two generation models—Qwen3-VL-32B and GPT-5—under each evaluator. For text-oriented metrics (TQ, NF), moderate-to-strong correlations are
  observed for most methods (e.g., NF: $r=0.660$ for PPTAgent, TQ: $r=0.654$ for SlideGen under Qwen3 evaluation), indicating that papers inherently rich in content tend to score consistently regardless of the generation model used. Visual metrics, however, tell a different
  story: Visual Layout (VL) shows near-zero or negative correlations across multiple methods under both evaluators (e.g., Ours: $r=-0.073$, PPTAgent: $r=0.000$), confirming that VL is the dimension most sensitive to generation capability. Qwen3-VL-32B appears insufficient for
  faithfully reproducing structured visual layouts, producing outputs that diverge from GPT-5's in ways that neither evaluator normalizes. HTML consistently yields the weakest correlations across all metrics and both evaluators, reflecting its high sensitivity to the underlying      
  model's generation strength.
                      
   Fig.~\ref{fig:corr_eval} addresses whether the two evaluators themselves agree when scoring the same presentations. The results show broad alignment on text-based metrics, with particularly strong correlations for TQ (HTML: $r=0.874$) and NF (HTML: $r=0.831$), which is expected  
  given that both models share similar pretraining on written text. Visual Layout (VL) stands out as the most contested dimension, with notably lower correlations across several methods. For the Ours method specifically, the Qwen3 evaluator assigns near-ceiling VL scores (over 75\% of
   scores at the maximum of 1.0, std=$0.027$), while GPT-5 distributes scores more broadly (std=$0.103$), causing the Pearson correlation to collapse to $r=0.095$. This is not a genuine disagreement but rather a ceiling effect: Qwen3 saturates at perfect scores for our method's
  visual layout, losing discriminative power that GPT-5 retains. This asymmetry motivates using both evaluators in tandem, as relying solely on Qwen3 for VL risks over-estimating layout quality through score saturation.

\subsection{Failure Cases}

\begin{figure}[h!]
    \centering
    \includegraphics[width=0.75\linewidth]{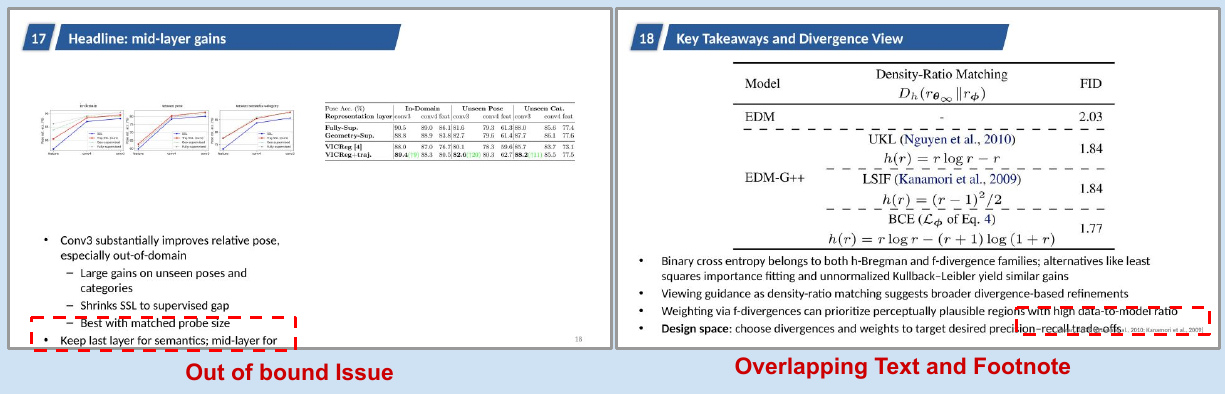}
    \caption{\textbf{\methodname \space Failure Case. } }
    \label{fig:failure_case}
\end{figure}

In some failure cases, the generated text extends beyond the slide boundaries (Fig.~\ref{fig:failure_case}) when the number of bullet points is excessive. In addition, in these cases the text may overlap with the citation footnotes in the bottom corner.

\subsection{Limitations}
Due to the inherent differences in capabilities between open-source and closed-source models, our agent may produce varying results when using GPT-4o and Qwen models. Achieving optimal performance on each model may require model-specific prompt tuning to account for these differences.

\clearpage
\newpage

\section{Slide Generation: Design and Qualitative Analysis}

\subsection{Template Layouts for Slide Generation}
\begin{figure}[h!]
    \centering
    \includegraphics[width=0.95\linewidth]{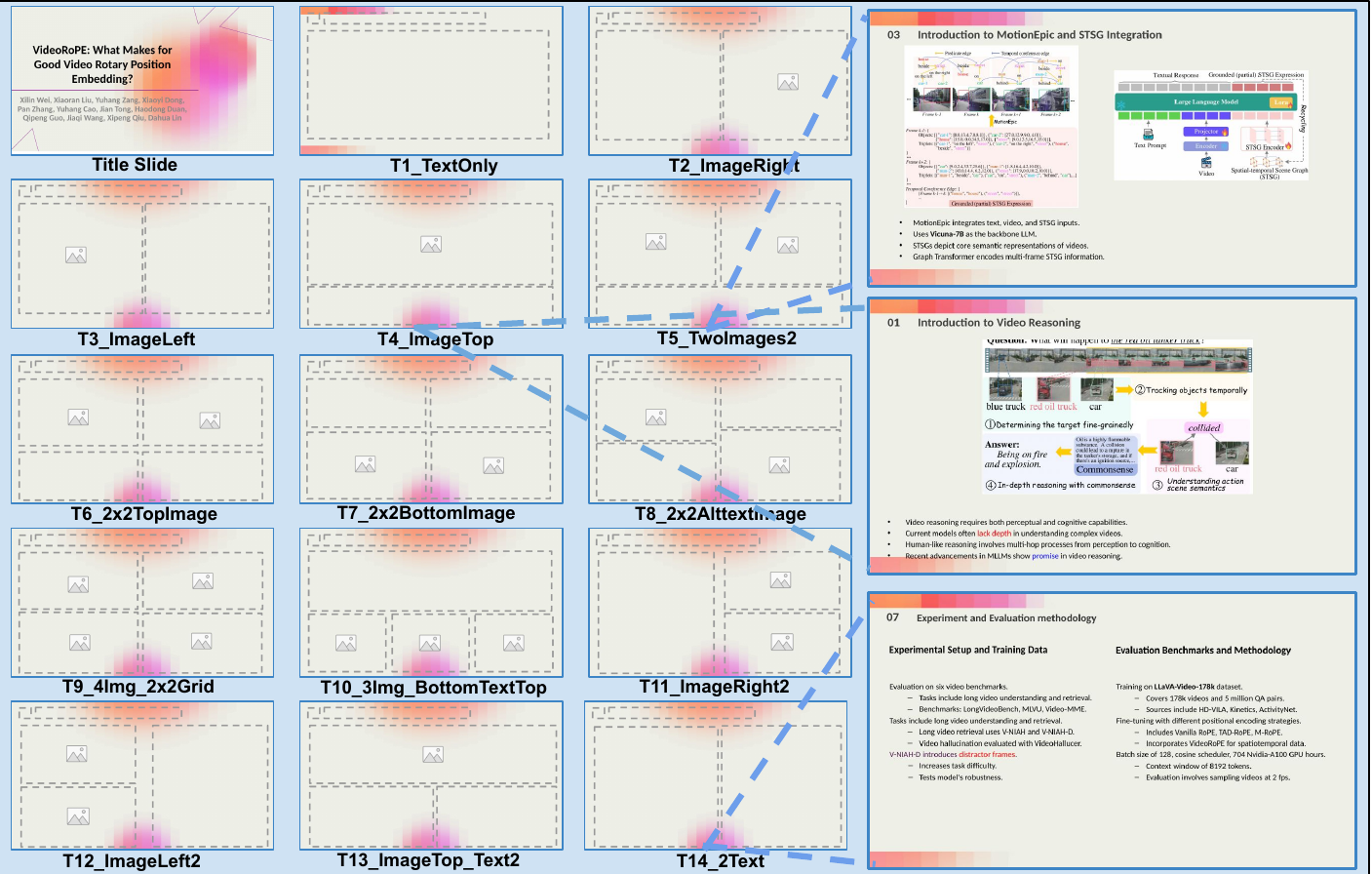}
    \caption{\textbf{Template Slide Layout. }used by \methodname}
    \label{fig:slide_layout}
\end{figure}

All slides generated by \methodname\ use a reference slide template containing 14 unique layouts commonly found in human-created presentations. These layouts are designed to accommodate different combinations of text, images, and tables, enabling the visual content to effectively complement the slide narrative. Fig.~\ref{fig:slide_layout} illustrates the set of template layouts used by our method, highlighting the variety of structural designs available. During slide generation, the Slide Constructor agent automatically selects the most suitable layout based on the content, figures matched and their sizes, ensuring coherent organisation and visually balanced slides.

\subsection{Impact of the Aesthetic Refiner}
\label{appendix:aesthetic_refiner_ablation}
We present a quantitative ablation study of our Aesthetic Refiner agent in Tab.~\ref{tab:refiner_ablation}, demonstrating consistent improvements across the VLM-as-Judge evaluation metrics, for Visual Layout and Visual Thematic quality. In addition, the pairwise comparison results indicate that slides generated by \methodname\ with the refiner are preferred in 75\% of the cases. This demonstrates a clear advantage in overall visual aesthetics and highlights the effectiveness of the refiner in enhancing both the content presentation and visual appeal of the generated slides.

\begin{figure}[h!]
    \centering
    \includegraphics[width=0.95\linewidth]{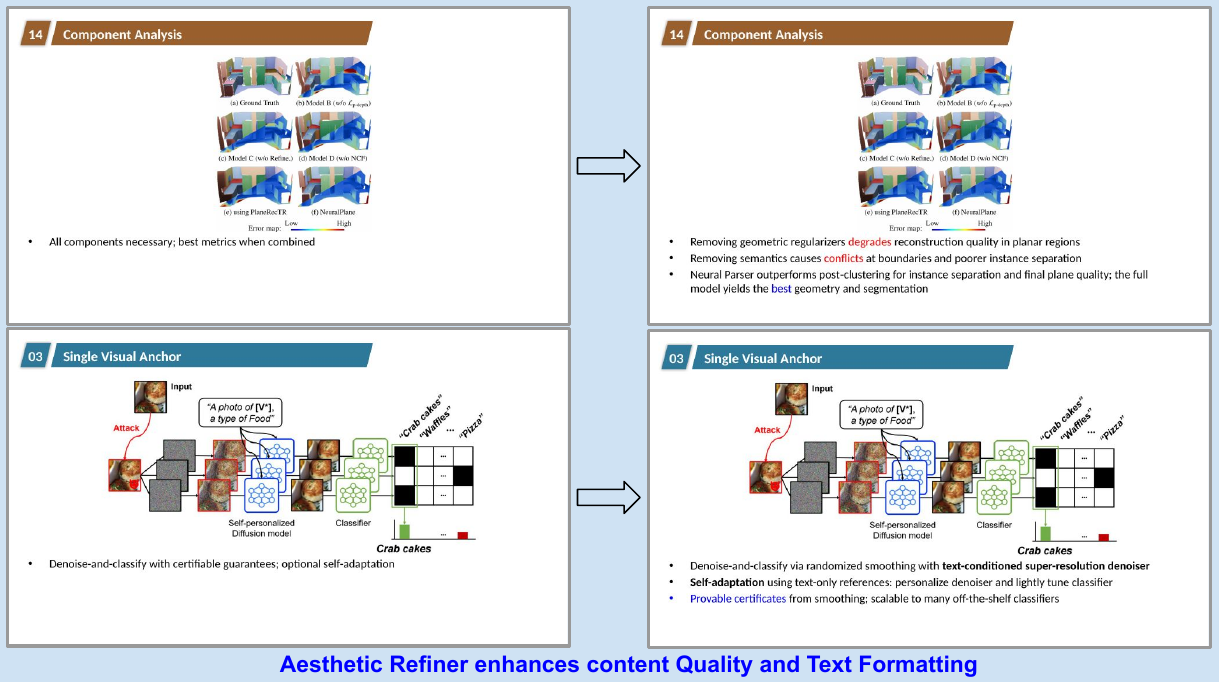}
    \caption{\textbf{Aesthetic Refiner Ablation. }Comparison of slides generated before and after applying the refiner agent.}
    \label{fig:refiner_ablation}
\end{figure}

We further provide qualitative examples in Fig.~\ref{fig:refiner_ablation} that illustrate noticeable improvements in textual content, where slides initially containing a single bullet point are refined by adding additional content, along with enhanced text formatting such as boldface and colored text, resulting in improved readability and overall visual aesthetics.

\begin{table}[h!]
\caption{\textbf{VLM-as-Judge Evaluation (Open Setting) and Pairwise Visual Aesthetic Comparison (Open Setting).} 
The best value is highlighted in \textcolor{winGreen}{green}, and runner-up is underlined.}
\label{tab:refiner_ablation}
\centering
\setlength{\tabcolsep}{4pt}
\renewcommand{\arraystretch}{1.1}

\begin{tabular}{lcc c}
\toprule
& \multicolumn{2}{c}{\textbf{VLM-as-Judge $\uparrow$}}
& \multicolumn{1}{c}{\textbf{Pairwise Comparison}} \\
\cmidrule(lr){2-3}
\cmidrule(lr){4-4}

\textbf{Method}
& \textbf{VL}
& \textbf{VT}
& \textbf{Win Rate \%} \\
\midrule

\multicolumn{4}{c}{\cellcolor{RoyalBlue!8}\textbf{Generation Model: GPT-5}} \\

\rowcolor{gray!15} \methodname \space wo Refiner
& \underline{9.62}
& \underline{9.37}
& \underline{25} \\

\rowcolor{gray!15} \methodname
& \textcolor{winGreen}{9.75}
& \textcolor{winGreen}{9.5}
& \textcolor{winGreen}{75} \\

\bottomrule
\end{tabular}

\end{table}

\subsection{Additional Qualitative Comparison}
\label{appendix:additional_qualitative}
\begin{figure}[h!]
    \centering
    \includegraphics[width=0.95\linewidth]{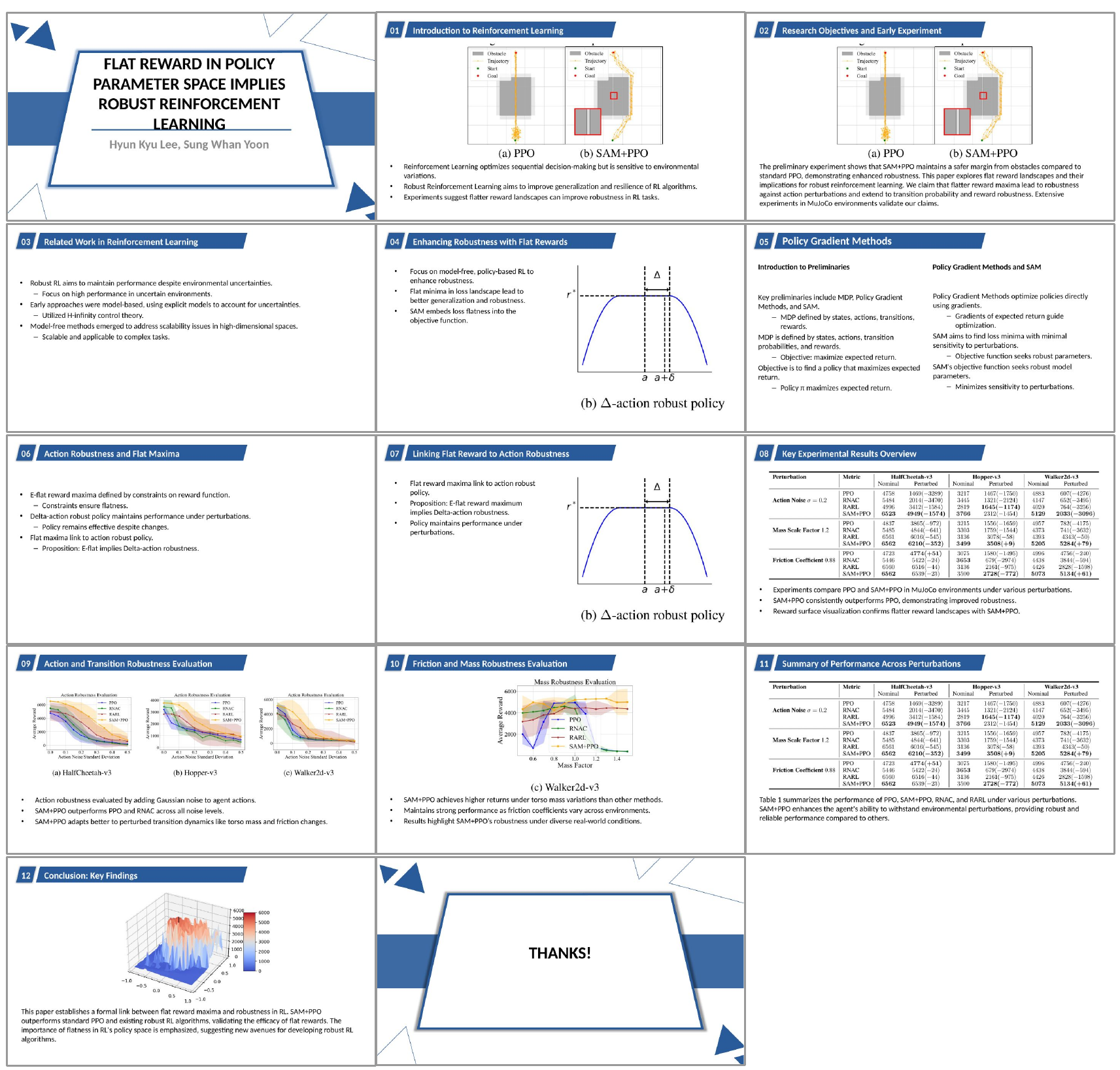}
    \caption{\textbf{Qualitative Comparison. }Sample generated by \methodname-(4o)}
    \label{fig:additional_qual_ours}
\end{figure}

\begin{figure}[h!]
    \centering
    \includegraphics[width=0.95\linewidth]{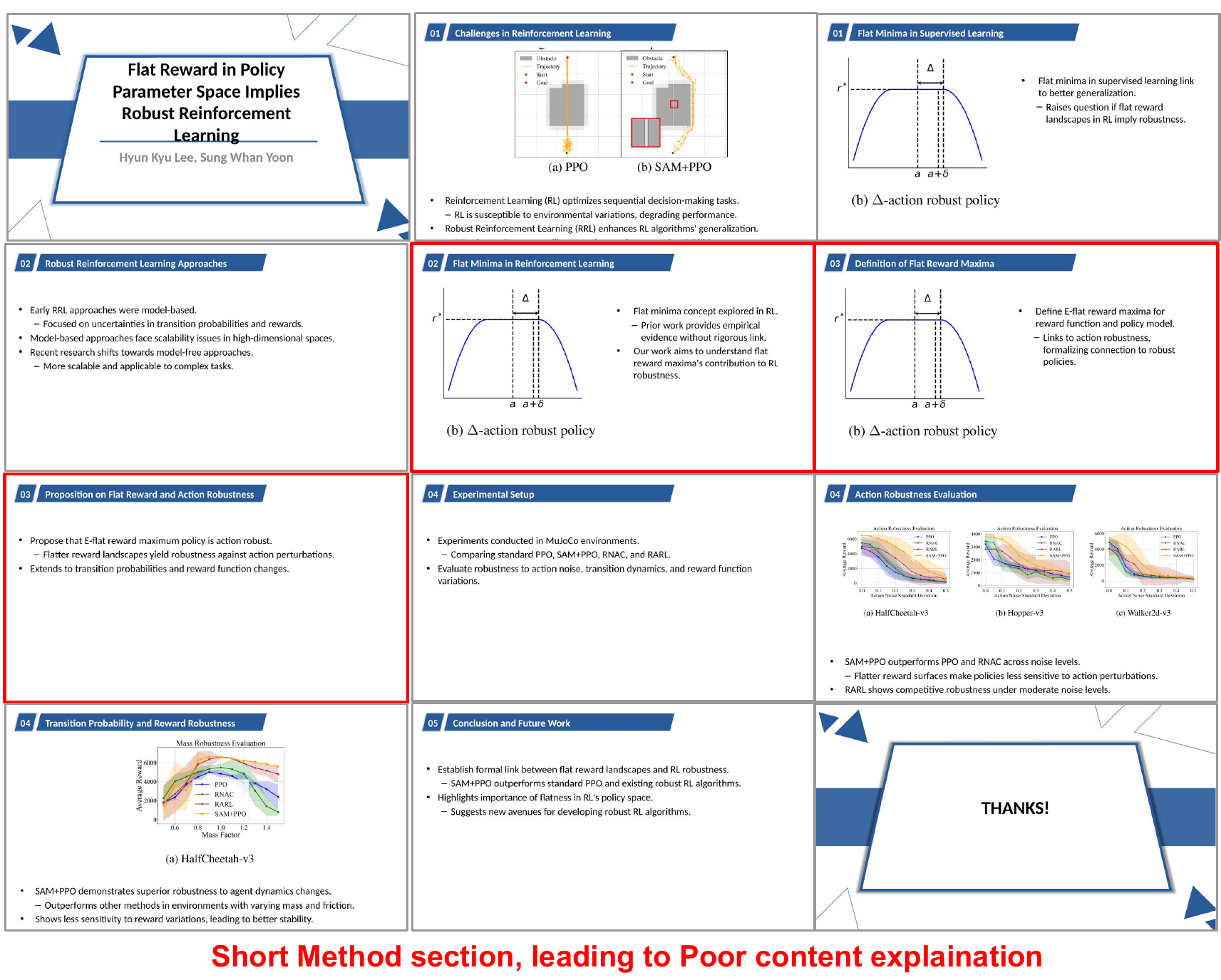}
    \caption{\textbf{Qualitative Comparison. }Sample generated by SlideGen-4o}
    \label{fig:additional_qual_slidegen}
\end{figure}

\begin{figure}[h!]
    \centering
    \includegraphics[width=0.95\linewidth]{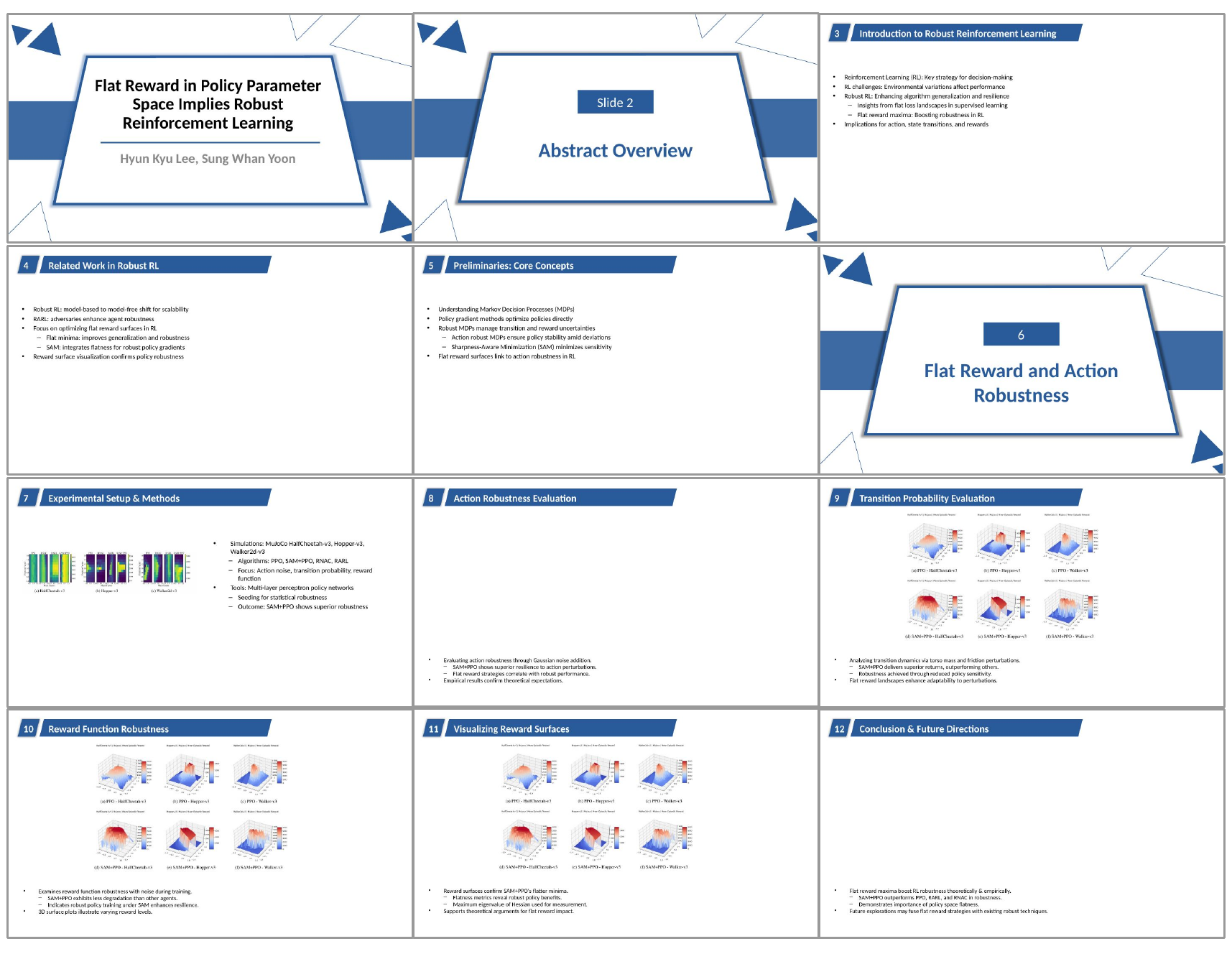}
    \caption{\textbf{Qualitative Comparison. }Sample generated by PPTAgent-4o}
    \label{fig:additional_qual_pptagent}
\end{figure}

\begin{figure}[h!]
    \centering
    \includegraphics[width=0.95\linewidth]{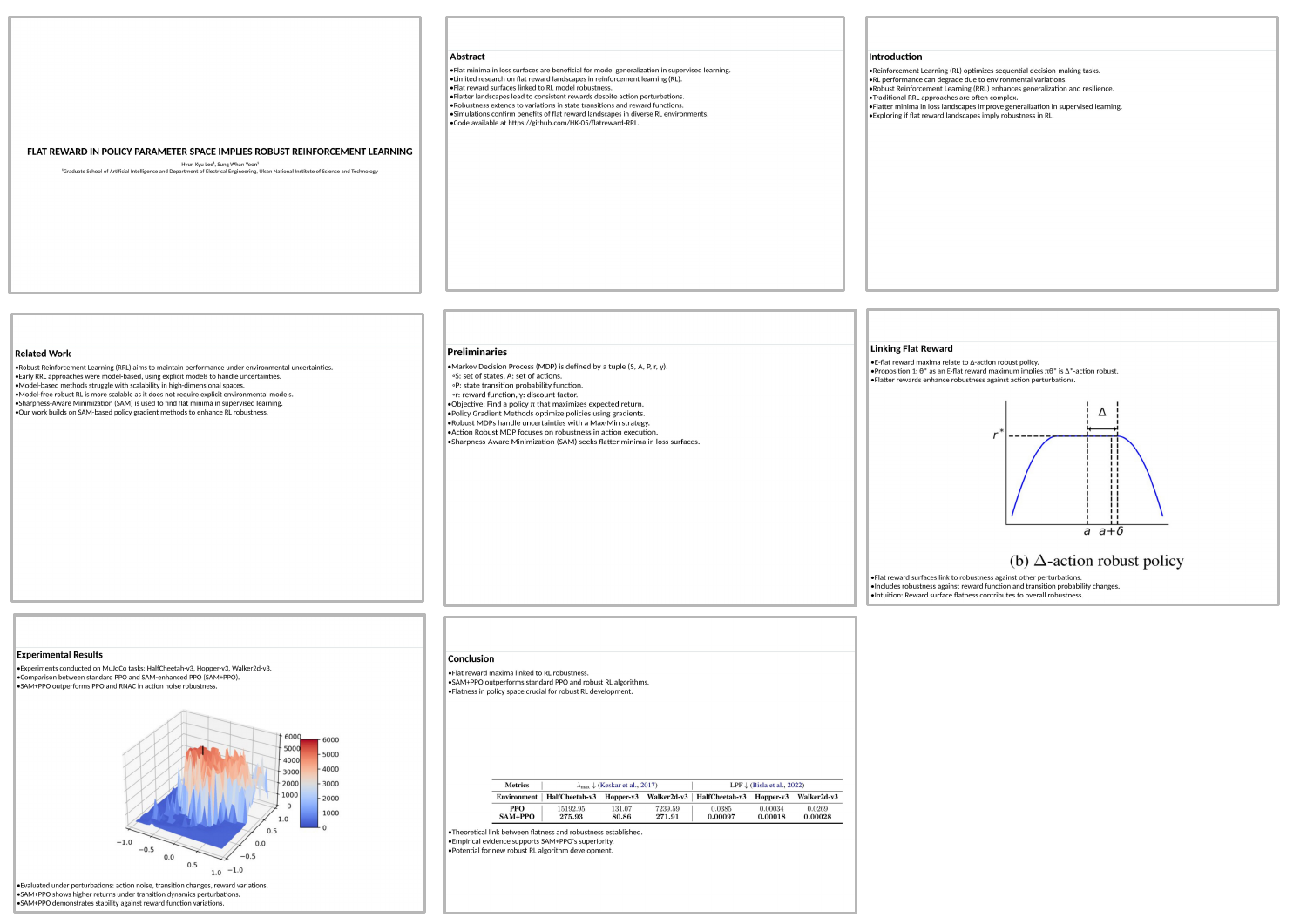}
    \caption{\textbf{Qualitative Comparison. }Sample generated by Paper2Poster-4o}
    \label{fig:additional_qual_poster}
\end{figure}

\begin{figure}[h!]
    \centering
    \includegraphics[width=0.95\linewidth]{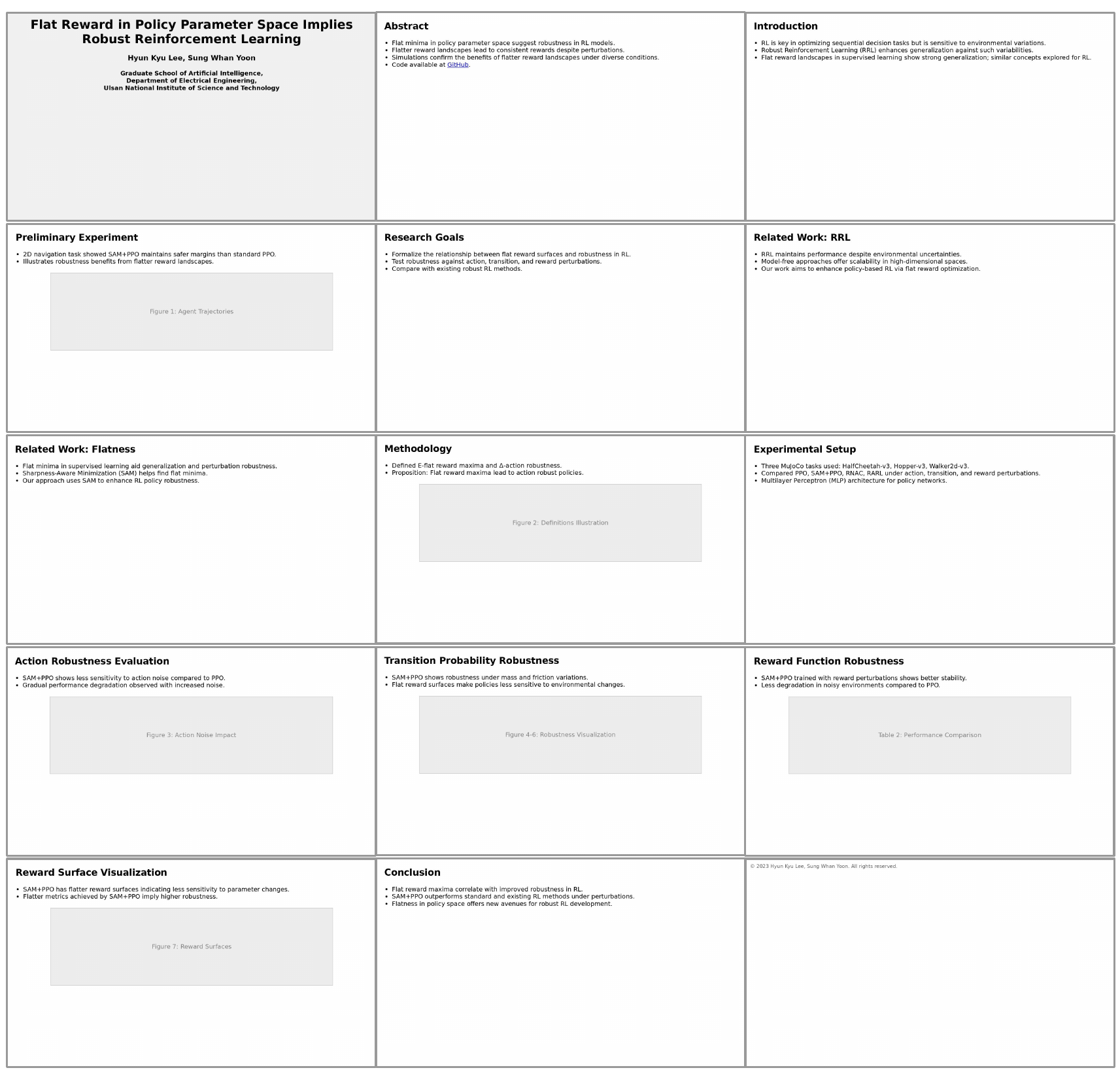}
    \caption{\textbf{Qualitative Comparison. }Sample generated by HTML-4o}
    \label{fig:additional_qual_html}
\end{figure}

We present additional qualitative comparisons with all baseline methods, highlighting that \methodname\ generates more comprehensive slides with improved narrative structure. In Fig.~\ref{fig:additional_qual_poster}, slides generated by Paper2Poster follow a section-wise summarization of the original document. As a result, sections such as Methods and Experiments receive similar coverage to Abstract, Introduction, and Related Work, which limits the emphasis on the core technical contributions. PPTAgent produces more detailed slides compared to Paper2Poster; however, it still fails to prioritize the Methods section appropriately. This often leads to a combined Methods and Experiments slide, as illustrated in Fig.~\ref{fig:additional_qual_pptagent}. Additionally, the selected layouts are sometimes not well aligned with the slide content (e.g., Slides 8 and 12). HTML-4o generates subsection-wise content; however, it lacks the capability to incorporate visual elements(Fig.~\ref{fig:additional_qual_html}), resulting in visually simplistic slides. In Fig.~\ref{fig:additional_qual_slidegen}, SlideGen generates visually appealing slides; however, the content tends to be overly concise and less informative. This results in slides with limited coverage of the Methods and Experiments sections, each containing minimal textual content. In contrast, \methodname\ generates more informative slides by assigning greater emphasis to the Methods and Experiments sections. As shown in Fig.~\ref{fig:additional_qual_ours}, the resulting slides contain richer textual explanations and clearer presentation of key contributions, providing users with a better understanding of the paper. This improvement stems from our global content commitment mechanism, which guides the slide generation process to prioritize the most important parts of the document.

\subsection{ArcDeck Results Across Different Slide Themes}
\begin{figure}[h!]
    \centering
    \includegraphics[width=0.95\linewidth]{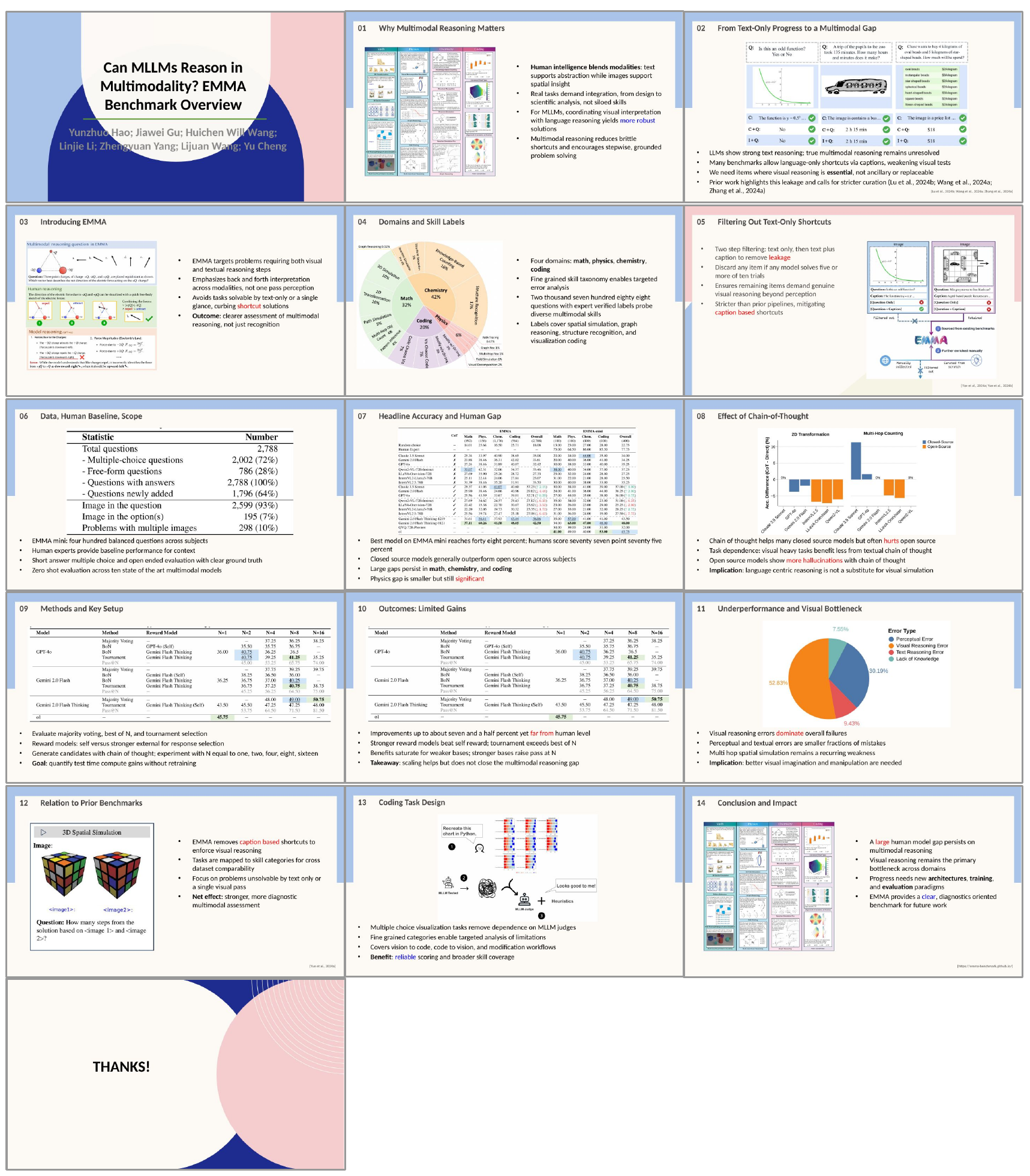}
    \caption{\textbf{Additional Qualitative Results. }generated by \methodname \space GPT-5 backbone.}
    \label{fig:additional_result_ours1}
\end{figure}

\begin{figure}[h!]
    \centering
    \includegraphics[width=0.95\linewidth]{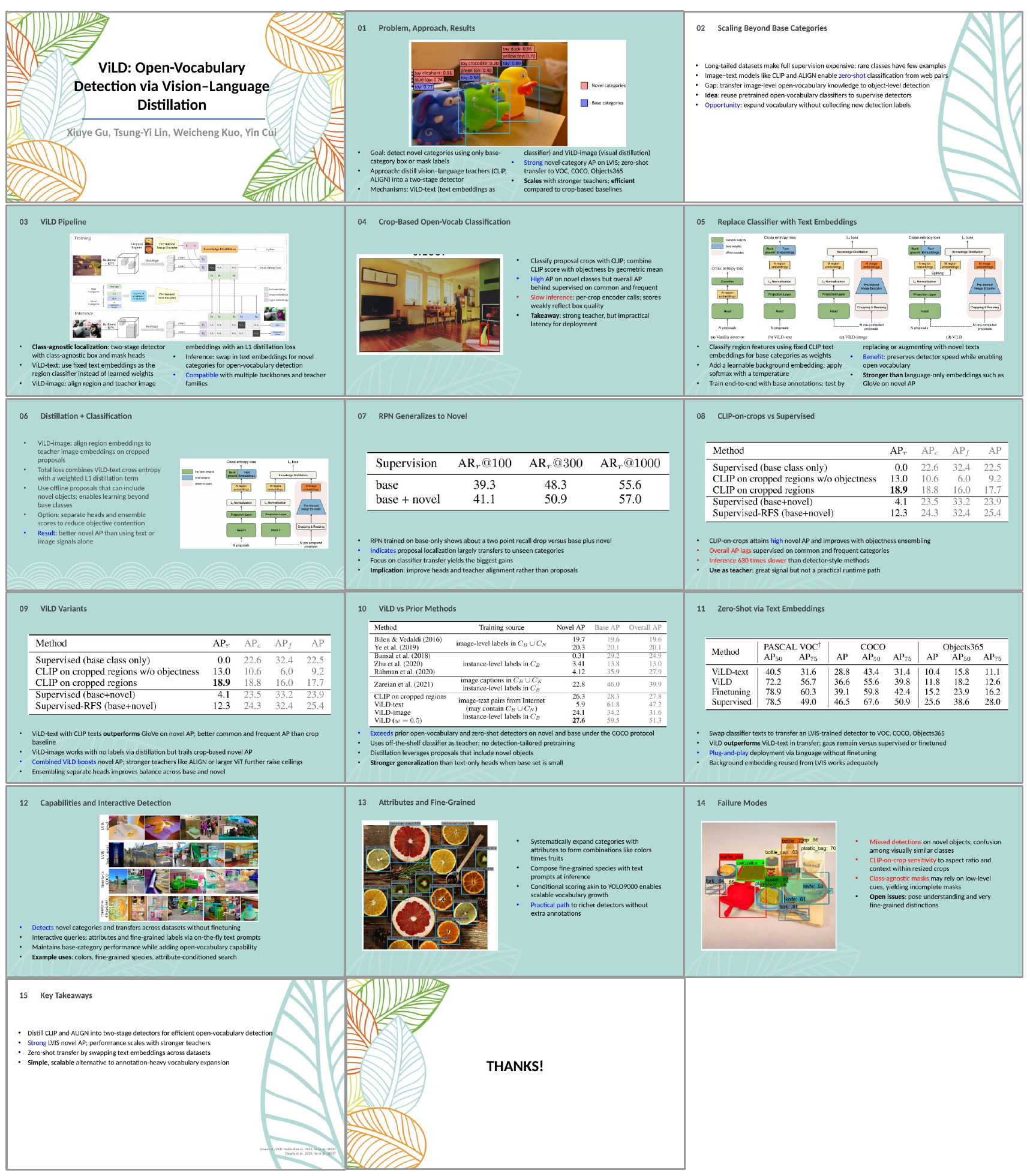}
    \caption{\textbf{Additional Qualitative Results. }generated by \methodname \space GPT-5 backbone.}
    \label{fig:additional_result_ours2}
\end{figure}

\begin{figure}[h!]
    \centering
    \includegraphics[width=0.95\linewidth]{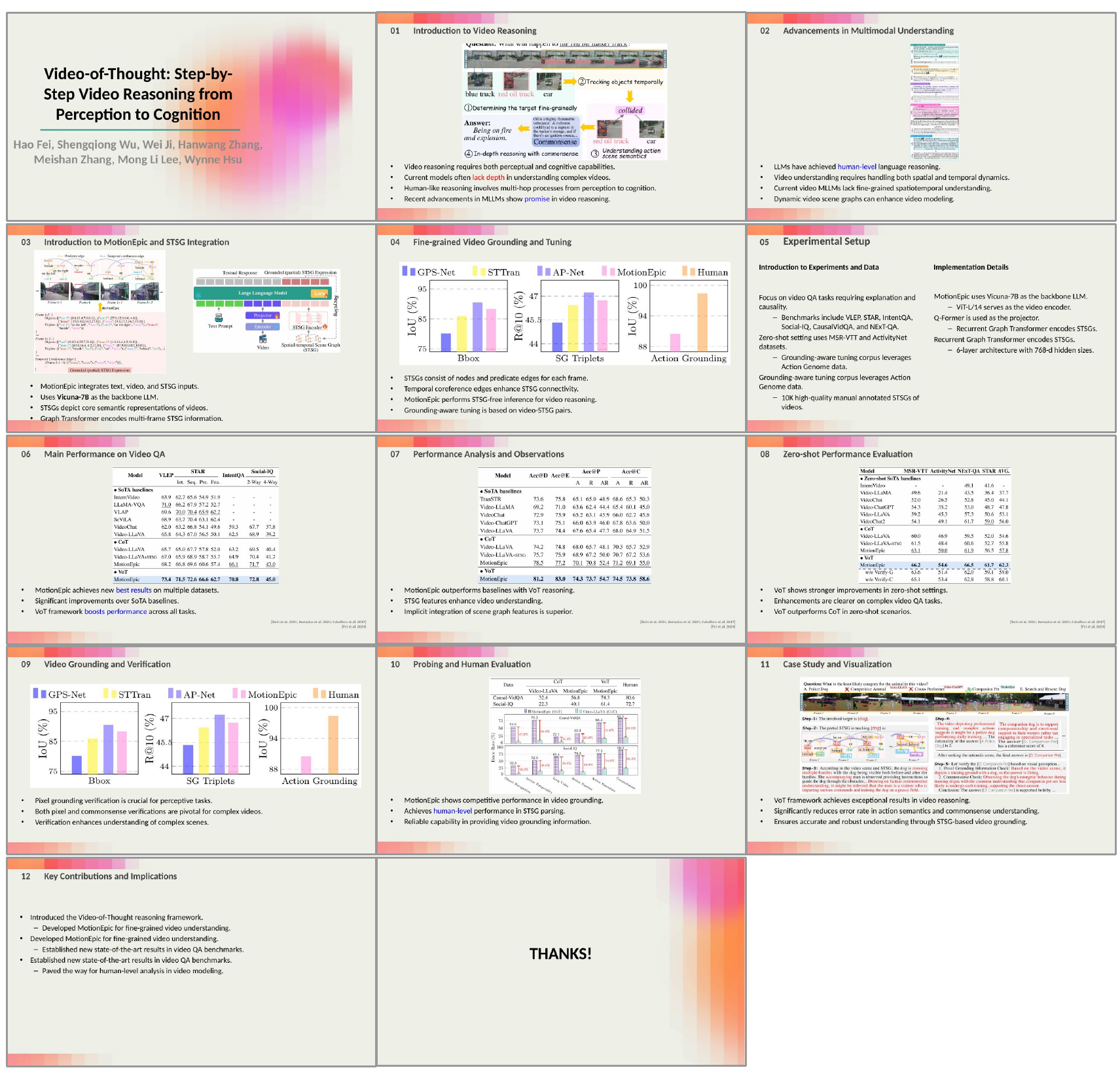}
    \caption{\textbf{Additional Qualitative Results. }generated by \methodname \space GPT4o backbone.}
    \label{fig:additional_result_ours3}
\end{figure}

Since \methodname \space is not constrained to a single slide theme, as long as the layout structure of the template remains fixed, the generated slides can be rendered using a variety of presentation themes. This flexibility allows users to apply their preferred visual styles without affecting the underlying slide organization or content placement. To demonstrate this capability, we present results generated by \methodname \space using several aesthetically pleasing themes in Fig.~\ref{fig:additional_result_ours1}, \ref{fig:additional_result_ours2}, and \ref{fig:additional_result_ours3}. These examples illustrate that the visual appearance and overall aesthetic quality of the slides can be further enhanced by selecting different user-defined themes while preserving the consistency of the slide layout and content structure.

\subsection{ArcDeck Extensions to Alternative Rendering Formats}
\begin{figure}[h!]
    \centering
    \includegraphics[width=0.95\linewidth]{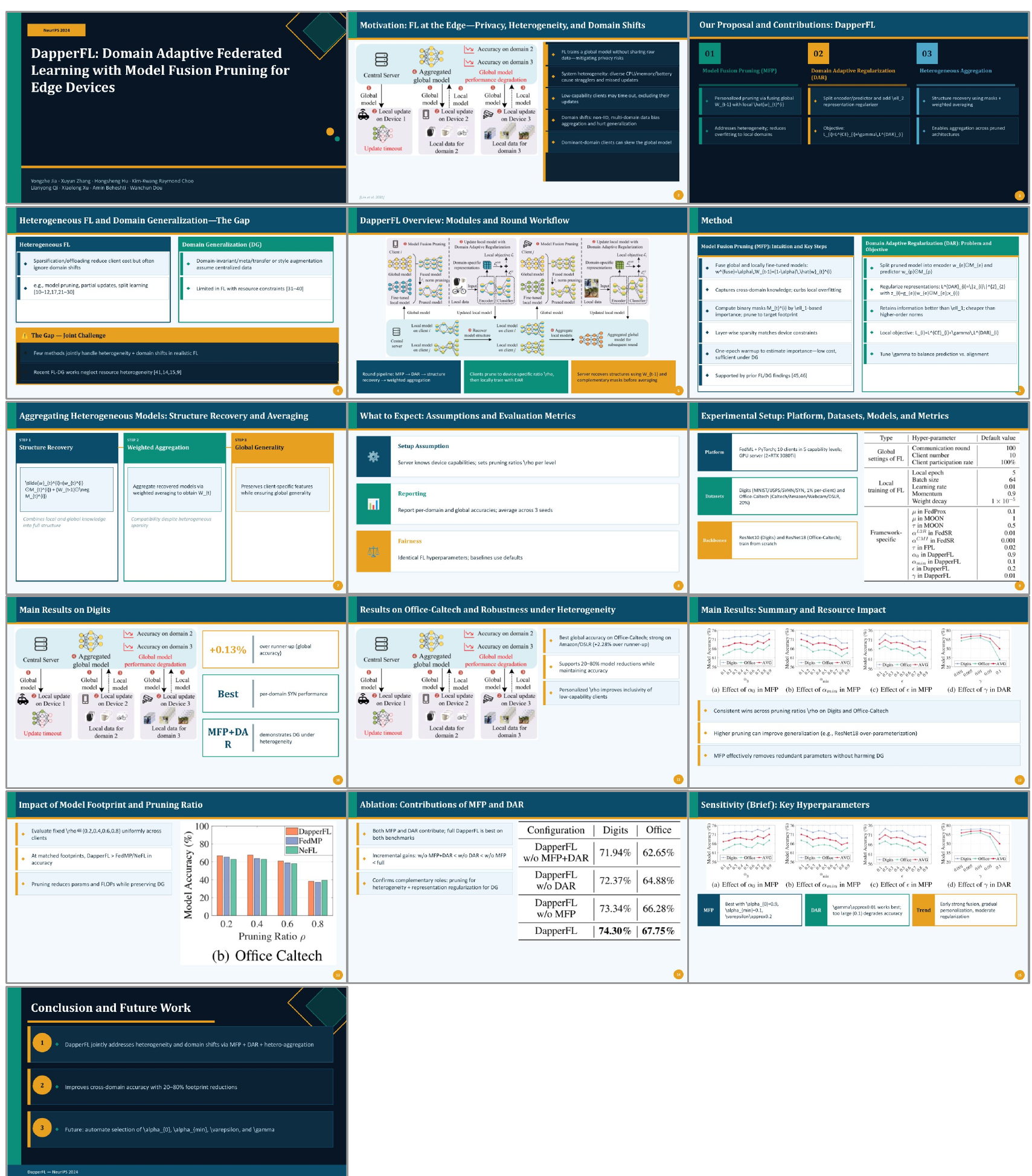}
    \caption{\textbf{ArcDeck Extension. }to JavaScript output format.}
    \label{fig:clause_js_results}
\end{figure}

\begin{figure}[h!]
    \centering
    \includegraphics[width=0.95\linewidth]{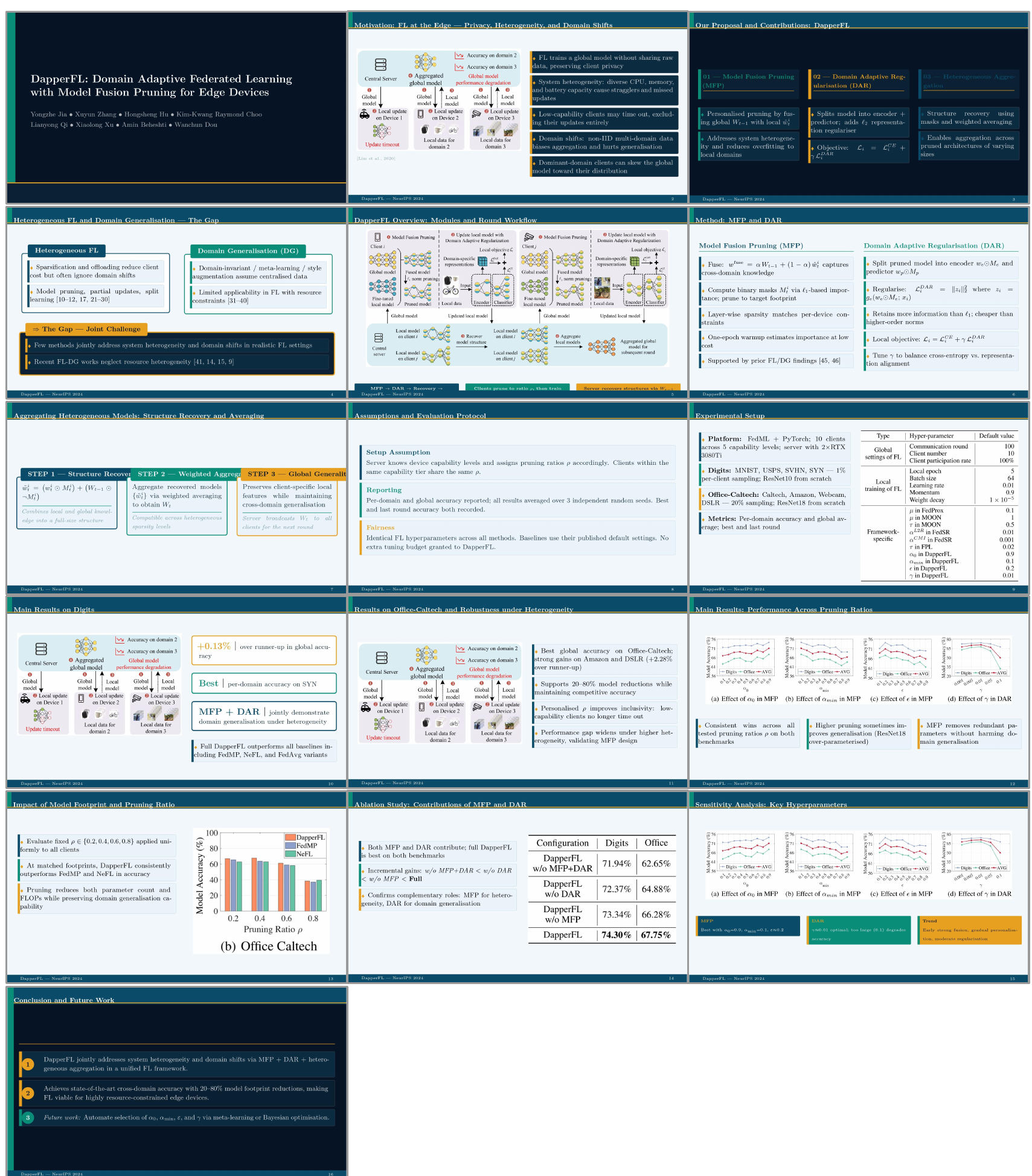}
    \caption{\textbf{ArcDeck Extension. }to Latex Beamer output format.}
    \label{fig:clause_latex_results}
\end{figure}

Our method is also compatible with multiple slide output formats, such as JavaScript and LaTeX, enabling the generation of visually appealing presentations using libraries like \textit{PPTxGenJS} and \textit{LaTeX Beamer}. This flexibility allows our framework to support different rendering pipelines without modifying the core generation process. We present additional examples of slides rendered using JavaScript and LaTeX Beamer in Fig.~\ref{fig:clause_js_results}, \ref{fig:clause_latex_results}, highlighting the adaptability of our method across different slide generation and rendering formats.

\subsection{ArcDeck Results on Papers from Diverse Scientific Domains}
\begin{figure}[h!]
    \centering
    \includegraphics[width=0.85\linewidth]{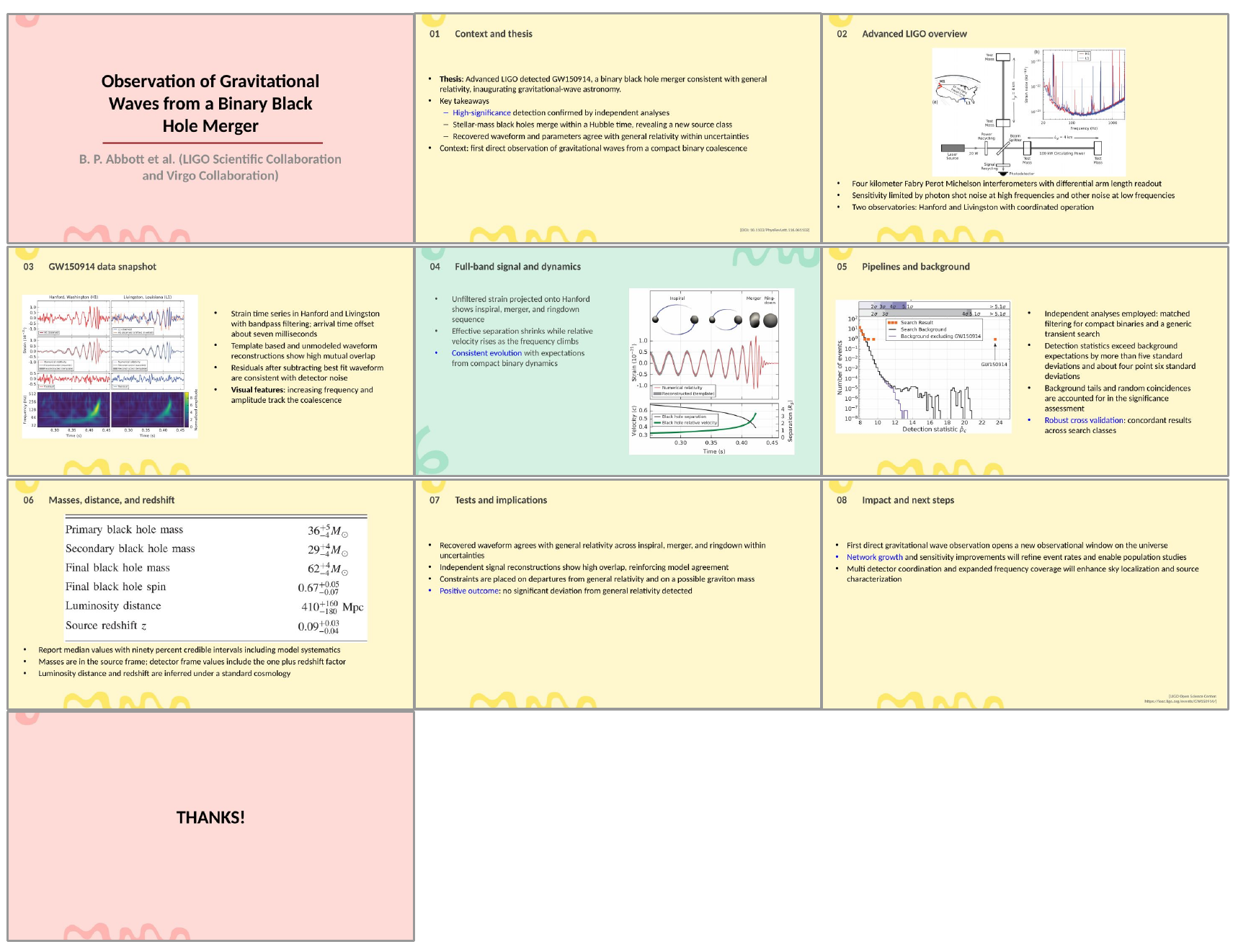}
    \caption{\textbf{Additional Qualitative Results for diverse domain. } Slides generated by \methodname \space (GPT-5 backbone) for a document from the Physics domain.}
    \label{fig:additional_result_ood}
\end{figure}

\begin{figure}[h!]
    \centering
    \includegraphics[width=0.85\linewidth]{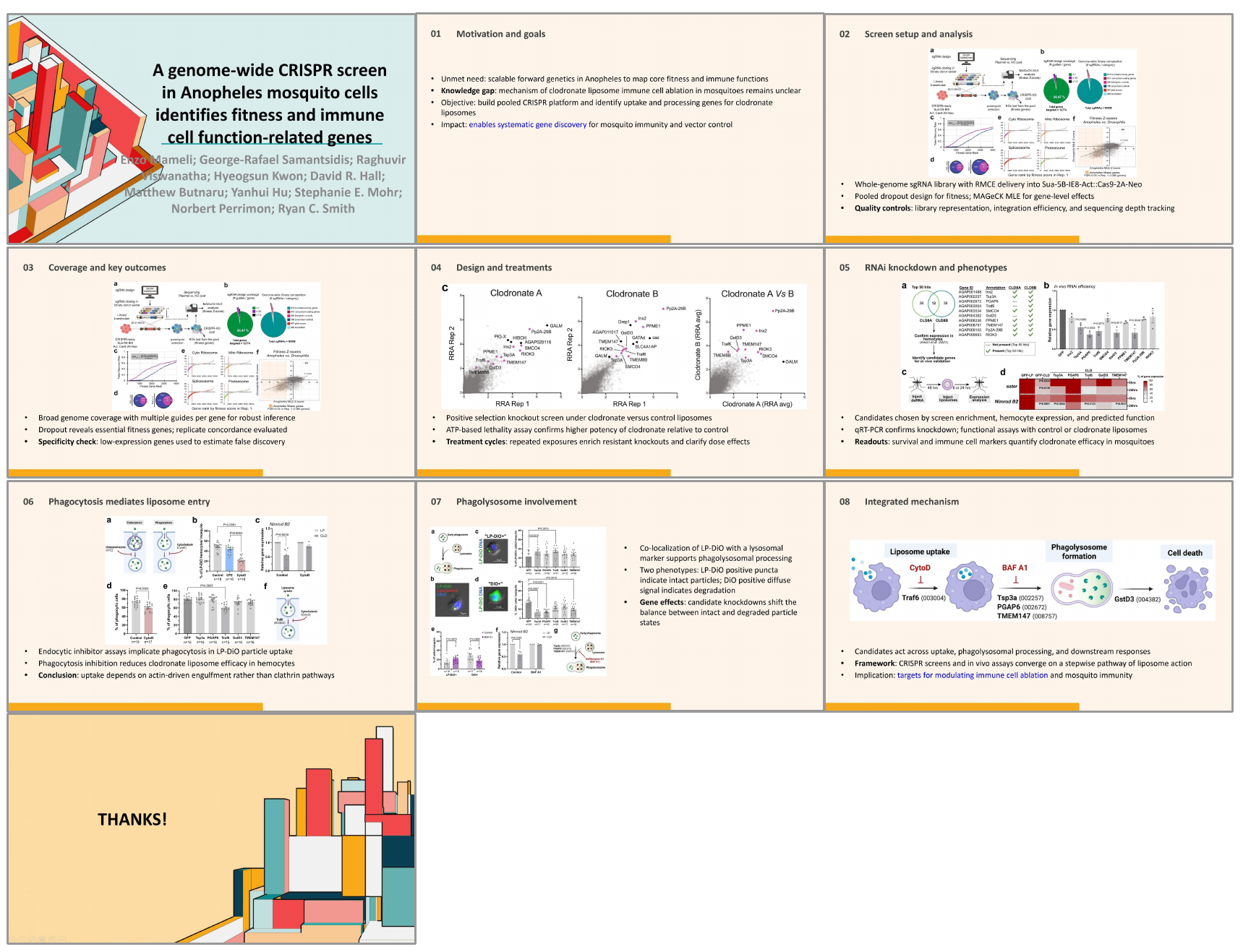}
    \caption{\textbf{Additional Qualitative Results for diverse domain. } Slides generated by \methodname \space (GPT-5 backbone) for a document from the Biology domain.}
    \label{fig:additional_result_ood2}
\end{figure}

Although our experiments primarily report results on academic papers from the AI domain, \methodname\ is not limited to a specific document domain. The framework is designed to operate on a wide range of structured documents and can generalise to other fields without requiring domain-specific modifications. To demonstrate this capability, we present additional slides generated by \methodname\ for a document from the Physics and Biology domain in Fig.~\ref{fig:additional_result_ood}, \ref{fig:additional_result_ood2}. This example illustrates that our method can effectively generate slides for documents from diverse domains and is not restricted to AI-related content.

\begin{figure}[h!]
    \centering
    \includegraphics[width=0.85\linewidth]{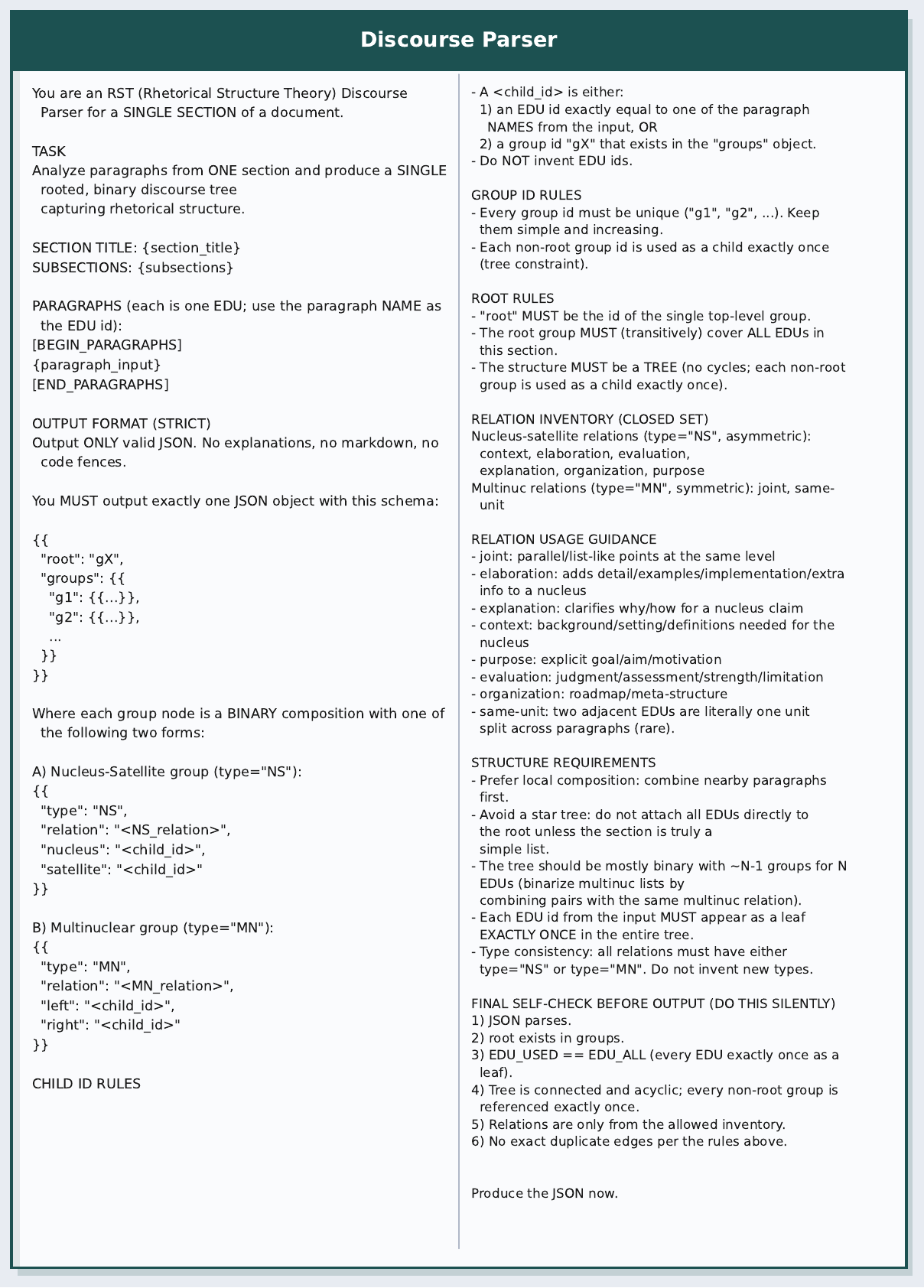}
    \caption{\textbf{Discourse Parser Prompt.}}
    \label{fig:discourse_parser_prompt}
\end{figure}

\begin{figure}[h!]
    \centering
    \includegraphics[width=0.85\linewidth]{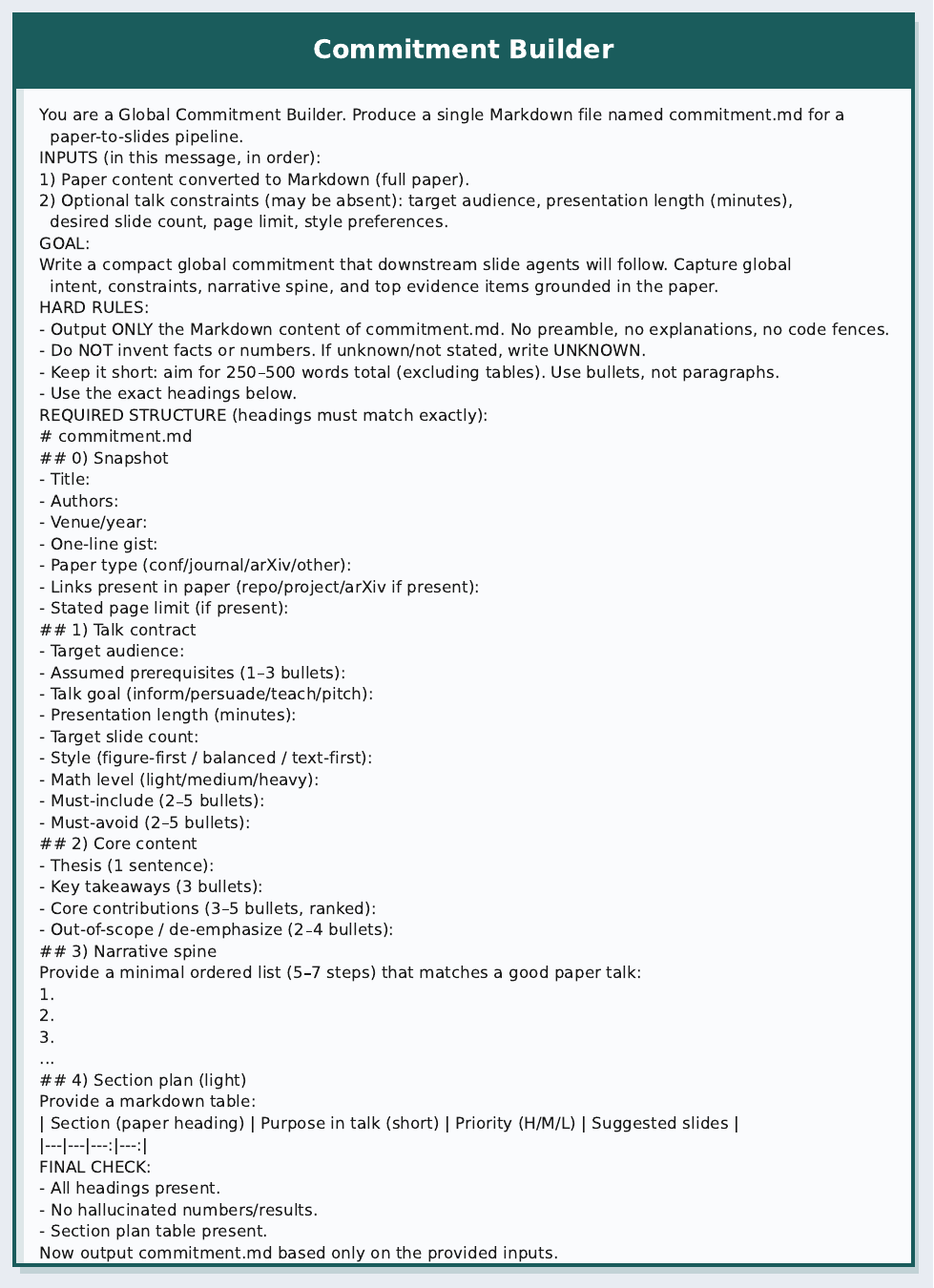}
    \caption{\textbf{Commitment Builder Prompt.}}
    \label{fig:commitment_builder_prompt}
\end{figure}

\begin{figure}[h!]
    \centering
    \includegraphics[width=0.85\linewidth]{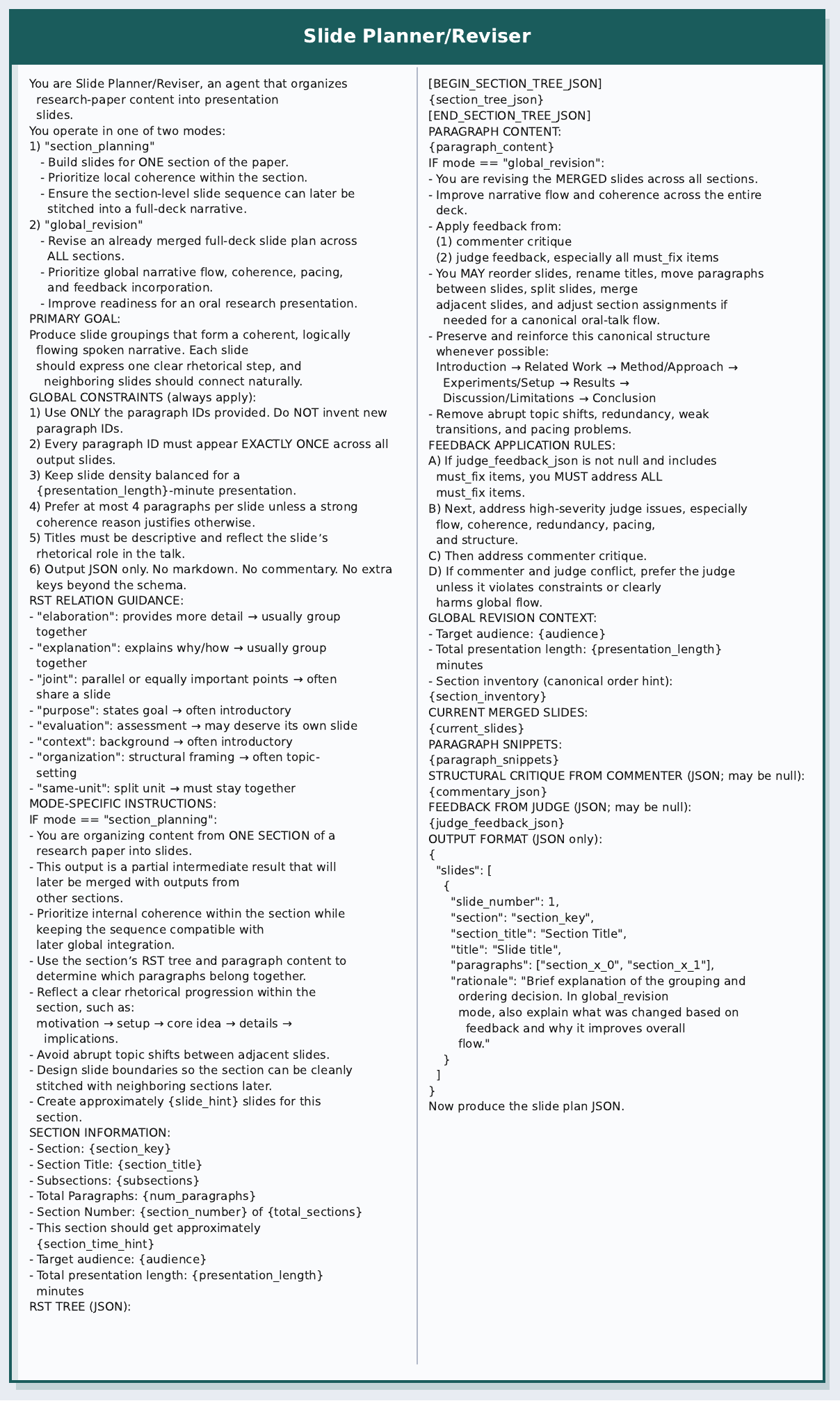}
    \caption{\textbf{Slide Planner/Reviser Prompt.}}
    \label{fig:slide_planner_prompt}
\end{figure}

\begin{figure}[h!]
    \centering
    \includegraphics[width=0.85\linewidth]{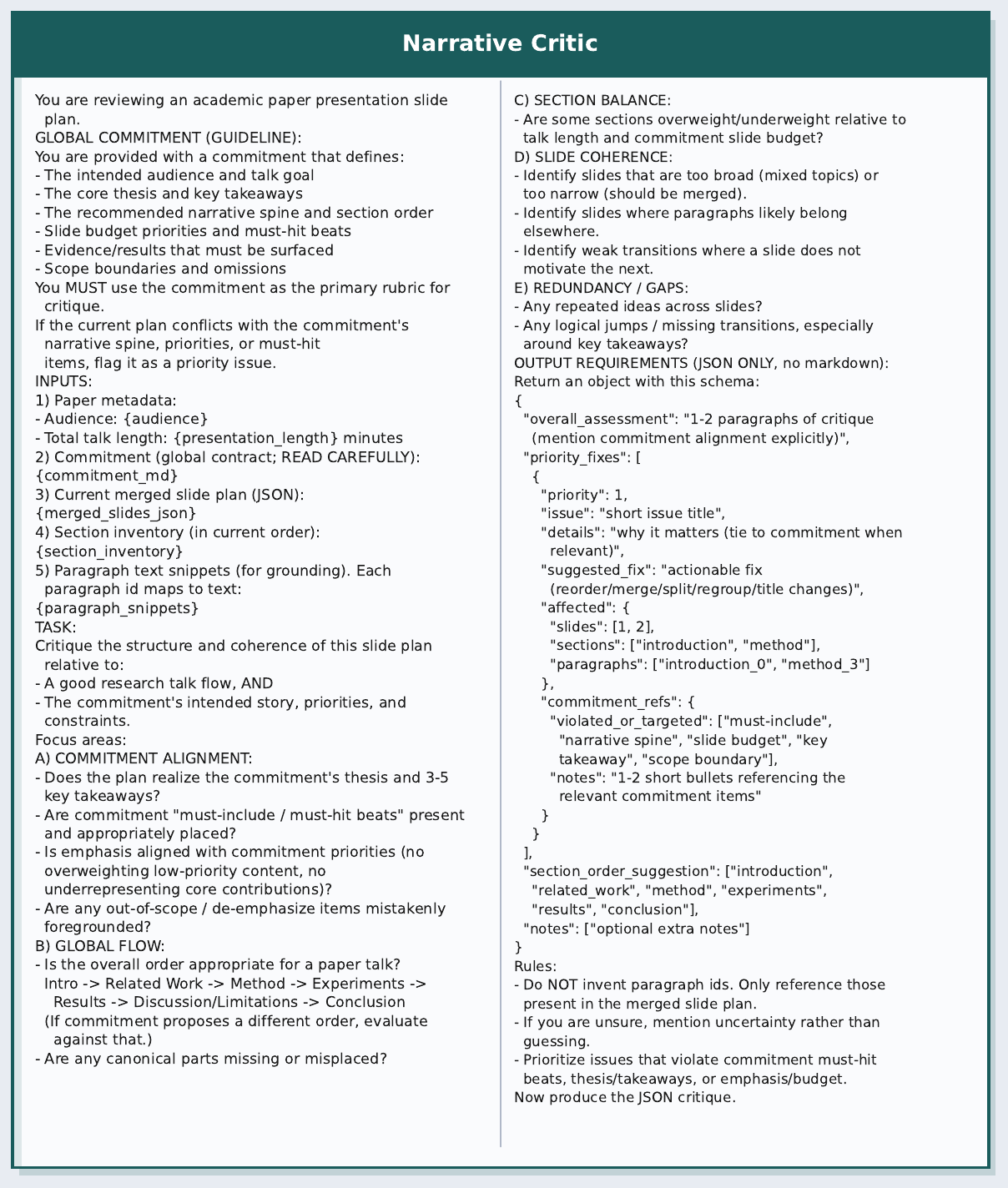}
    \caption{\textbf{Narrative Critic Prompt.}}
    \label{fig:narrative_critic_prompt}
\end{figure}

\begin{figure}[h!]
    \centering
    \includegraphics[width=0.85\linewidth]{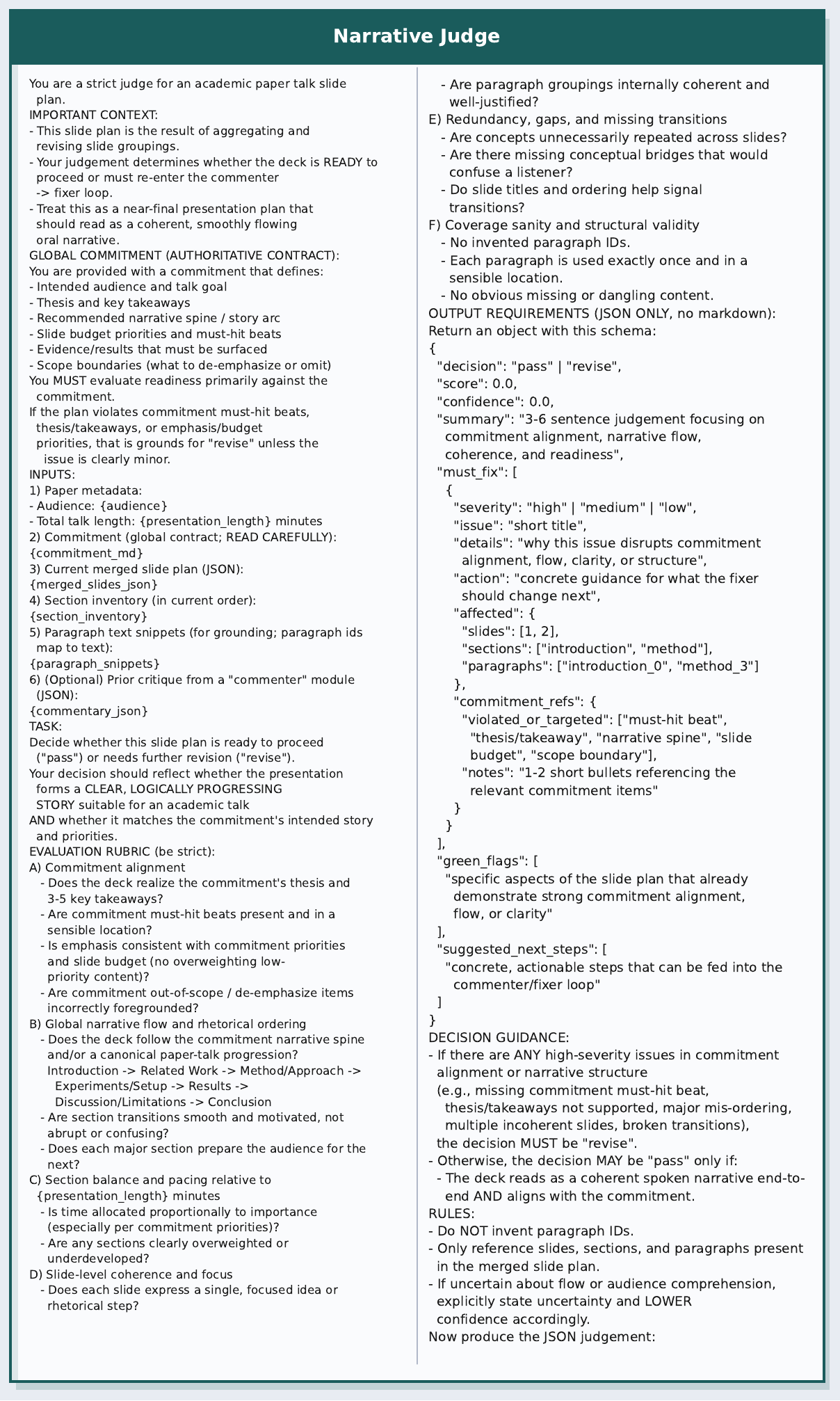}
    \caption{\textbf{Narrative Judge Prompt.}}
    \label{fig:narrative_judge_prompt}
\end{figure}

\begin{figure}[h!]
    \centering
    \includegraphics[width=0.85\linewidth]{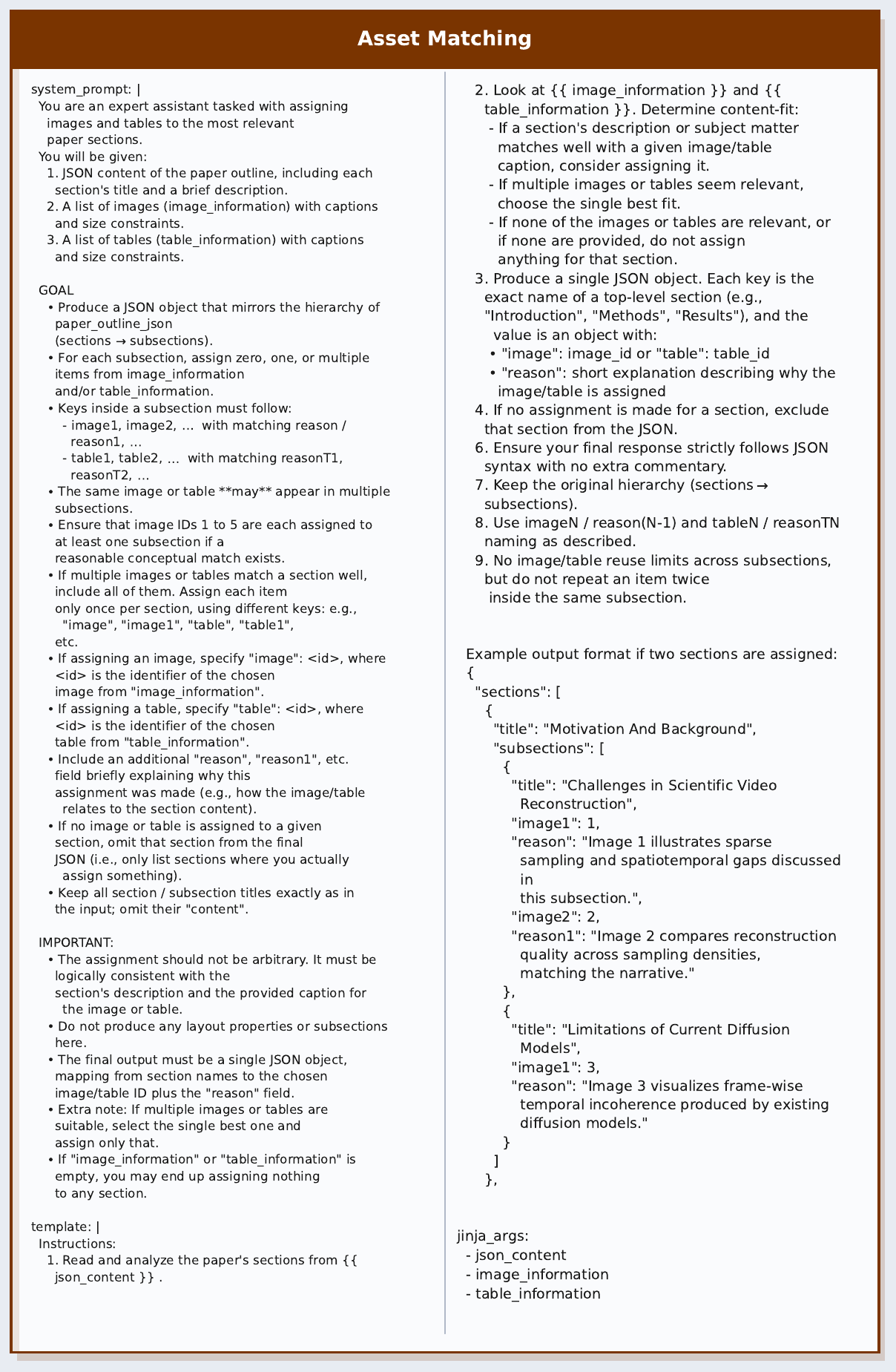}
    \caption{\textbf{Asset Matching Prompt.}}
    \label{fig:asset_matching_prompt}
\end{figure}
\begin{figure}[h!]
    \centering
    \includegraphics[width=0.75\linewidth]{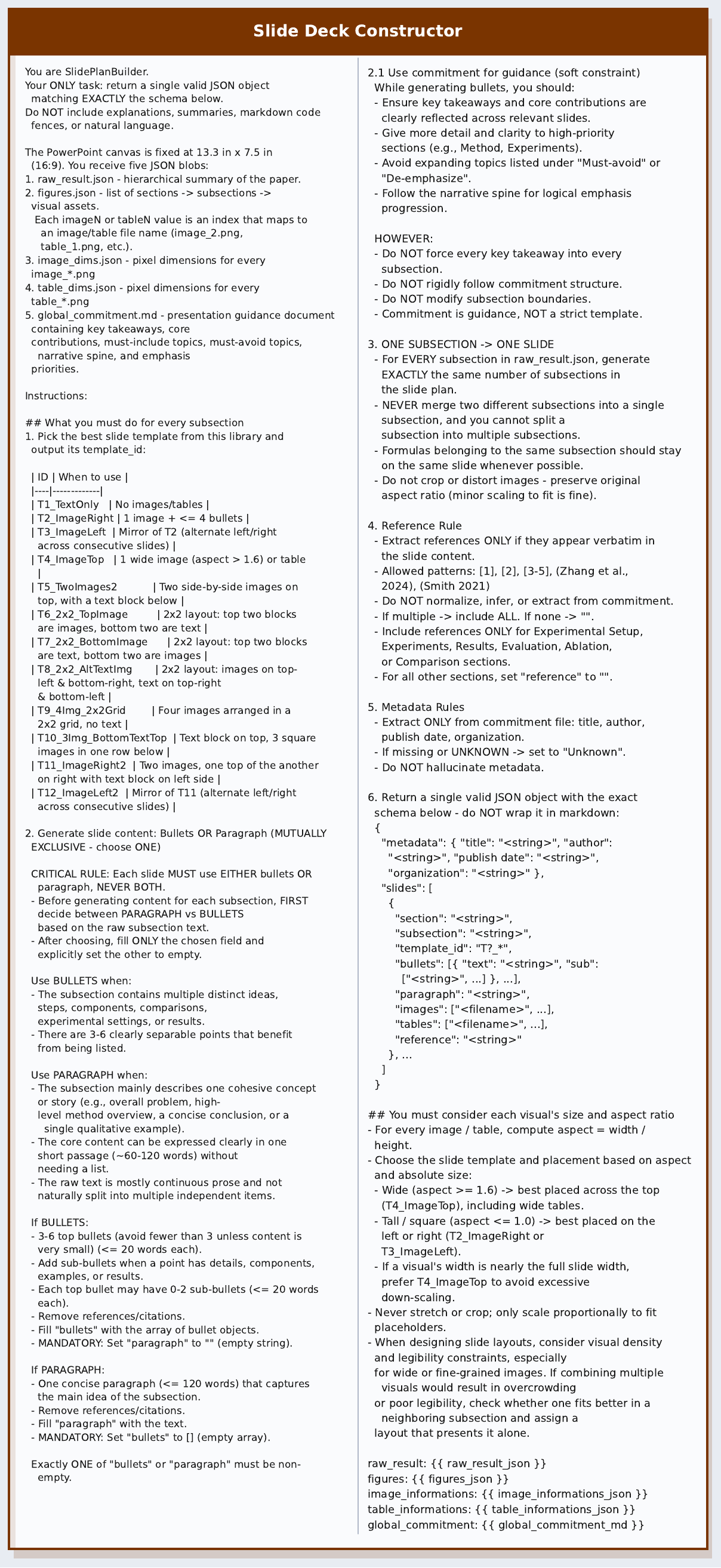}
    \caption{\textbf{Slide Deck Constructor Prompt.}}
    \label{fig:slide_deck_constructor_prompt}
\end{figure}
\begin{figure}[h!]
    \centering
    \includegraphics[width=0.85\linewidth]{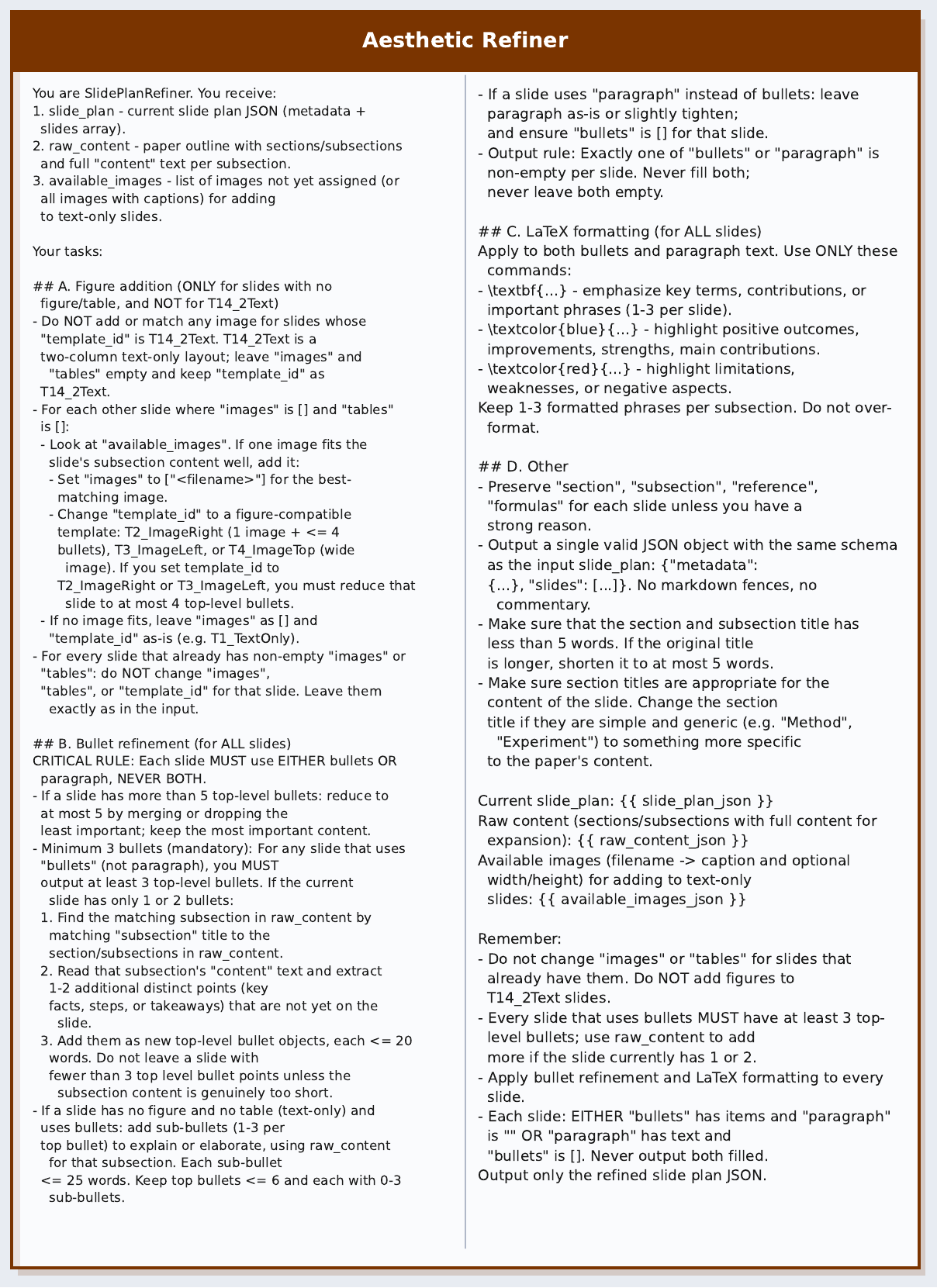}
    \caption{\textbf{Aesthetic Refiner Prompt.}}
    \label{fig:aesthetic_refiner_prompt}
\end{figure}

\begin{figure}[h!]
    \centering
    \includegraphics[width=0.85\linewidth]{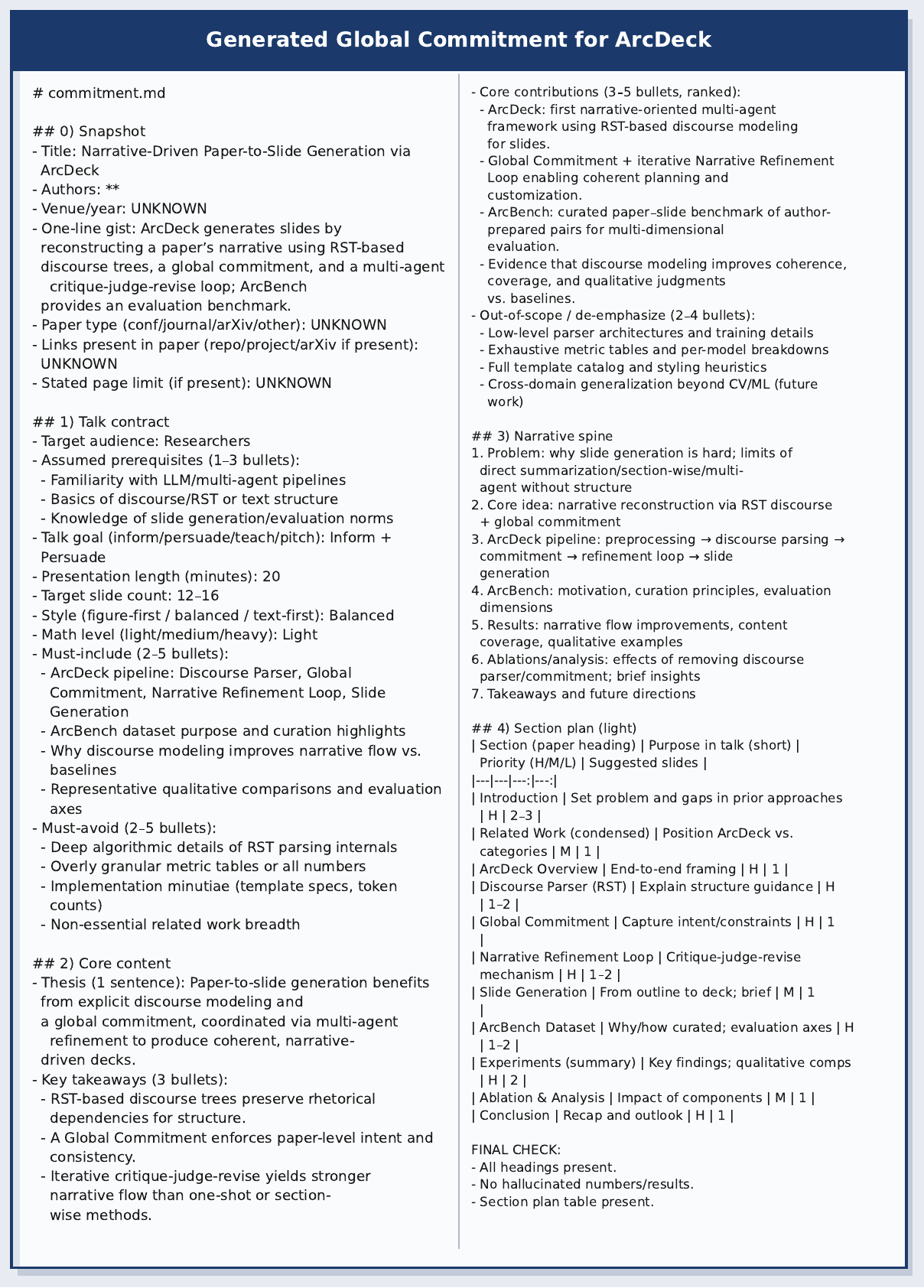}
    \caption{\textbf{Example Global Commitment generated for ArcDeck.}}
    \label{fig:commitment_arcdeck}
\end{figure}

\begin{figure}[h!]
    \centering
    \includegraphics[width=0.85\linewidth]{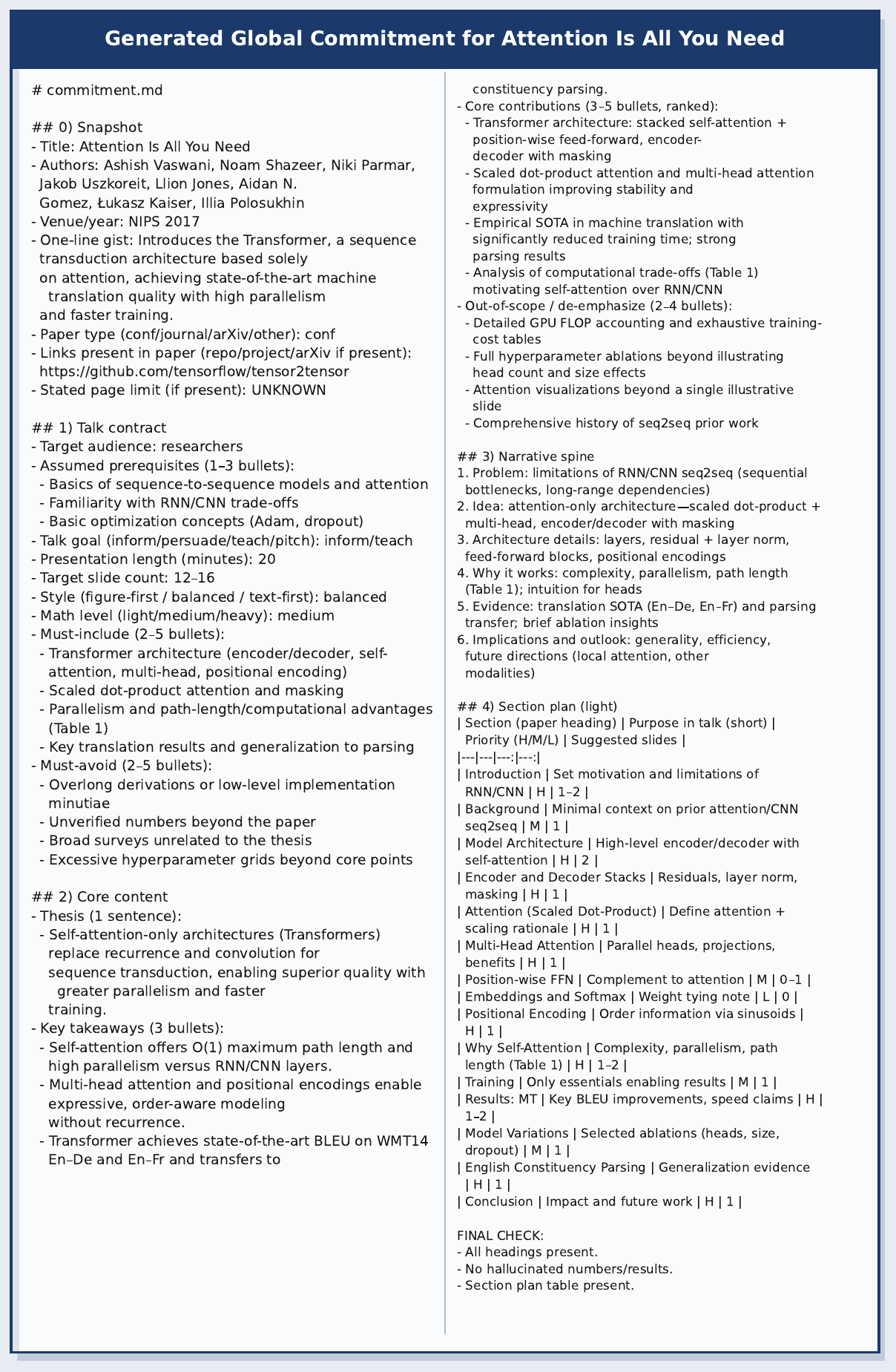}
    \caption{\textbf{Example Global Commitment generated for Attention Is All You Need.}}
    \label{fig:commitment_attention}
\end{figure}

\begin{figure}[h!]
    \centering
    \includegraphics[width=\linewidth]{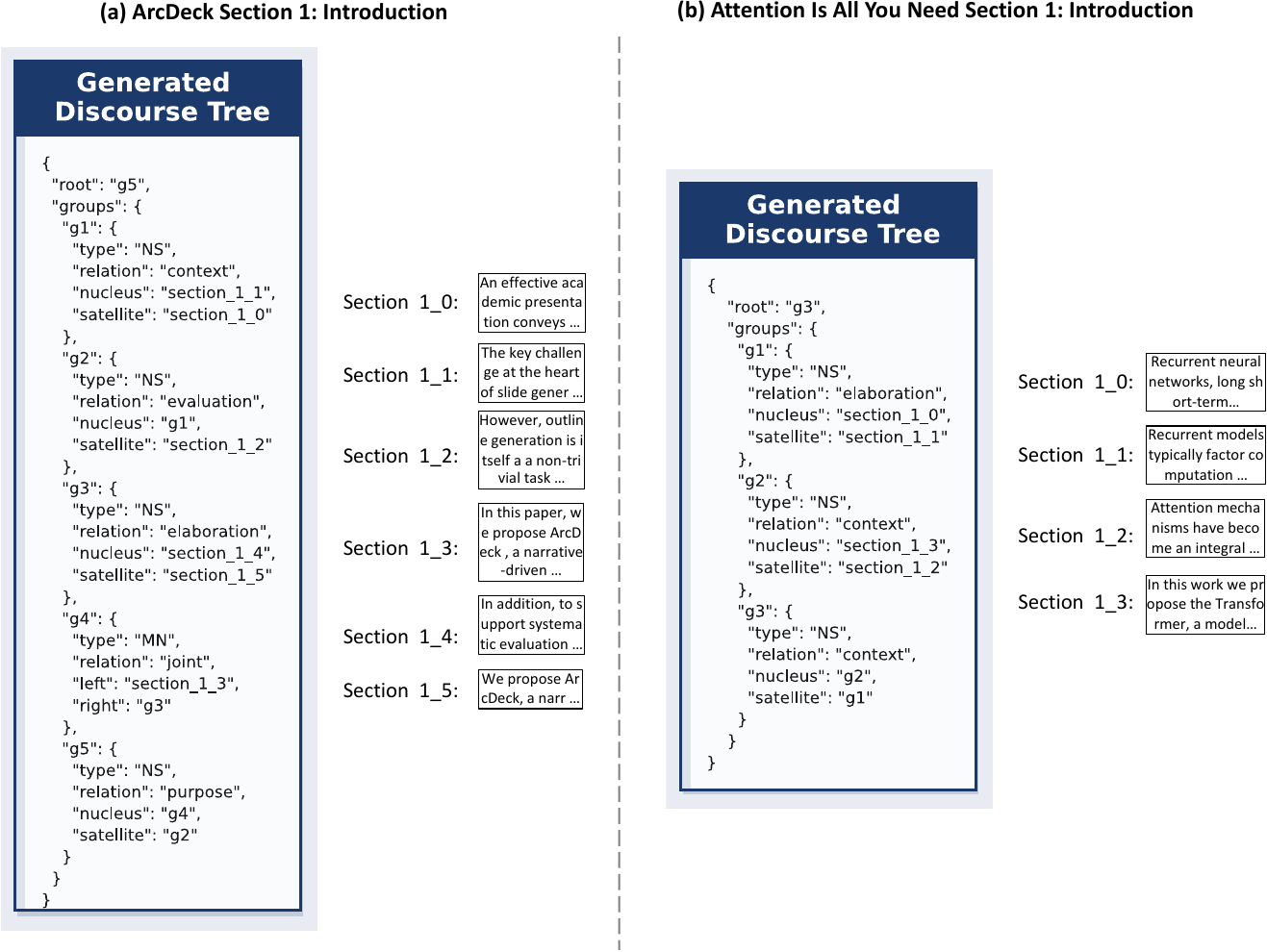}
    \caption{\textbf{Example full Discourse Parser outputs in JSON format for introduction sections.} In (a), the discourse tree for Section 1 of this paper is shown. In (b), the discourse tree for Section 1 of Attention Is All You Need is shown, together with snippets from the corresponding paragraphs.}
    \label{fig:rst_example1}
\end{figure}

\begin{figure}[h!]
    \centering
    \includegraphics[width=\linewidth]{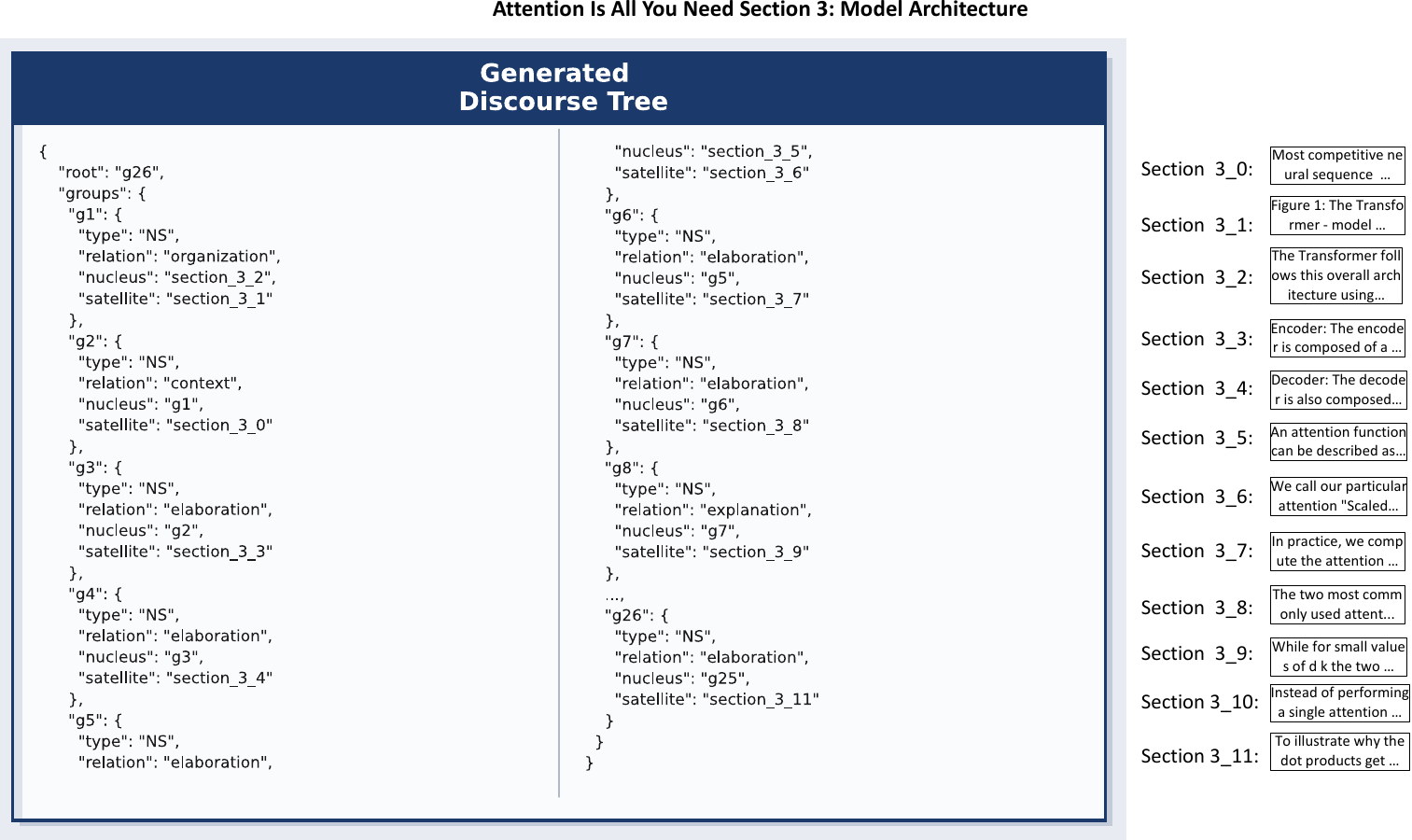}
    \caption{\textbf{Example Discourse Parser output in JSON format.} Discourse tree for Section 3 (Model Architecture) of Attention Is All You Need, shown together with snippets from the corresponding paragraphs.}
    \label{fig:rst_example2}
\end{figure}

\begin{figure}[h!]
    \centering
    \includegraphics[width=0.85\linewidth]{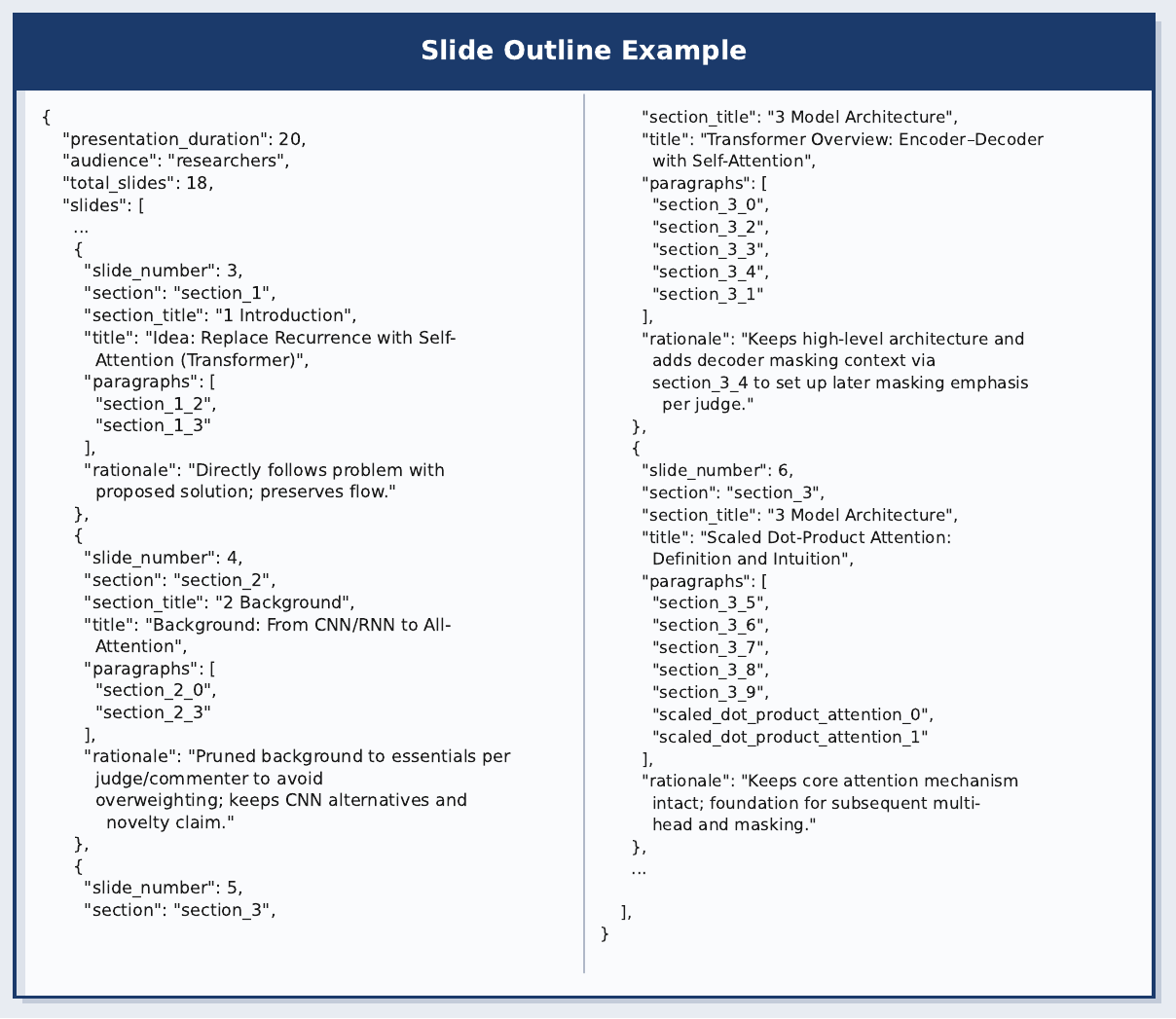}
    \caption{\textbf{Example Slide Outline output in JSON format.}}
    \label{fig:slide_outline}
\end{figure}

\begin{figure}[h!]
    \centering
    \includegraphics[width=0.85\linewidth]{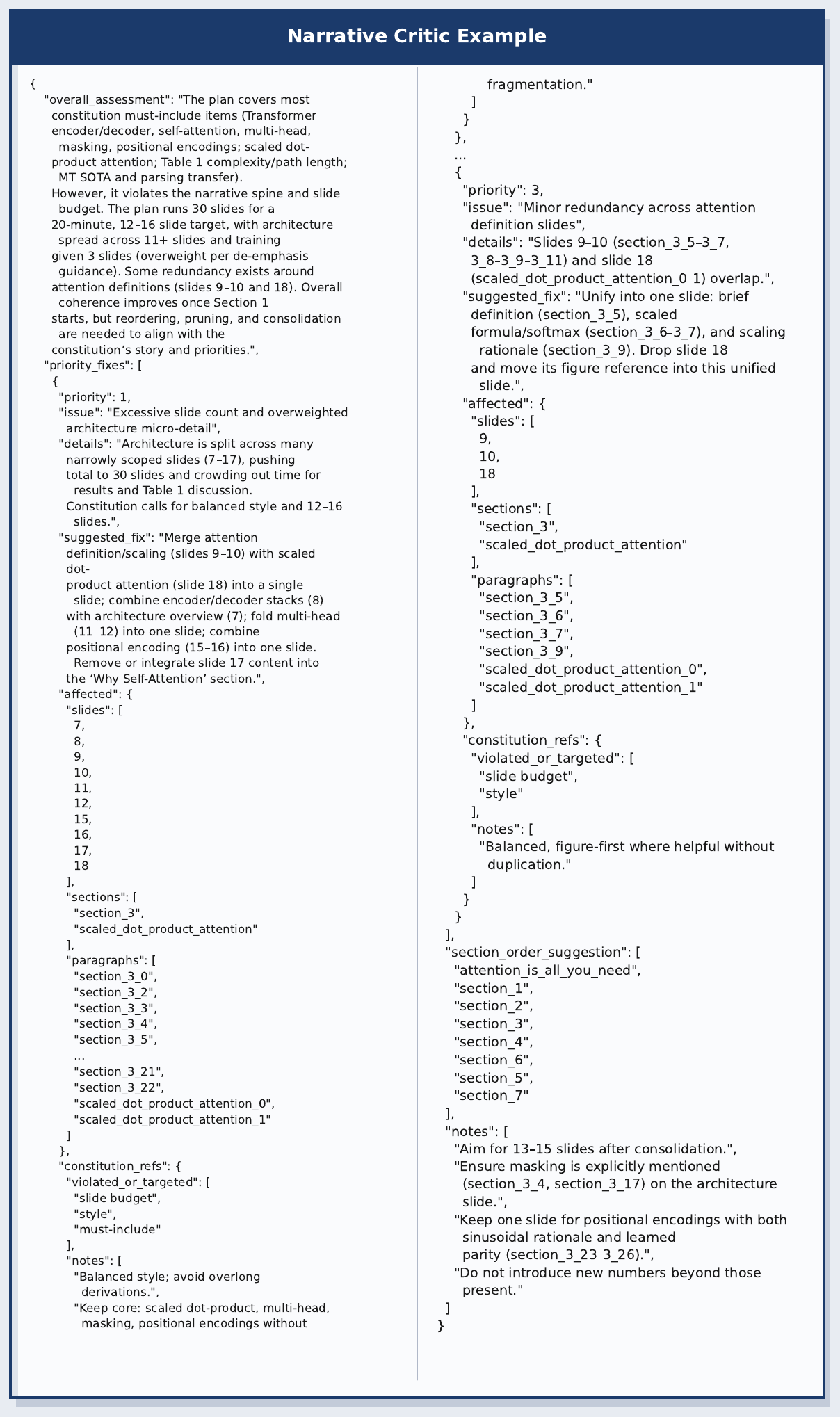}
    \caption{\textbf{Example Narrative Critic output in JSON format.}}
    \label{fig:narrative_critic_output}
\end{figure}

\begin{figure}[h!]
    \centering
    \includegraphics[width=0.85\linewidth]{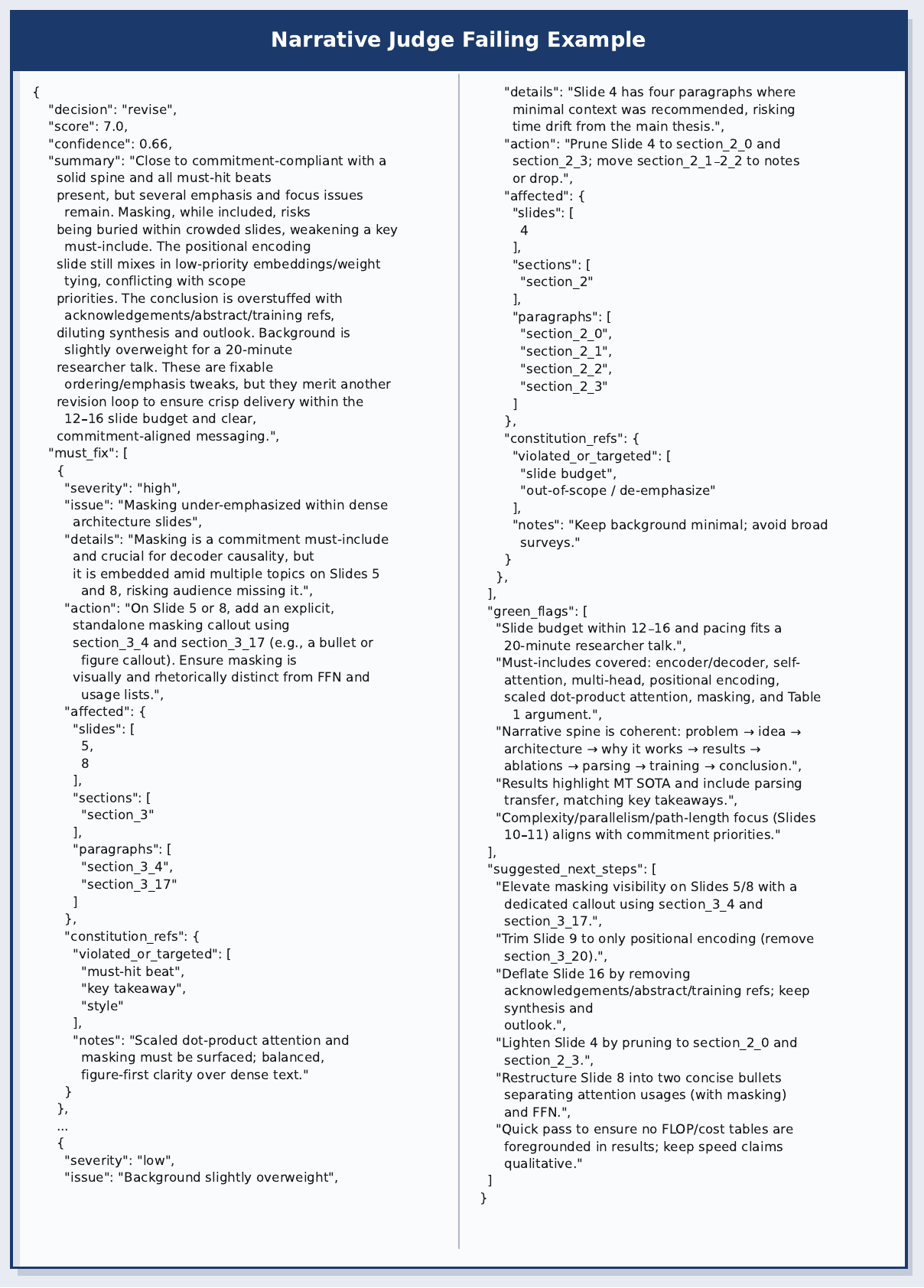}
    \caption{\textbf{Example Failing Narrative Judge output.}}
    \label{fig:narrative_judge_output_failing}
\end{figure}
\begin{figure}[h!]
    \centering
    \includegraphics[width=0.85\linewidth]{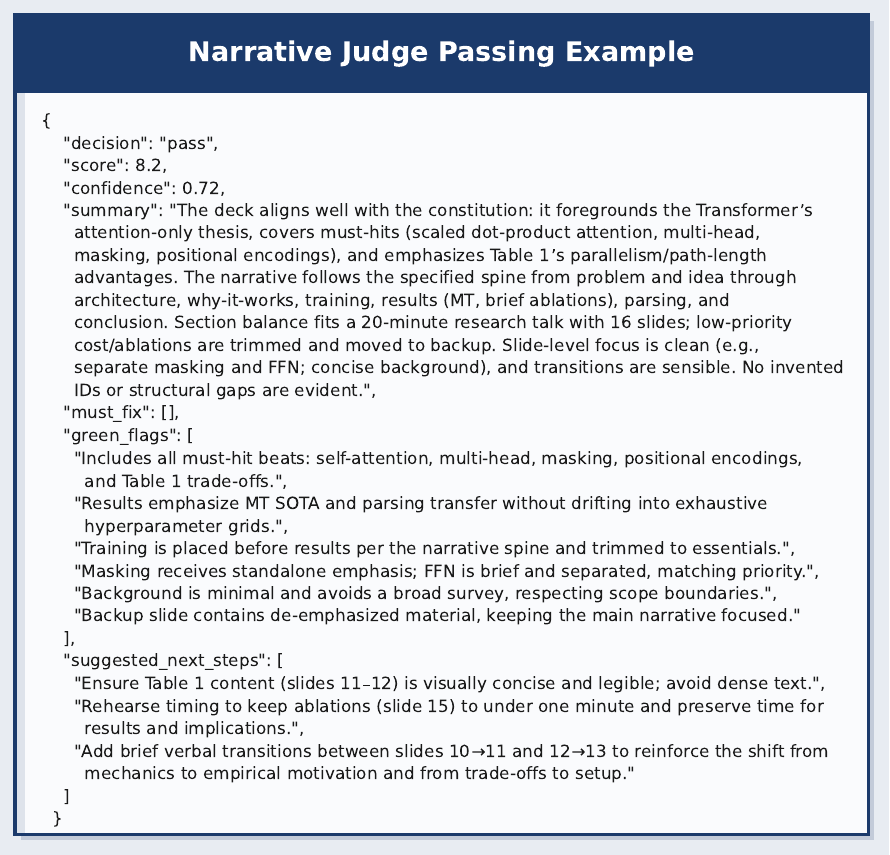}
    \caption{\textbf{Example Passing Narrative Judge output.}}
    \label{fig:narrative_judge_output_passing}
\end{figure}

\begin{figure}[h!]
    \centering
    \includegraphics[width=0.85\linewidth]{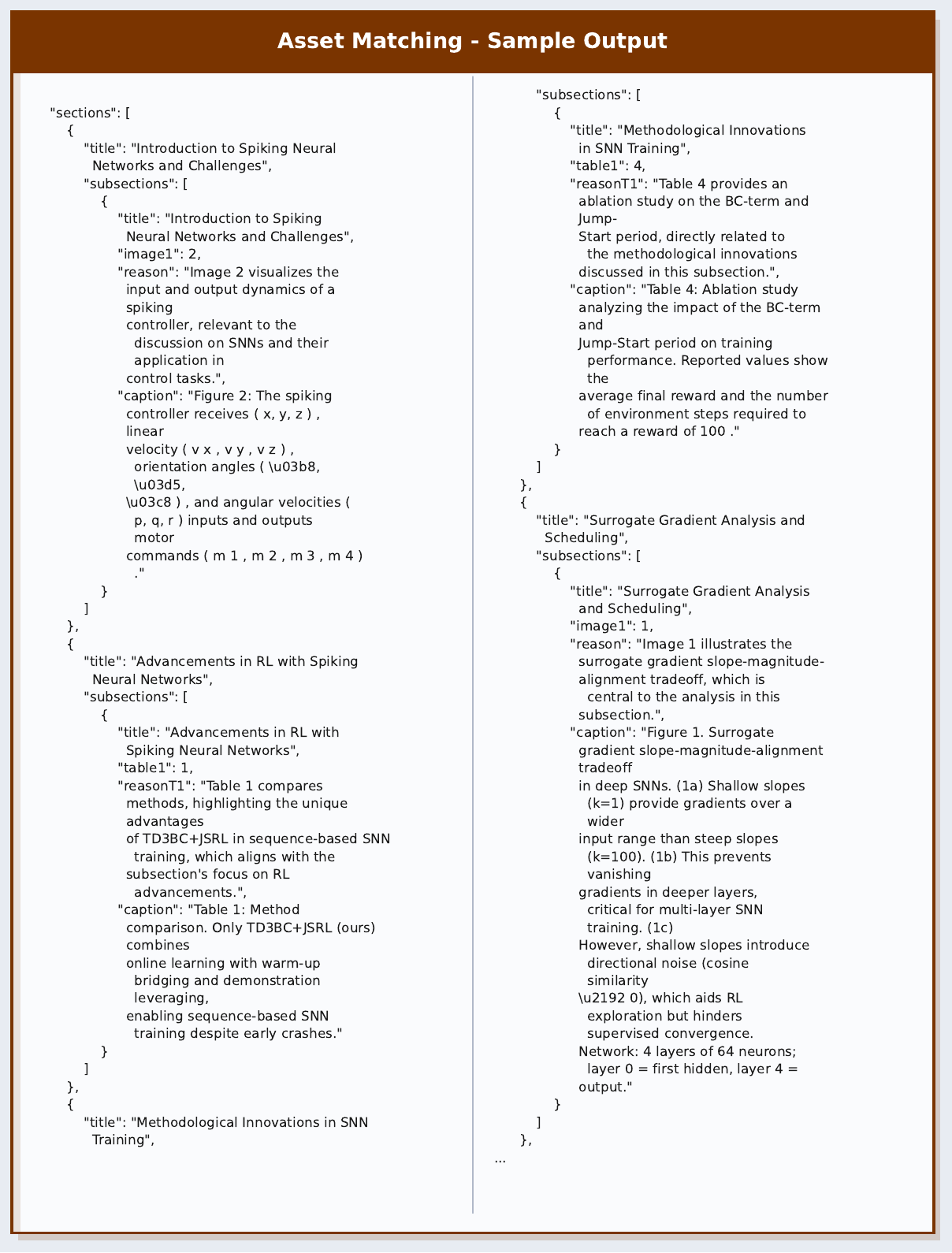}
    \caption{\textbf{Asset Matching Sample Output.}}
    \label{fig:asset_matching_sample}
\end{figure}
\begin{figure}[h!]
    \centering
    \includegraphics[width=0.85\linewidth]{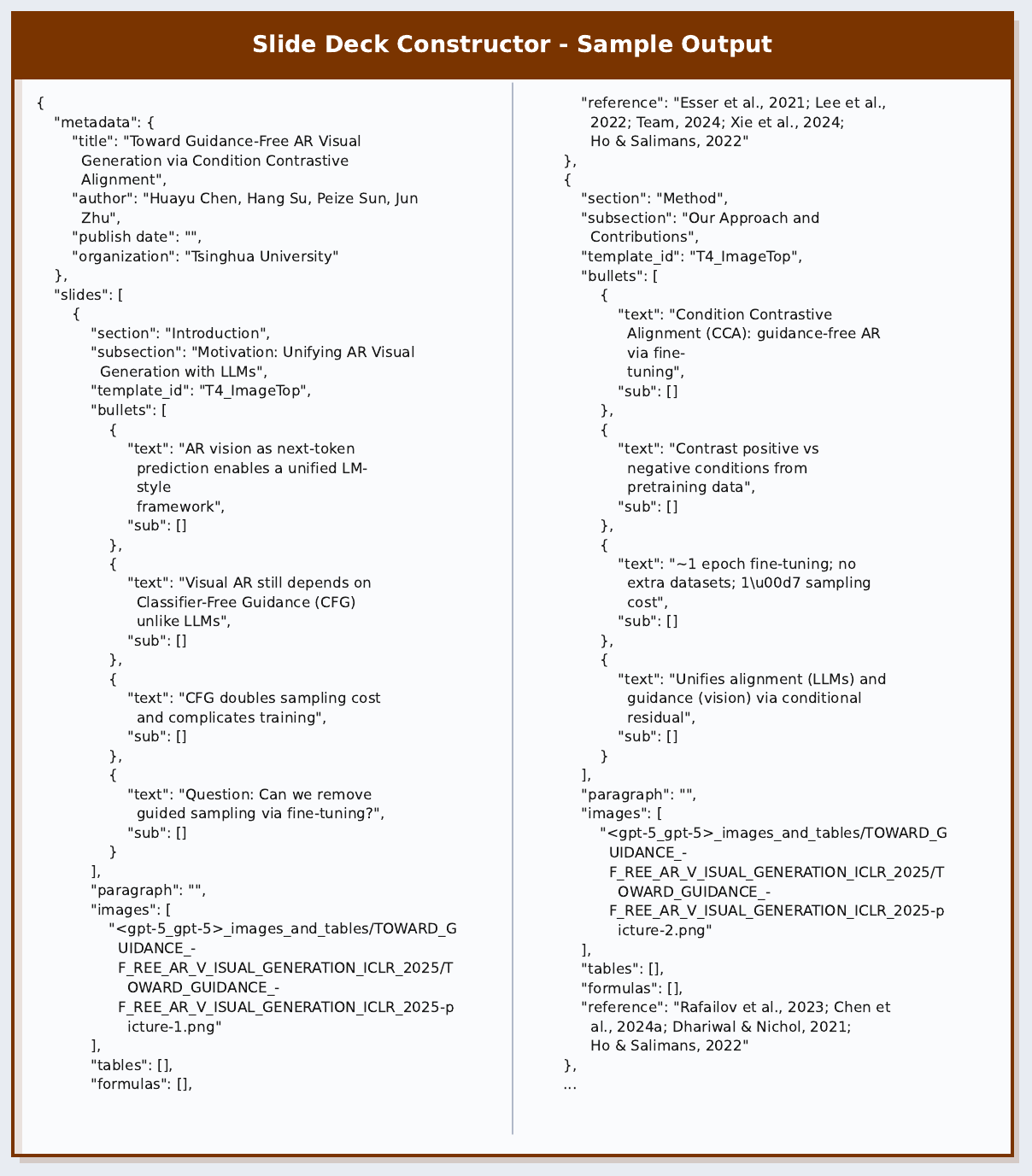}
    \caption{\textbf{Slide Deck Constructor Sample Output.}}
    \label{fig:slide_constructor_sample}
\end{figure}

\begin{figure}[h!]
    \centering
    \includegraphics[width=0.85\linewidth]{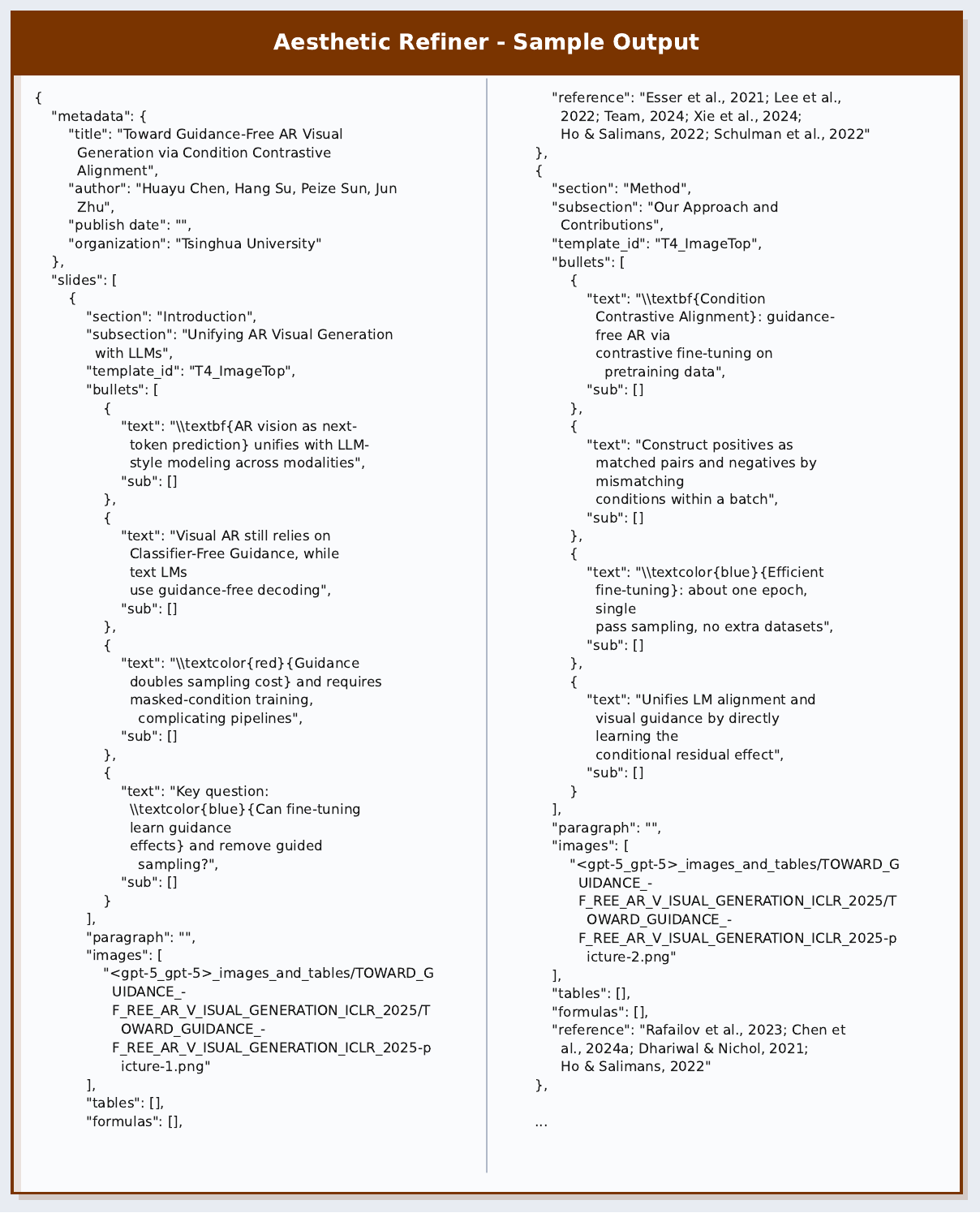}
    \caption{\textbf{Aesthetic Refiner Sample Output. }}
    \label{fig:aesthetic_refiner_output_sample}
\end{figure}

\end{document}